# Vehicle-in-Virtual-Environment (VVE) Method for Developing and Evaluating VRU Safety of Connected and Autonomous Driving with Focus on Bicyclist Safety

Haochong Chen, Xincheng Cao, Bilin Aksun-Guvenc, Levent Guvenc

*Automated Driving Lab*

*Department of Mechanical and Aerospace Engineering*

*Ohio State University*

# Contents

# Chapter 1: Overview

With rapid urbanization and technological development, the number of privately owned vehicles has dramatically increased each year. The excessive numbers of private vehicles have led to severe traffic congestion and an alarming increase in road accidents, which gradually become a new set of challenges that every modern city must confront. According to the Global Status Report on Road Safety by the World Health Organization (WHO), over 50 million people are injured, and 1.3 million individuals lose their lives each year due to car accidents, with approximately 75% of these incidents attributed to human errors such as drowsy driving, driving under the influence, and distracted driving [1-3]. Vulnerable road users (VRUs), including pedestrians and bicyclists, are particularly at risk, suffering disproportionately high fatality rates in collisions, which may need extra attention to improve their safety and reduce accident severity. Autonomous Driving Systems (ADS), benefit from powerful and robust autonomous driving algorithms and hence offer a promising solution for reducing human error and enhancing road safety. Extensive research has already been conducted in the autonomous driving field to help vehicles navigate safely and efficiently. At the same time, plenty of current research on vulnerable road user (VRU) safety are performed which largely concentrates on perception, localization, or trajectory prediction of VRUs [4-8]. However, existing research still exhibits several gaps, including the lack of a unified planning and collision avoidance system for autonomous vehicles, limited investigation into delay-tolerant control strategies, and the absence of an efficient and standardized testing methodology. Ensuring VRU safety remains one of the most pressing challenges in autonomous driving, particularly in dynamic and unpredictable environments.

In this two-year project, we focused on applying the Vehicle-in-Virtual-Environment (VVE) method to develop, evaluate, and demonstrate safety functions for Vulnerable Road Users (VRUs) using automated steering and braking of ADS. In the current second year project, our primary focus was on enhancing the previous year's results while also considering bicyclist safety. We began by analyzing five key bicyclist crash scenarios identified by the Fatality Analysis Reporting System (FARS), an organization under the National Highway Traffic Safety Administration (NHTSA) that compiles vehicle crash data. The bicyclist crash scenarios we examined include: "Motorist Overtaking Bicyclist" (FARS 230), "Bicyclists Failed to Yield, Midblock" (FARS 310), "Bicyclist Failed to Yield Sign, Controlled Intersection" (FARS 145), "Bicyclist Left Turn / Merge" (FARS 220), and "Motorist Left Turn / Merge" (FARS 210). Figure 1.1 demonstrates these five bicyclist crash cases. These scenarios along with the pedestrian crash scenarios considered in our first-year project formed the basis of our research. We recreated these five traffic crash scenarios using the CARLA (Car Learning to Act) virtual environment to ensure realistic and accurate simulations for our analysis. The detailed traffic crash scenario simulation videos are provided in the following. Videos Link: 1. FARS230: https://youtu.be/Tk3ZEXVd7ow 2. FARS310: https://youtu.be/s7spUE1gCek 3. FARS145: https://youtu.be/vD5c2wH2PFU 4. FARS220: https://youtu.be/c5bRPeQJjD0 5. FARS210: https://youtu.be/yrBi7FteFfA.

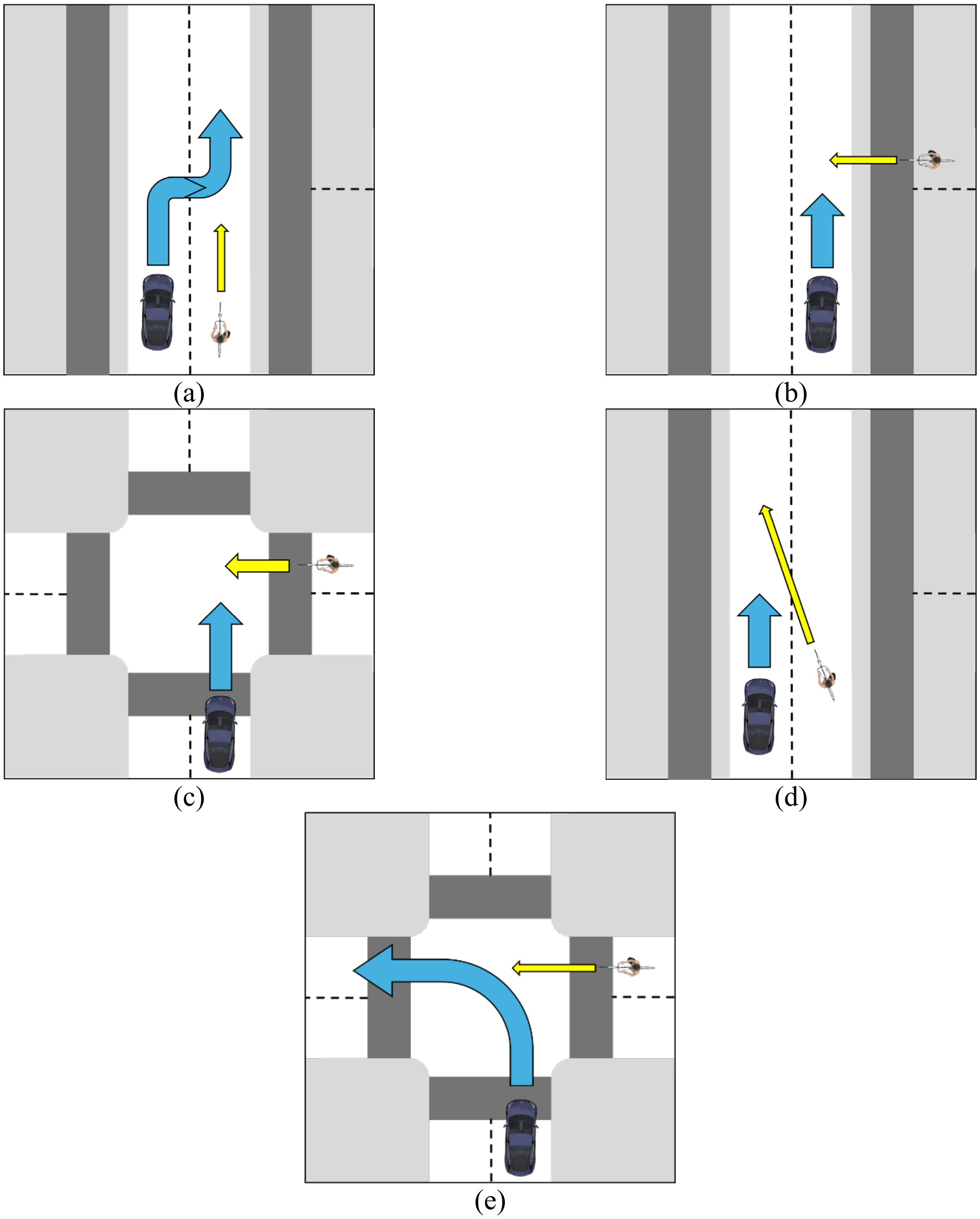

**Figure 1.1.** Bicyclist FARS Crash Scenarios (a). FARS 230: Motorist Overtaking Bicyclist (b). FARS 310: Bicyclists Failed to Yield, Midblock (c). FARS 145: Bicyclist Failed to Yield Sign, Controlled Intersection (d). FARS 220: Bicyclist Left Turn / Merge (e). FARS 210: Motorist Left Turn / Merge

Building on these cases, our research fills critical gaps in the current literature by proposing practical solutions to the types of VRU-related crashes represented in the FARS dataset. We aimed to systematically develop and test autonomous driving technologies that can prevent such incidents from occurring. Specifically, this research makes the following three contributions:

(1) Proposing a communication-disturbance-observer (CDOB) based delay-tolerant control strategy to enhance robustness of the control under computation latency, network latency and packet loss.

(2) Developing a hierarchical control framework that integrates deep reinforcement learning (DRL) for high-level decision-making with a CLF-CBF-QP-based controller for safe and smooth low-level execution.

(3) Introducing a novel VVE-based testing pipeline that enables efficient and rigorous evaluation of autonomous driving functions under various complex and high-risk traffic scenarios.

In addition, this research also incorporates existing perception methods for VRU detection and trajectory prediction, which serve as a benchmark and provide a solid foundation for integrating these components into a unified, end-to-end autonomous driving strategy in future development.

The study starts with a comprehensive literature review that addresses VRU detection and trajectory prediction which is presented in Chapter 2. In Chapter 3, a delay-tolerant control strategy based on a communication disturbance observer (CDOB) is introduced to ensure stable and accurate path-tracking performance under undesirable computation and communication delays. In Chapter 4, we propose a novel hierarchical control framework that integrates traditional optimization-based control with deep reinforcement learning (DRL), aiming to achieve both high performance and robustness. This framework is implemented and validated on two distinct vehicle models to demonstrate its generalizability and effectiveness. In Chapter 5, a novel testing approach called Vehicle-in-Virtual-Environment (VVE) is proposed to enable efficient, safe, and scalable validation of autonomous driving functions across different deployment stages. Finally, in Chapter 6, we conclude our second-year research and outline future work.

The resulting publications of this project are listed below.

# Chapter 2: Vulnerable Road User (VRU) Detection

## 2.1 Camera Based Vulnerable Road User (VRU) Detection

Designing robust path planning and collision avoidance functions starts with the accurate perception of the surrounding traffic environment, particularly the detection of VRUs including pedestrians and bicyclists. In complex traffic scenarios, it is essential to reliably identify VRUs, accurately estimate their location and orientation, and predict their future trajectories based on raw sensor data. This perception stage serves as the foundation for effective decision-making in autonomous driving systems.

Extensive research has demonstrated the effectiveness of using camera images for obstacle identification and categorization. High-resolution visual data from cameras enable the precise detection of VRUs, while advanced computer vision algorithms facilitate their classification. Currently, there are two approaches to perform obstacle detection using image processing. The first approach is to perform object detection using a two-stage process: proposal generation and object detection. The proposal generation step uses a selective search algorithm to generate multiple region proposals which indicate potential object locations. Then, a neural network (usually convolutional neural network) is applied to classify the object within the proposed region and refine its bounding box. Girshick et al. introduced R-CNN, a novel framework that leverages rich feature hierarchies from pre-trained convolutional neural networks to perform accurate object detection and semantic segmentation [9]. Building on the foundation of R-CNN, Girshick further refined the model with Fast R-CNN by integrating the region proposal and feature extraction steps. This integration was achieved by introducing a Region of Interest (RoI) pooling layer that extracts a fixed-length feature vector from the feature map for each object proposal, followed by fully connected layers that classify the features into object categories and regress the bounding box coordinates [10]. Ren et al. explored the concept further and proposed Faster R-CNN. This algorithm incorporated a Region Proposal Network (RPN) that shares full-image convolutional features with the detection network, thus enabling nearly cost-free region proposals [11]. Sun et al., introduced Sparse R-CNN which simplifies the previously complex pipeline and reduces the dependency on heuristic design, pushing the boundaries of object detection with a sparse set of highly effective proposals [12]. However, the major limitations of the R-CNN architecture are its deficiency in real-time performance caused by complicated procedure and its computational complexity. The second approach is to merge the aforementioned two stages by applying a single neural network to the whole image, dividing the image into regions and predicting bounding boxes and probabilities for each region simultaneously. Redmon introduced a convolutional neural network (CNN)-based architecture known as "YOLO" (You Only Look Once) for multi-object detection. This architecture segments an image into multiple small grids, assigning each grid the task of detecting objects whose center points fall within its boundaries. For each grid cell, the model predicts multiple bounding boxes and assigns labels to these boxes, each with associated class probabilities. To enhance the accuracy and reduce redundancy, the model employs Non-Maximum Suppression (NMS) to eliminate overlapping bounding boxes that detect the same object [13]. Currently, since its debut, the YOLO architecture has undergone numerous iterations and enhancements, evolving to its twelfth generation—YOLOv12. These iterations have not only improved the model's accuracy and speed but also enabled YOLO to effortlessly recognize a wide variety of objects [14].

## 2.2 Lidar Data Processing and Vulnerable Road User (VRU) Trajectory Prediction

Lidar technology offers significant advantages over traditional front-facing vehicle cameras, particularly in the field of autonomous vehicles [15]. Early work on automotive applications used two-dimensional (2D) LiDAR for object detection with Kalman filtering for prediction [16]. Unlike cameras, which are limited to capturing visual data from the front of the vehicle, three dimensional (3D) Lidar provides a 360-degree view, allowing it to detect obstacles all around the vehicle. This comprehensive coverage is important for the complex decision making required in autonomous driving. Lidar sensors work by emitting laser beams and measuring the time it takes for the reflection to return, thereby creating detailed and accurate 3D maps of the environment. This capability makes Lidar exceptionally good at detecting and tracking pedestrians, even in challenging conditions such as low light or obstructed views. Consequently, there has been substantial research and development in the field, focusing on leveraging Lidar for pedestrian detection, which is important for improving safety and operational efficiency in autonomous vehicle technologies. It is also possible to use Vehicle-to-VRU communication to determine the pose of nearby VRUs and predict their future trajectories [17].

The integration of additional sensors, such as LiDAR, can complement visual data by providing accurate three-dimensional spatial information, thereby providing accurate estimation of VRUs' location and orientation [18-19]. Unlike cameras, which are limited to capturing visual data from the front of the vehicle, LiDAR provides a 360-degree view, allowing it to detect obstacles all around the vehicle. LiDAR sensors work by emitting laser beams and measuring the time it takes for the reflection to return, thereby creating detailed and accurate 3D maps of the environment. This capability makes LiDAR exceptionally good at detecting and tracking pedestrians and bicyclists, even in challenging conditions such as low light or obstructed views. Plenty of recent research has focused on combining LiDAR point cloud data with camera visual data to leverage the complementary strengths of both sensor types. It is also possible to combine these perception sensors with Vehicle to VRU communication for better handling of no line-of-sight cases [20]. Future Vehicle to VRU communication may also include communication with unmanned aerial vehicles (UAV) [21]. Muhammad et al focused on enhancing object detection in autonomous driving through a neural network approach that integrates visual data with LiDAR point clouds. They proposed a framework aiming to address the inaccuracies common in LiDAR detections by using separate processing streams for visual and LiDAR data, which allows for a lightweight LiDAR-only setup during runtime if needed. The approach is designed to work in real-time on embedded platforms, suggesting significant potential for practical applications in dynamic environments [22]. Sahba et al. introduced an effective method for 3D object detection using LiDAR data through the PointPillars network. Their study utilizes the nuScenes dataset to train the model for detecting cars, pedestrians, and buses, demonstrating that increasing the number of LiDAR sweeps substantially improves detection performance. Their research emphasizes the potential of integrating different types of sensor data to further enhance the encoder's effectiveness in autonomous vehicle applications [23]. Liu et al. developed a LiDAR-camera fusion algorithm for 3D object detection, focusing on autonomous driving applications. The proposed FuDNN network used a 2D backbone for image feature extraction and an attention-based fusion sub-network for integrating features from camera and LiDAR data. Their model was tested on the KITTI dataset and has shown high accuracy in detecting cars, reflecting significant improvements over existing LiDAR-camera fusion techniques [24]. Naich et al. introduced a LiDAR-based intensity-aware 3D object detection approach for outdoor environments. They proposed a voxel encoder that generates intensity histograms to enhance the feature set for robust detection, integrated within a single-stage detector. The method was evaluated using the KITTI dataset which not only matches but in some cases surpasses state-of-the-art performance, especially in detecting pedestrians and bicyclists while maintaining high frame rates during inference [25].

After estimating VRU location and orientation, the data will be used to predict their future movement pattern and trajectory. Typically, the spatial information of the surrounding traffic environment is vectorized and fed into a neural network to predict the trajectory. Zhu et al. introduced the Spatio-Temporal Graph Transformer Network (STGFNet) for predicting multi-pedestrian trajectories, leveraging both spatial and temporal data. The proposed model integrates a novel decoder structure and a memory mechanism to enhance trajectory continuity and uses HuberLoss for the first time as a loss function, showing notable improvements in prediction accuracy across multiple datasets. This research exemplifies the utility of combining transformer architectures with graph neural networks to address the dynamic complexities of pedestrian movement in crowded spaces [26].

## 2.3 Proposed Approach

From the literature review, we identified key limitations in existing perception and prediction pipelines. In response, we proposed a theoretically feasible approach that connects VRU perception to trajectory prediction in a unified manner, forming a perception and prediction component for future decision-making and control development. As described earlier, camera-based detection of VRUs involves two primary stages: proposal generation and object detection. To streamline this process, end-to-end neural networks like YOLO are employed to process RGB visual data and perform VRU detection due to its effectiveness and simplicity. The YOLO architecture implemented for VRU detection is YOLO v11, which demonstrates robust performance in identifying pedestrians and bicyclists [27]. However, the original model occasionally struggles to accurately recognize VRUs and other vehicles within the Unreal Engine 4 virtual environments. Given that the VVE evaluation method assesses the system's performance in such settings, it's imperative that the model effectively identify VRUs in these contexts. To address this challenge, we propose applying transfer learning to the pre-trained YOLOv11 model using the CARLA Object Detection Dataset [28]. This dataset, generated within the CARLA simulator (built on Unreal Engine 4), offers a large collection of labeled images that serve as valuable training data for improving object detection models. By fine-tuning YOLOv11 with this data, we aim to improve its capability to detect VRUs accurately in virtual environments. Post-transfer learning, the model is expected to proficiently utilize camera-derived visual data to identify VRUs, thereby ensuring reliable performance during VVE evaluations.

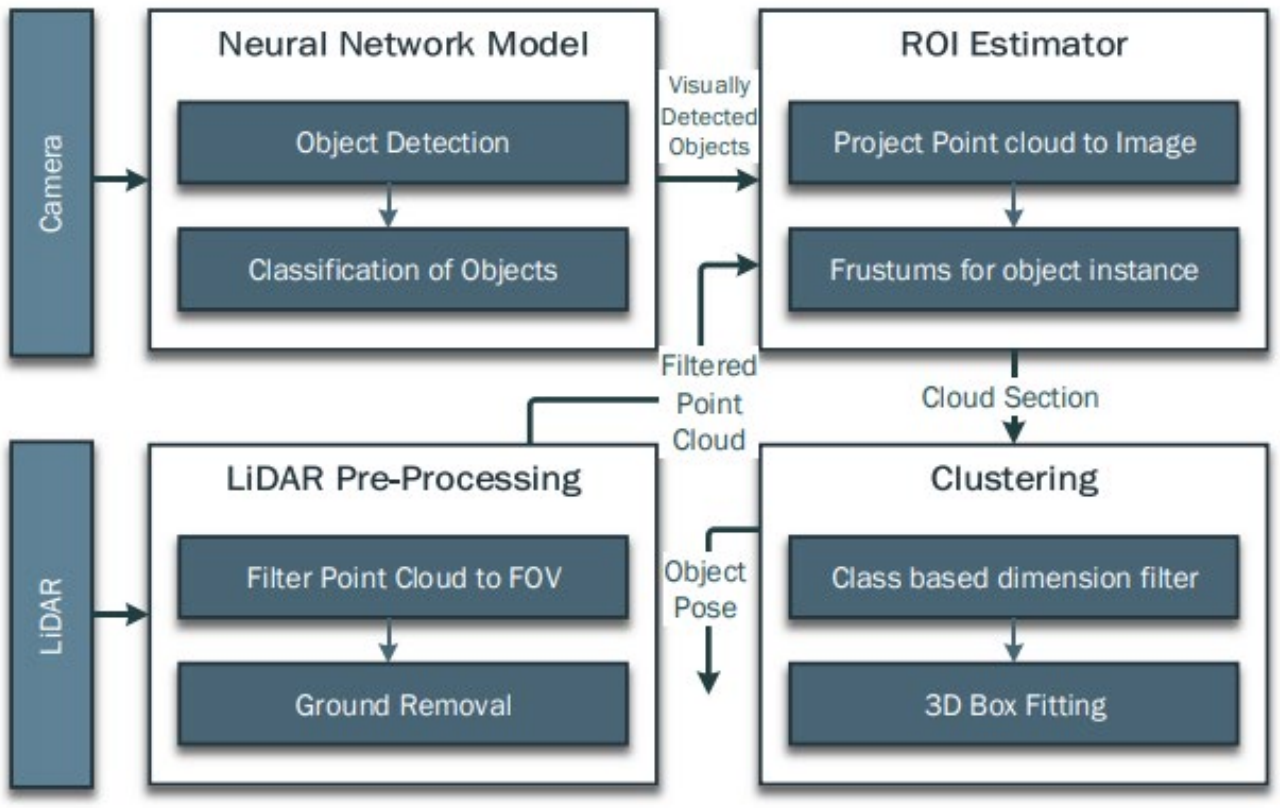

**Figure 2.1.** Camera and LiDAR based VRU detection [22]

After using visual data to perform object detection, point cloud data from LiDAR will be used as spatial information to further estimate the location and orientation of the VRUs. Figure 2.1 illustrates a multi-sensor fusion pipeline that combines camera and LiDAR data to detect and estimate the location and orientation of objects [22]. LiDAR Pre-Processing filters the point cloud to focus on the field of view (FOV) and removes ground points to improve object detection accuracy. The ROI Estimator then projects the detected objects onto the LiDAR point cloud to refine their spatial representation. The Clustering module further processes the filtered point cloud by applying class-based dimension filtering and 3D box fitting, which helps in estimating object poses. By integrating visually detected objects with LiDAR point cloud data, this approach enables precise localization and orientation estimation of objects in complex environments, enhancing perception for autonomous systems.

After estimation of the location and orientation of the VRUs, the position data will be used to further predict VRU trajectory and movement pattern. Spatio-Temporal Graph Transformer Network (STGFNet) framework is applied for pedestrian trajectory prediction [26]. The process starts with multi-pedestrian trajectory data input, containing absolute coordinates (temporal information) and relative coordinates (spatial interactions). The multi-modal preprocessing module separately processes the temporal sequence, which captures motion patterns over time, and the spatial sequence, which encodes pedestrian interactions. The encoder then extracts spatial-temporal features using a spatial encoder for pedestrian interactions at each timestep and a temporal encoder for motion pattern encoding across time. A memory mechanism is included to maintain consistency in long-term predictions by storing historical temporal features. Finally, the decoder utilizes STDecoder with masked attention layers to predict future trajectories based on the extracted high-dimensional spatial-temporal features. This model enhances trajectory prediction for vulnerable road users (VRUs) by effectively modeling their movement patterns and interactions in dynamic environments. This is crucial for autonomous driving and pedestrian/bicyclist safety applications

## 2.4 Conclusion

In this chapter, we explored various advanced methodologies for detecting VRUs using camera-based and LiDAR based technologies for autonomous driving. Additionally, we have found that LiDAR technology, in comparison to cameras, provides superior accuracy in locating the positions of road users due to its ability to generate precise 3D maps of the surrounding environment. Therefore, we conducted an extensive review of the latest studies in LiDAR technology and VRU trajectory prediction, which will undoubtedly enhance our foundation for future research in enhancing the safety and efficiency of autonomous driving systems.

# Chapter 3: Delay-Tolerant Path Tracking Control

## 3.1 Introduction

Our year 1 project report presented an effective path-tracking control design using the disturbance-observer (DOB). Information about the DOB method can be found in the reference [29-30]. One shortcoming of this method, however, is the significant deterioration of path-tracking performance when the system is subjected to unknown time delays. Hence, this chapter introduces a delay-tolerant communication disturbance observer (CDOB) design to handle path-tracking maneuvers even with undesirable time delays.

## 3.2 Path Generation

The collision avoidance maneuver in this case is assumed to be a single lane-change. The overall procedure of obtaining such a reference path remains identical to the approach used in the Year 1 report and can be condensed to the following: (a) generate limited number of sample waypoints to ascertain the general shape of the path; (b) generate dense waypoints based on the sample waypoints to complete the path waypoint design; (c) Apply segmentation to the dense waypoints to cut the path into several segments, ideally with each segment containing a minimum amount of features (i.e. corners); (d) Perform polynomial fit optimization to obtain the desired path expression that guarantees smooth curvature within each segment as well as smooth transition between the segments. The detailed step of this procedure is outlined in [31]. In Figure 3.1, an example of the optimized reference path and its path curvature is displayed, demonstrating the smoothness of such a path generated using this approach.

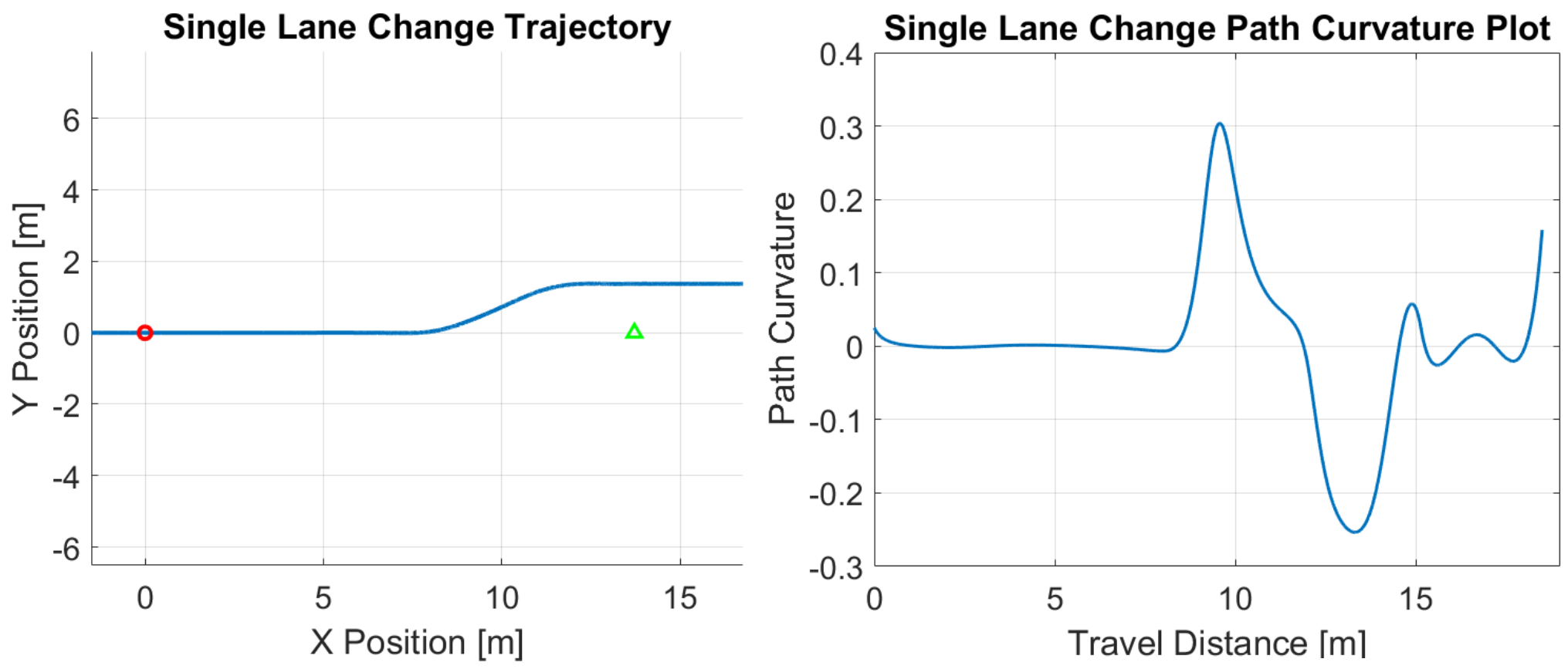

**Figure 3.1.** Reference path optimization: (a) optimized path; (b) path curvature of the optimized path

## 3.3 Linear Path-tracking Model

This section presents the linear path-tracking model that serves as the basis for the proposed control routine. The detailed derivation of this model can be found in [31]. It should also be noted that the same linear path-tracking model as also used in Year 1 project report's path-tracking control design section. Similar models that have been derived using a similar approach have also been applied to articulated vehicle configurations in the literature [32].

$$\begin{bmatrix} \dot{\beta} \\ \dot{r} \\ \Delta\dot{\psi}_p \\ \dot{e_y} \end{bmatrix} = \begin{bmatrix} \frac{-C_f-C_r}{MV} & -1+\frac{C_r l_r - C_f l_f}{MV^2} & 0 & 0 \\ \frac{C_r l_r - C_f l_f}{I_z} & \frac{-C_f {l_f}^2 - C_r {l_r}^2}{I_z V} & 0 & 0 \\ 0 & 1 & 0 & 0 \\ V & l_s & V & 0 \end{bmatrix} \begin{bmatrix} \beta \\ r \\ \Delta\psi_p \\ e_y \end{bmatrix} + \begin{bmatrix} \frac{C_f}{MV} & \frac{C_r}{MV} \\ \frac{C_f l_f}{I_z} & \frac{C_r l_r}{I_z} \\ 0 & 0 \\ 0 & 0 \end{bmatrix} \begin{bmatrix} \delta_f \\ \delta_r \end{bmatrix} + \begin{bmatrix} 0 \\ 0 \\ -V \\ -l_s V \end{bmatrix} \rho_{ref} + \begin{bmatrix} 0 \\ \frac{1}{I_z} \\ 0 \\ 0 \end{bmatrix} M_{zd} \quad (3.1)$$

where: $l_s = KV$, K is a constant

**Table 3.1.** Linear path-tracking model parameters.

| Model | Explanation |
|---|---|
| $\beta$ | Vehicle side slip angle |
| r | Vehicle yaw rate |
| $\Delta\psi_p$ | Heading error |
| $e_y$ | Path-tracking error |
| $C_f$ | Front tire cornering stiffness |
| $l_f$ | Distance between CG and front axle |
| $C_r$ | Rear tire cornering stiffness |
| $l_r$ | Distance between CG and rear axle |
| M | Vehicle mass |
| V | Vehicle velocity |
| $l_s$ | Preview distance |
| $I_z$ | Vehicle yaw moment of inertia |
| $\rho_{ref}$ | Reference path curvature |
| $M_{zd}$ | Yaw moment disturbance |
| K | Preview distance scheduling constant |

This linear path-tracking model contains two components: a (linear) lateral single-track model and a path-tracking model augmentation. The path-tracking scenario is illustrated in Figure 3.2, and the resulting linear path-tracking model is described in Equation 3.1. The parameters of this model are specified in Table 3.1. It can be observed that, for generality, the model presented in Equation 3.1 has both front and rear wheel steering angles $\delta_f$ and $\delta_r$ as inputs. In our case, the vehicle is assumed to be front-wheel-steer only. It can also be noticed that path curvature $\rho_{ref}$ and yaw moment disturbance $M_{zd}$ enter the model as external disturbances. Additionally, the preview distance $l_s$ is chosen to be a linear function of vehicle speed. It should also be remarked that vehicle speed can be scheduled according to the refence path curvature to make sure vehicle lateral acceleration stays within an acceptable limit.

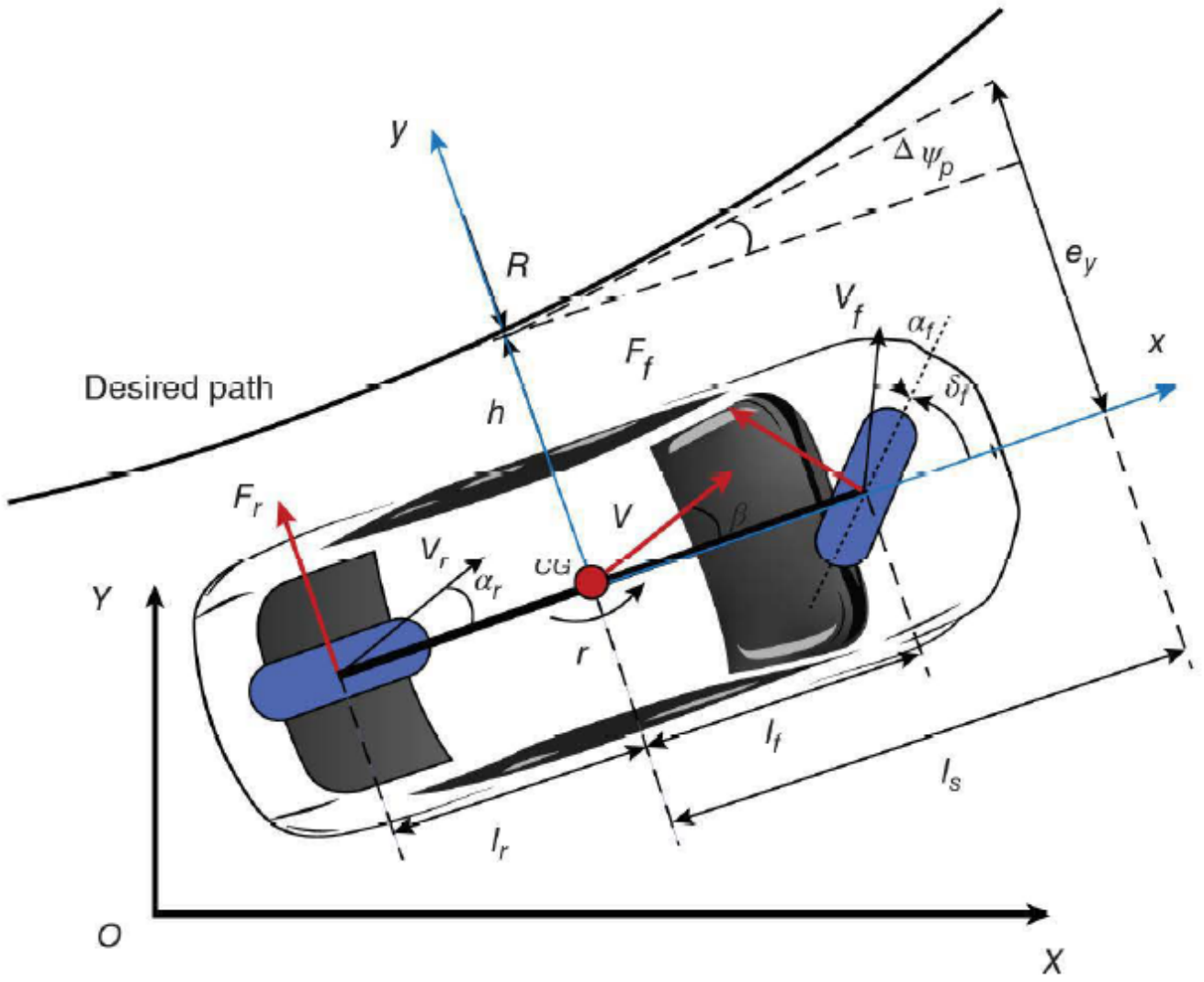

**Figure 3.2.** Path-tracking scenario [33]

## 3. 4 Delay-Tolerant Path-Tracking Control Design

### 3.4.1 Communication Disturbance Observer (CDOB)

This sub-section presents a general overview of the communication disturbance observer (CDOB), which is an approach inspired by the disturbance observer (DOB). Please see references [29], [34], [35] for more details on the CDOB.

Given a time-delayed input-output system as shown in Figure 3.3(a), an equivalent system can be constructed as displayed in Figure 3.3(b), where a term, $D(s)$, incorporates the time delay and is fed into the system as a disturbance. It must be remarked that the magnitude of the time delay is not necessarily known.

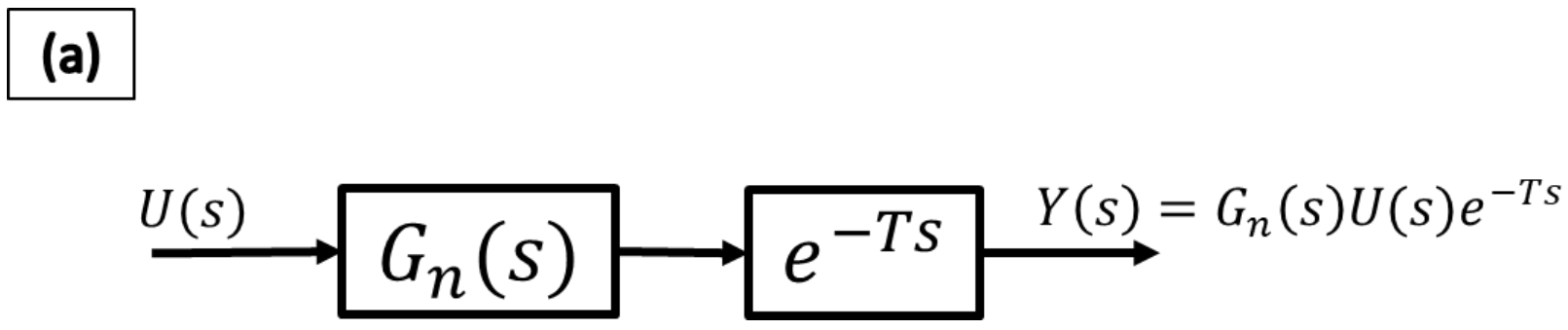

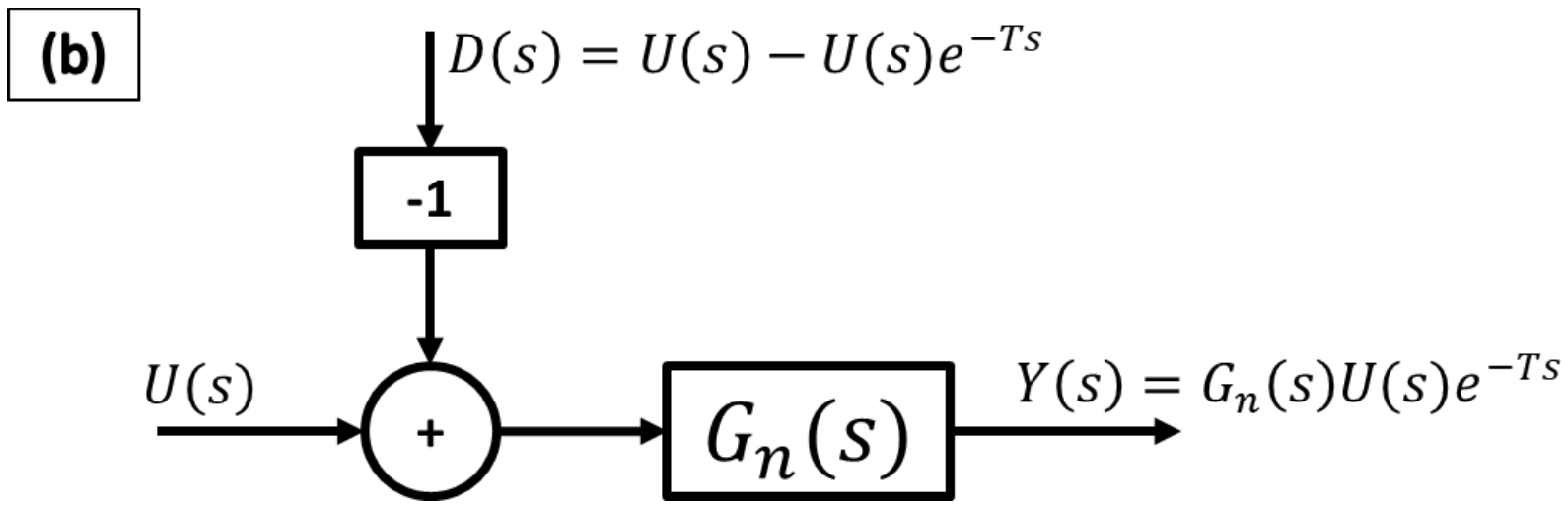

**Figure 3.3.** Sample input-output system with time delay

Once it has been established that the unknown time delay can be remodeled as a disturbance, the concept of DOB can be used to estimate and compensate for the time delay. Figure 3.4 shows the basic structure of the CDOB consisting of a time delay estimation loop and a time delay compensation loop. $Q(s)$ in the time delay estimation loop is a unity-gain low-pass filter of the appropriate order introduced to ensure that $Q(s)/G_n(s)$ is proper, hence ensuring that the scheme is implementable. Assuming that the analysis is carried out at low frequency where $Q(s) = 1$, it can be derived that the output of the time delay estimation loop yields $\bar{D}(s)$ which is an estimation of $D(s)$ as shown in Figure 3.3(b). The additional time delay compensation loop cancels out the term containing time delay and yields the desired output form $G_n(s)U(s)$ that is not affected by the unknown time delay.

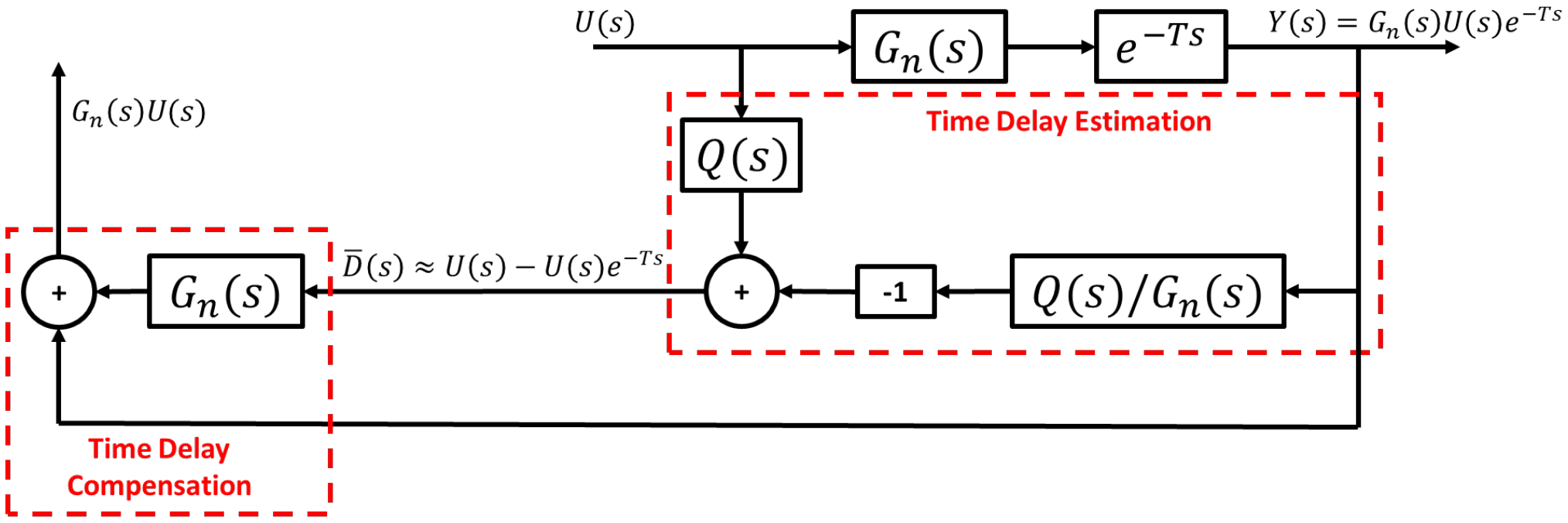

**Figure 3.4.** Standard CDOB block diagram

### 3.4.2 Modification for Path Curvature Rejection

As mentioned in Section 3.3, the vehicle model used for path-tracking is derived such that reference path curvature enters the model as an external disturbance. Denoting the path curvature disturbance as $d$, the desired output of the CDOB hence becomes $G_n(s)U(s) + d$. Adding this disturbance $d$ into the CDOB block diagram as illustrated in Figure 3.5, however, does not yield the desired outcome, where the actual output remains in the form of $G_n(s)U(s)$, lacking the disturbance term $d$.

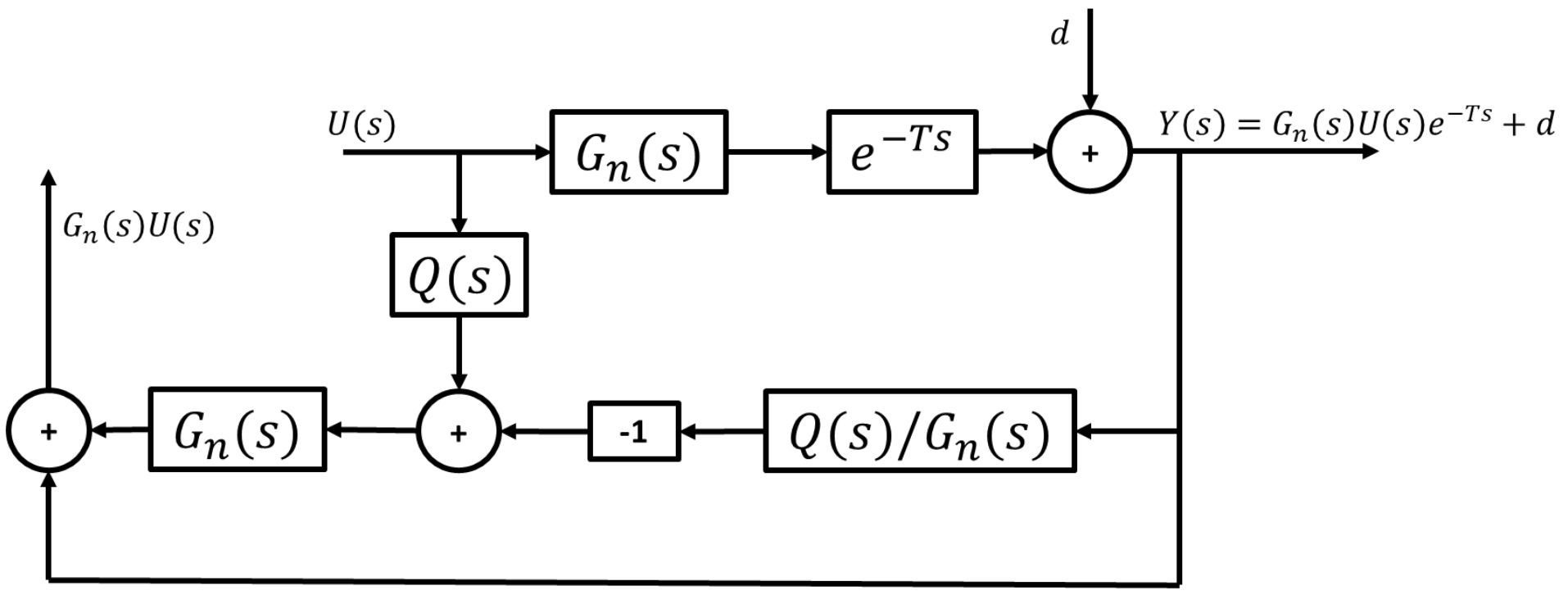

**Figure 3.5.** Standard CDOB block diagram with external disturbance

To account for the above issue, modifications must be made to the standard CDOB structure to accommodate the path curvature rejection requirement. Figure 3.6 shows the modified CDOB block diagram, where the same path curvature disturbance is added to the output of the CDOB delay compensation loop. It should be remarked that this structure works because the curvature of the reference path is known.

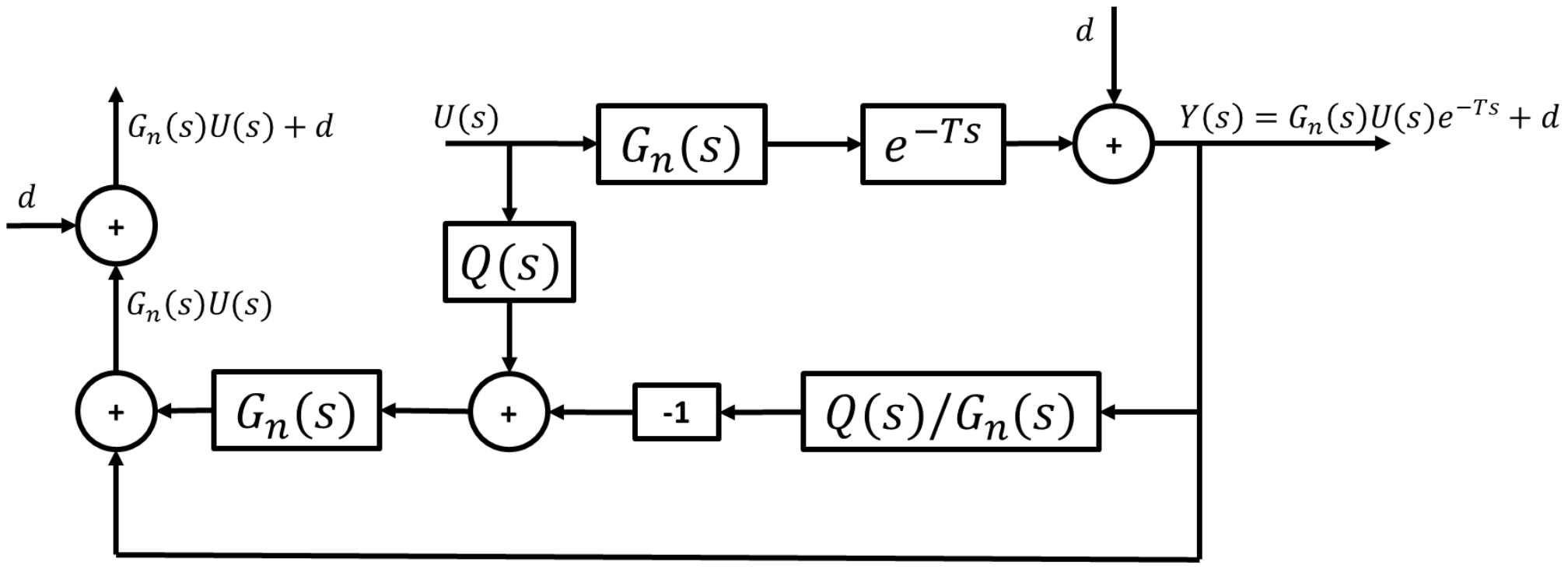

**Figure 3.6.** Modified CDOB block diagram with external disturbance

### 3.4.3 Feedback Controller Design

With the modified CDOB capable of outputting desired output form without the interference of time delay, closed-loop control system can be constructed for this non-time-delayed disturbance rejection problem. In Figure 3.7, a generic feedback controller $C(s)$ is added to generate input $U(s)$ such that reference input $R(s)$ can be tracked.

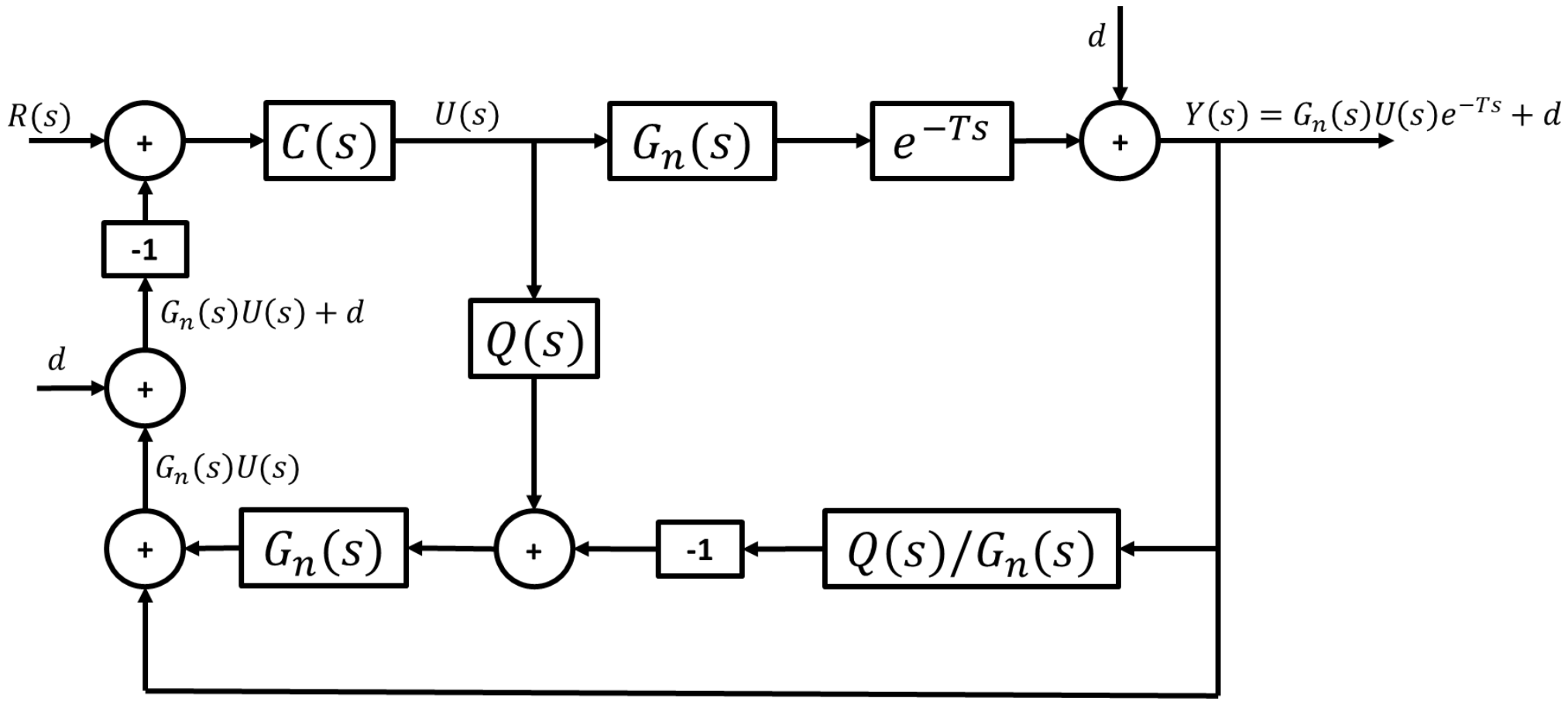

**Figure 3.7.** Control system with modified CDOB and feedback controller

The feedback controller mentioned above can be of any design. We present an example design that features a speed-scheduled, parameter-space PID controller, the details of which can be found in [31]. The parameter-space method is discussed in detail in [33] and also in references [36-38] that focus on application to autonomous path following. The reference input (denoted as $R(s)$ in Figure 3.7) should be zero in this case noting that the goal of the control system is to eliminate path-tracking error $e_y$. The form of the controller is presented in Equation 3.2. The controller gains $(k_p, k_i, k_d)$ are the parameters to be tuned. Since the controller is speed-scheduled in this case as well, the tunable parameter set has four elements: $(V, k_p, k_i, k_d)$. A D-stability region, as displayed in Figure 3.8, is established for pole placement. An example of the admissible controller gain region at a certain scheduled speed is shown in Figure 3.9. It should be remarked that during the process of controller gains value selection, a general rule of thumb is to choose the gains to be as small as possible within the admissible region so that the control efforts can be minimized while the energy efficiency maximized.

$$C(s) = k_p + \frac{k_i}{s} + k_d s \qquad (3.2)$$

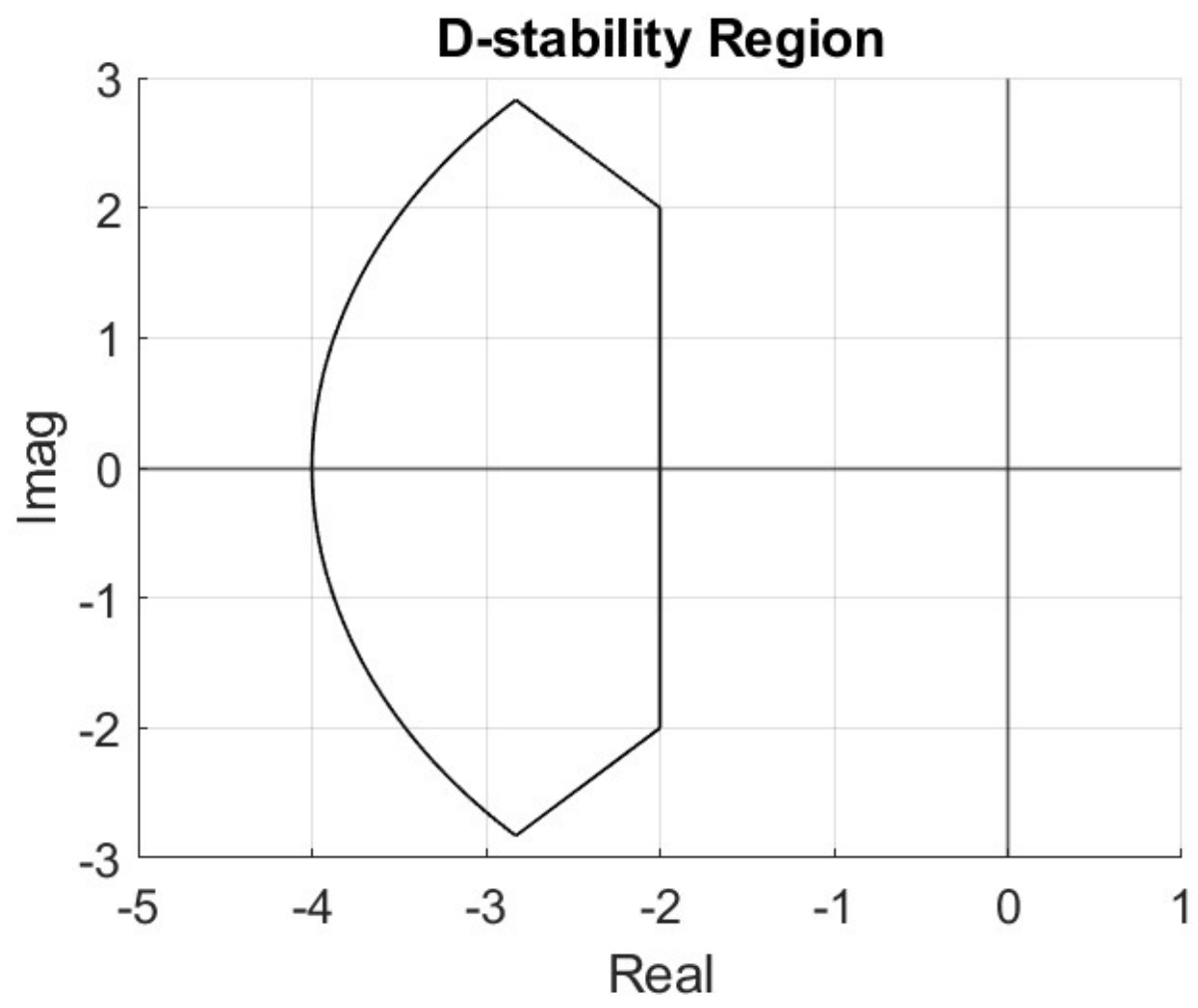

**Figure 3.8.** D-stability region

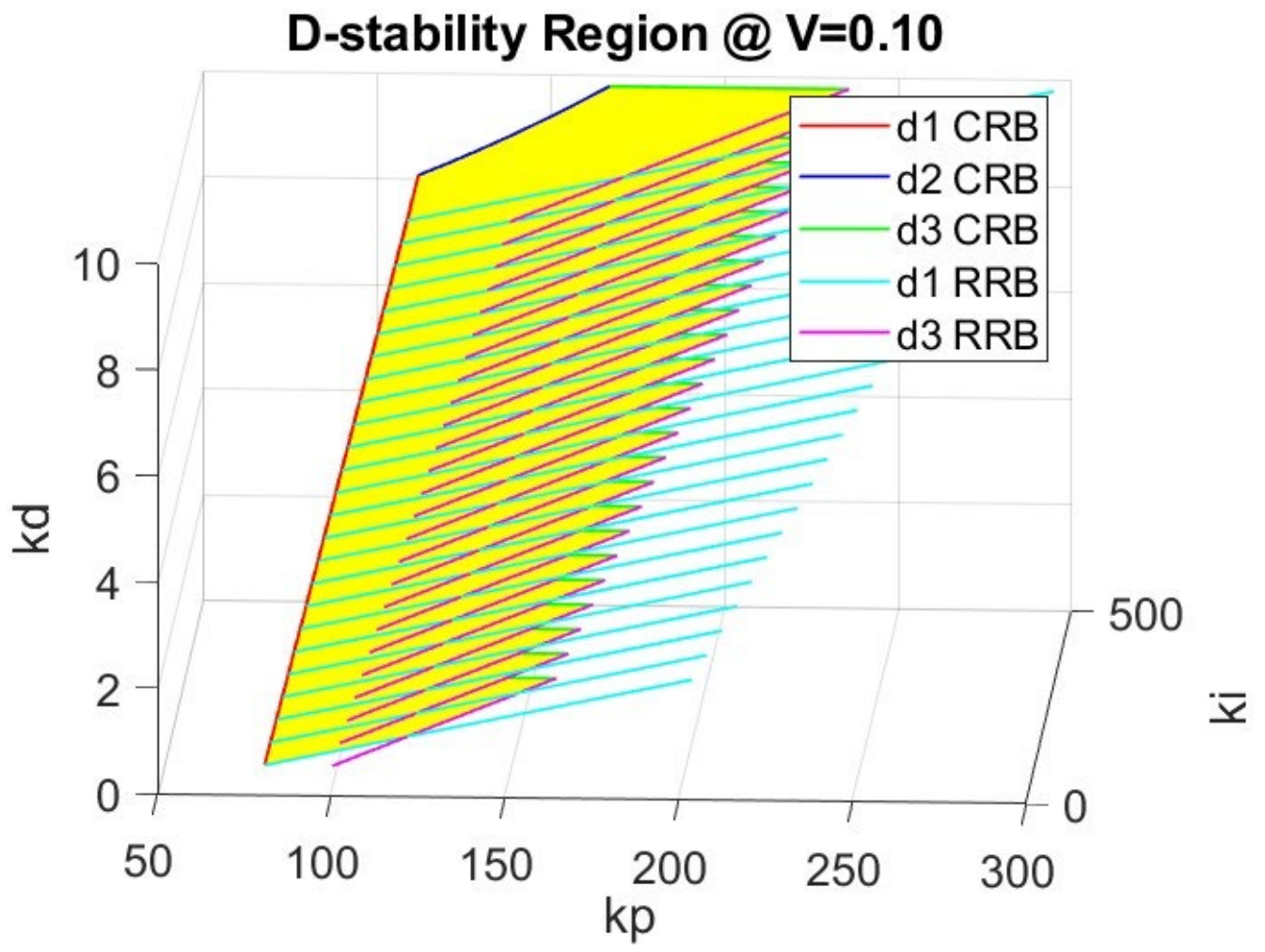

**Figure 3.9.** Admissible control region at a certain speed

# 3.5 Experiments

## 3.5.1 Simulation Study

Simulation studies are first performed to demonstrate the efficacy of the proposed control design. A Simulink model is constructed to simulate the motions of the vehicle. The parameter values used in the simulations are listed in Table 3.2. The example path shown in Figure 3.1 is used as the desired path in this experiment.

**Table 3.2.** Parameter value selections for simulation.

| Symbol | Parameter | Value |
|---|---|---|
| M | Mass | 3000 kg |
| $I_z$ | Yaw moment of inertia | 5.113e3 kg*m^2 |
| $C_f$ | Front tire cornering stiffness | 3e5 N/rad |
| $C_r$ | Rear tire cornering stiffness | 3e5 N/rad |
| $l_f$ | Distance between CG and front axle | 2 m |
| $l_r$ | Distance between CG and rear axle | 2 m |
| $t_d$ | Time Delay | 0.3 sec |

The forward motion simulation results for the combined modified CDOB and PID control system are displayed in Figure 3.10. It can be observed that the vehicle is able to track the reference path satisfactorily with reasonably small path-tracking errors by applying smooth steering inputs. To demonstrate the effectiveness of the CDOB component in the control system, a baseline result showing the trajectory of applying PID controller alone without the CDOB loop is included in Figure 3.10(a). It can be observed that with the same PID controller, the tracking operation will fail in very early stage without the CDOB feature.

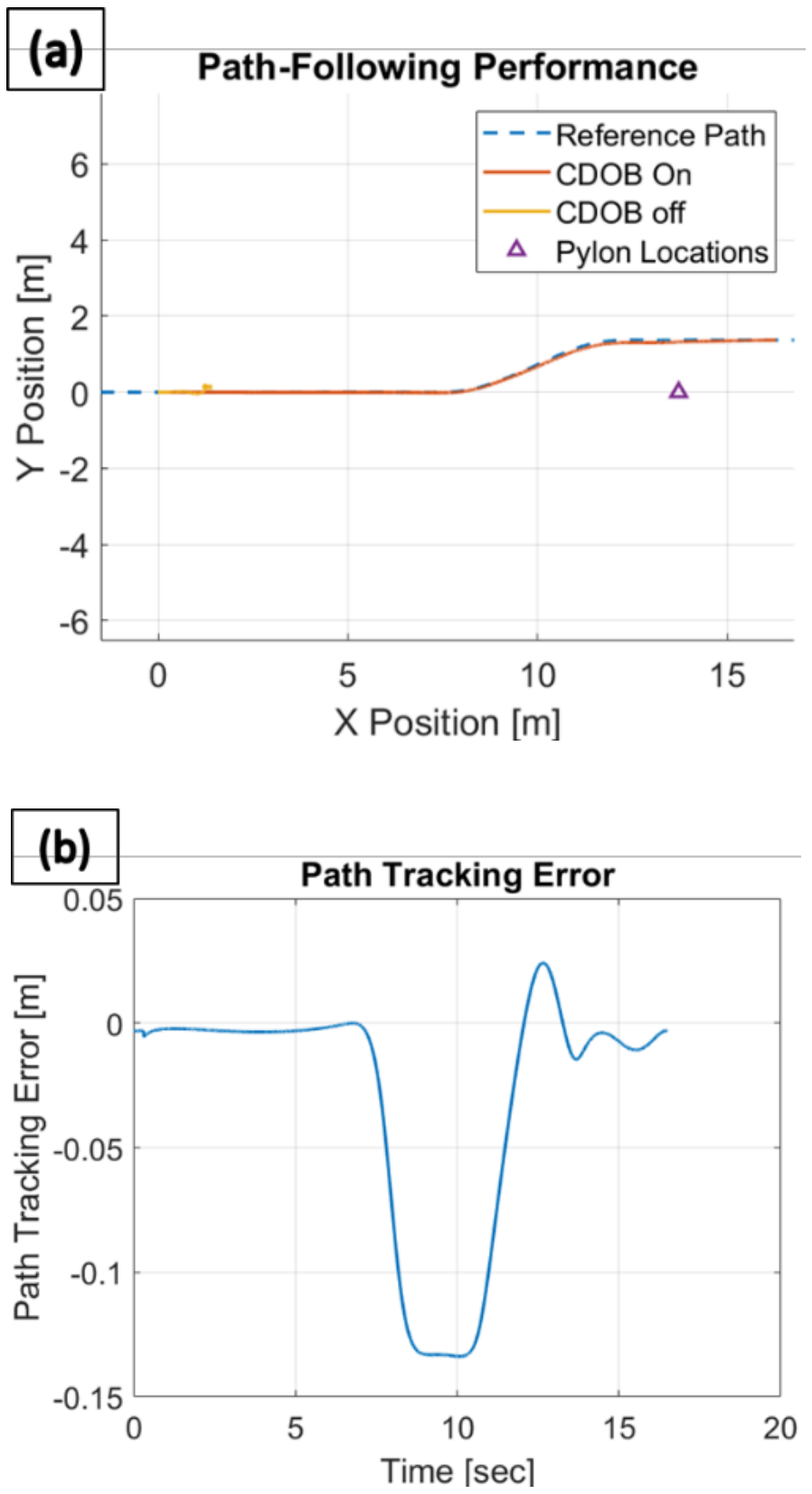

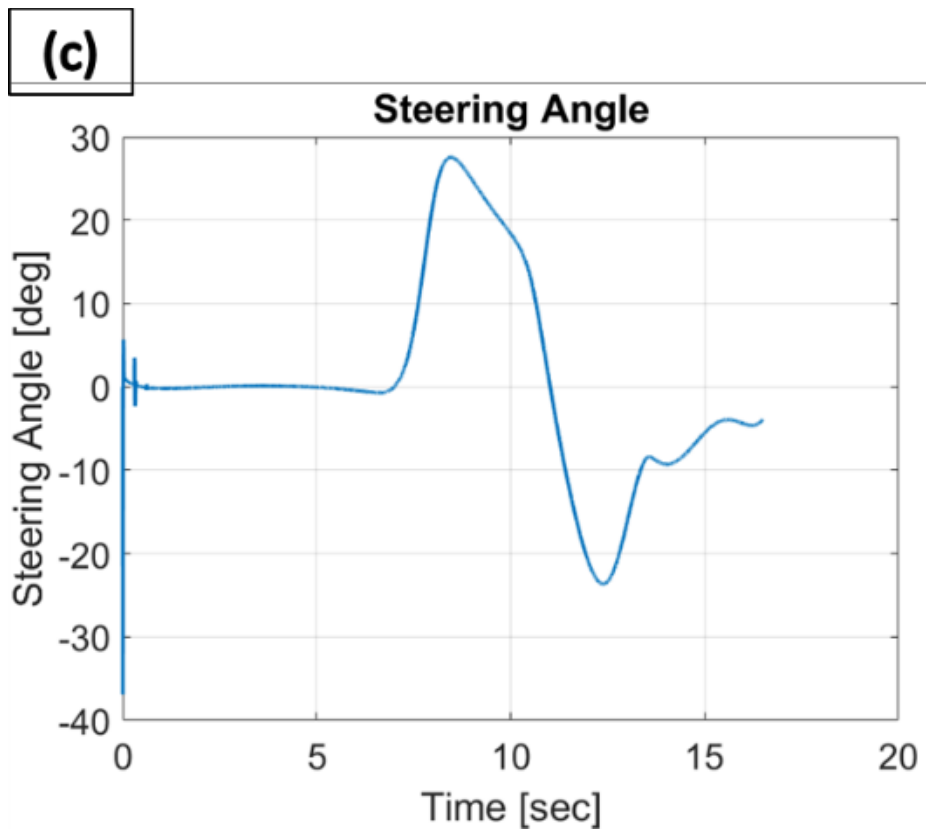

**Figure 3.10.** Forward motion modified CDOB + PID simulation results

### 3.5.2 Hardware-in-the-Loop (HIL) Experiment

In addition to the simulation study, hardware-in-the-loop (HIL) experiments are also performed to further demonstrate the suitability of this proposed control scheme for real-life implementations. The same parameter choices as shown in Table 3.2 as well as the same example path are used, and the results are shown in Figure 3.11. It can be observed that even for online operation in HIL simulator, the proposed control scheme is able to follow the desired path effectively with small path-tracking errors.

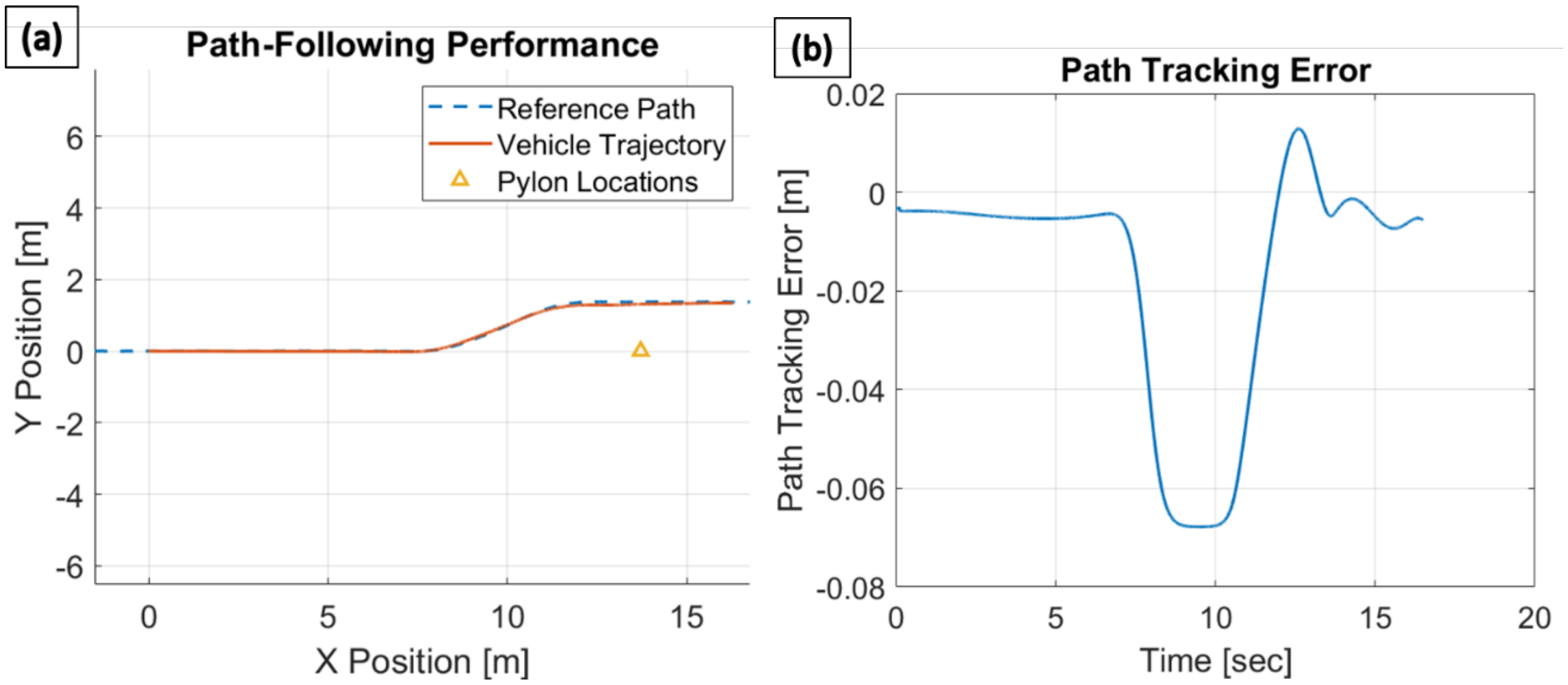

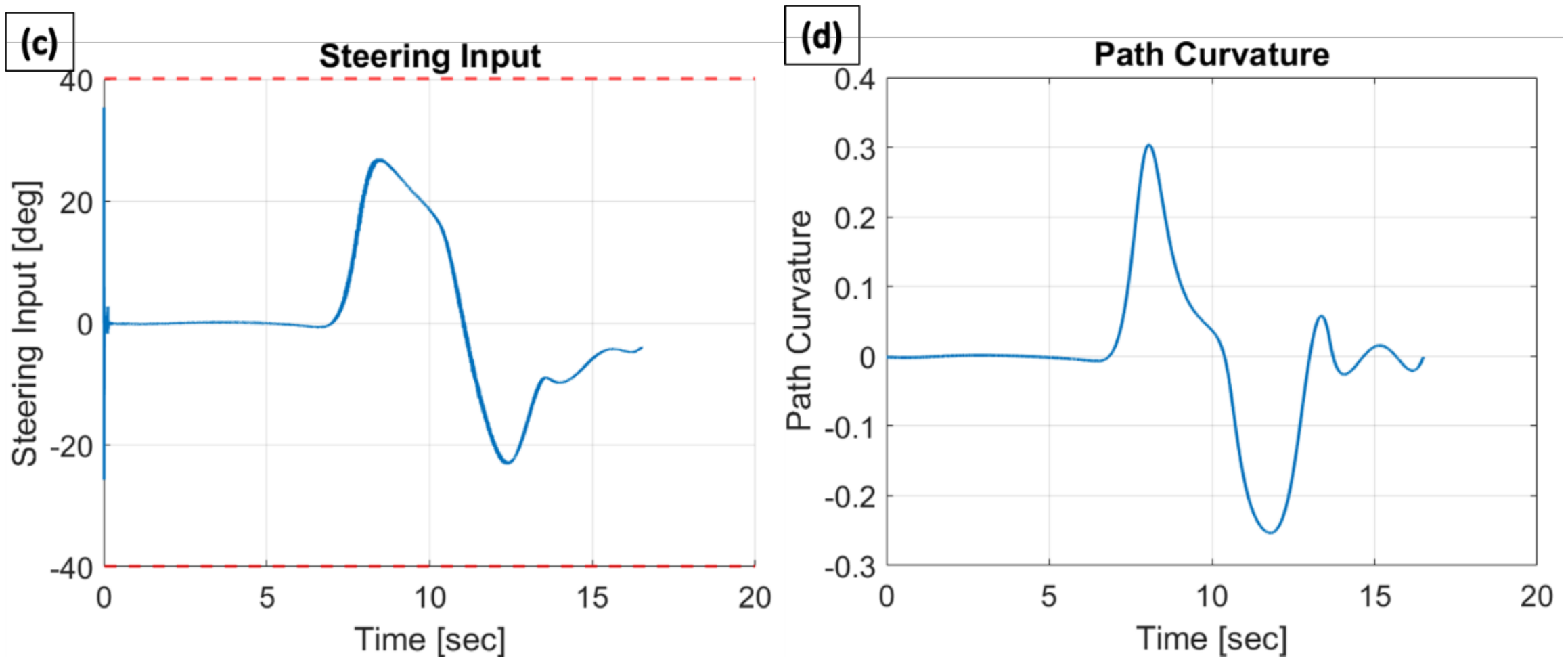

**Figure 3.11.** Forward motion modified CDOB + PID HIL results

# Chapter 4: Hierarchical Collision Avoidance Framework

## 4.1 Introduction

Autonomous driving is currently a highly popular research topic in the mobility area with immense potential to enhance safety, reduce traffic congestion, and revolutionize urban transportation [33], [39], [40], [41]. However, one of the major challenges in the development of autonomous vehicles is ensuring collision-free navigation in dynamic and unpredictable environments [33],[42],[43], [44]. This challenge becomes particularly significant in environments ranging from multi-lane highways with fast-moving traffic [45] to crowded urban settings where vehicles must navigate amidst pedestrians [46]. To address these challenges, it is essential to develop more robust and intelligent path planning and collision avoidance methods that can operate reliably in complex and uncertain environments. A well-designed planner can generate safe and feasible trajectories for autonomous vehicles in complex dynamic traffic environments.

The development of path planning and collision avoidance functions for connected automated vehicles is an intricate and complex process. Based on the specific vehicle type and traffic scenario, various planning algorithms and strategies are employed to plan a collision-free trajectory. Extensive research has been conducted to develop high-performance and robust collision-free path planning strategies, which can generally be categorized into two major approaches: optimization-based methods and machine learning-based methods.

The optimization-based approach formulates path planning and collision avoidance as a mathematical optimization problem with well-defined constraints, aiming to compute an optimal, collision-free trajectory by minimizing or maximizing specific objective functions. These constraints generally come from two primary sources: vehicle dynamic limitations (such as acceleration, steering angle, and braking capabilities) and traffic environment constraints (such as road boundaries and curvatures). Ensuring feasible and safe navigation requires optimizing trajectory planning within these constraints while maintaining computational efficiency for real-time applications.

Among various methods, the Control Lyapunov Function - Control Barrier Function - Quadratic Programming (CLF-CBF-QP) approach has gained significant attention due to its ability to balance safety and stability in an optimization framework. In this approach, CLFs enforce system stability, ensuring the vehicle follows a desired trajectory, while CBFs define safety boundaries, preventing collisions with traffic barriers and other road users. The optimal control input is then obtained by solving a QP problem, which ensures that both stability and safety constraints are satisfied [47-49]. This formulation allows for real-time control adaptation, making it particularly effective for dynamic traffic scenarios. The CLF-CBF-QP framework has been widely adopted in the fields of robotics [50-51] and autonomous driving [52-54], especially for vehicle control problems involving various system models, such as kinematic [55-56] and dynamic vehicle [57] representations. In addition to being used as a standalone controller, CBFs are also frequently integrated as safety filters alongside other control strategies, such as Deep Reinforcement Learning (DRL) and Model Predictive Control (MPC) [58-59], where they act as safety-check layers to ensure constraint satisfaction during control calculation. In addition, as system dynamics grow in complexity, High-Order CBF (HOCBF) formulations are introduced to allow CBF-based control to be applied to more general nonlinear systems [60-61].

Beyond CLF-CBF-based methods, various other optimization-based approaches have been explored to improve path planning and collision avoidance. These include the Elastic Band approach [62-63], the Potential Field approach [64], Support Vector Machines (SVM) based approach [65], geometry based optimization (quintic spline) [66], and hybrid A* search in spatiotemporal map [67].

The machine learning-based approach, on the other hand, frames autonomous driving and collision avoidance as a Markov Decision Process (MDP) problem and employs reinforcement learning (RL) methods to optimize the decision-making process. This approach has been widely applied in autonomous driving research [68-70]. Early work by Kendall et al. introduced an end-to-end deep reinforcement learning (DRL) framework for ADS [71], while Yurtsever et al. proposed a hybrid DRL system combining rule-based control with DRL-based algorithms [72]. Researchers have further extended DRL with innovations [73-75], such as short-horizon safety mechanisms for highway driving [76], dueling architectures for efficient learning [77], and hierarchical reinforcement learning (H-REIL) to balance safety and efficiency in near-accident scenarios [78]. Additionally, many DRL-based models have been trained using simulation platforms like CARLA and The Open Racing Car Simulator (TORCS) [79-80], demonstrating their effectiveness in various driving conditions [81].

While optimization-based methods offer robustness and reliability due to their mathematical rigor, they often lack real-time feasibility and struggle to scale in high-dimensional, dynamic environments due to computational complexity and limited adaptability. In contrast, learning-based approaches, especially DRL based methods, provide flexibility and data-driven decision-making, but they require extensive training data and extra hard-coded safety assurances to ensure reliability in real-world scenarios.

To address these limitations, researchers are increasingly exploring data-driven hybrid approaches that integrate optimization-based planning with machine learning and reinforcement learning techniques. A particularly promising direction is the integration of CLF-CBF-QP with DRL, which combines rule-based safety constraints with adaptive learning-based decision-making [39-40]. This fusion allows autonomous systems to leverage the robustness and stability of optimization-based methods while incorporating the real-time adaptability of learning-based approaches, significantly enhancing safety, efficiency, and decision-making capabilities in complex and dynamic driving environments.

In this section, we propose a novel approach that combines traditional optimization-based control design with DRL to develop a high-performance and robust control strategy for autonomous vehicles. We implement it on two distinct vehicle models to demonstrate its effectiveness. The contributions of these sections are as follows.

- This section introduces a novel low-level controller based on the CLF-CBF-QP framework. For complex vehicle models, HOCLF and HOCBF are applied. This controller enables accurate path tracking under normal conditions and ensures effective collision avoidance in the presence of obstacles.

- Building on this foundation, we integrate the low-level CLF-CBF-QP based control with a high-level DRL decision making algorithm for path planning. The DRL algorithm generates high-level decisions based on surrounding traffic information, which the proposed optimization-based control then refines and executes.

- This hierarchical architecture takes advantage of both the learning-based approach and the optimization-based approach. The DRL-based high-level planner enables flexible decision-making in various traffic scenarios, while the optimization-based low-level controller adds hard safety constraints through CBFs, preventing unsafe behavior.

## 4.2 Methodology

### 4.2.1 Vehicle model

To comprehensively evaluate the proposed hierarchical collision avoidance framework, we implement it on two distinct vehicle models. Initial feasibility is verified using a simplified unicycle model, followed by a deployment on a more realistic vehicle dynamics model, where its effectiveness is further tested in a complex multi-lane driving environment.

#### 4.2.1.1 Unicycle vehicle dynamics

Indeed, the unicycle model is widely adopted in the field of mobile robotics due to its simplicity. To achieve more realistic vehicle dynamic simulations, incorporating more complex vehicle models would be necessary. However, the use of advanced vehicle models introduces significant implementation challenges for the CLF-CBF approach and increases computational complexity. To balance real-time performance with realistic simulation, the unicycle model is used first as proof-of-concept. As the simplest model, it provides a reasonable approximation of vehicle dynamics while maintaining computational efficiency. This choice aligns with many related studies, which also adopt the unicycle model to simplify the design and analysis of controllers while retaining sufficient representational fidelity for practical applications.

Figure 4.1 illustrates the plane motion of this unicycle vehicle. The vehicle can move forward with various linear speed $v$ and rotate with various angular speed $\omega$ around its geometry center. Note that this is also the Dubins model of a vehicle if we fix the speed and limit the rotation angle.

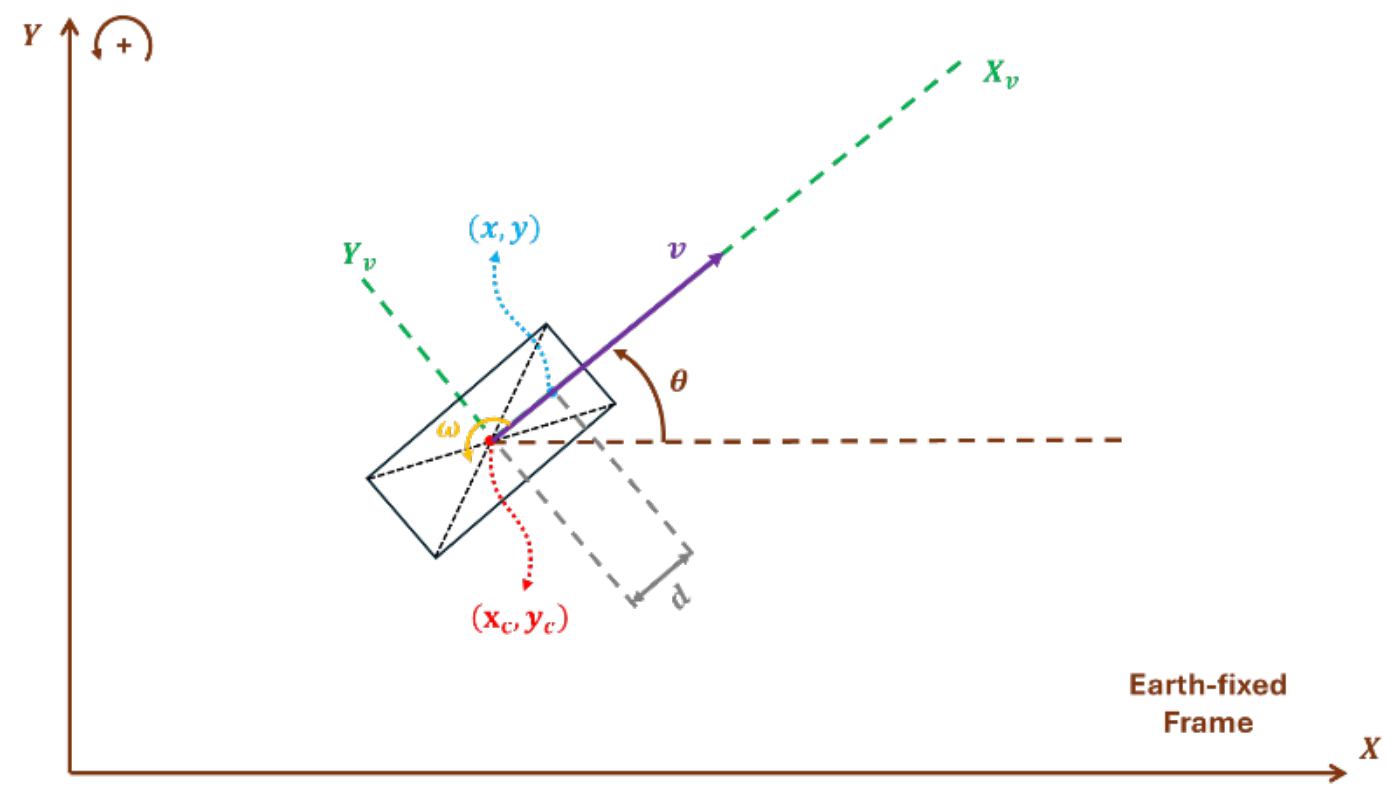

**Figure 4.1.** Unicycle vehicle dynamic model.

The state-space equation for this vehicle model is

$$\begin{bmatrix} \dot{x}_c \\ \dot{y}_c \\ \dot{\theta} \end{bmatrix} = \begin{bmatrix} \cos(\theta) & 0 \\ \sin(\theta) & 0 \\ 0 & 1 \end{bmatrix} \begin{bmatrix} v \\ \omega \end{bmatrix}, \tag{4.1}$$

where $[x_c \quad y_c]^T$ are the geometric center coordinates of the vehicle and $\theta$ denates the orientation of the vehicle. Now, consider another point $[x \quad y]^T$ located at a different position along the longitudinal axis $x_v$ of the vehicle. This point is located at an offset distance $d$ along the vehicle's x-axis relative to the geometric center. The relationship between point $[x \quad y]^T$ and vehicle's geometry center $[x_c \quad y_c]^T$ can be represented using equations below.

$$\begin{bmatrix} x \\ y \end{bmatrix} = \begin{bmatrix} x_c \\ y_c \end{bmatrix} + \begin{bmatrix} \cos(\theta) & -\sin(\theta) \\ \sin(\theta) & \cos(\theta) \end{bmatrix} \begin{bmatrix} d \\ 0 \end{bmatrix} \tag{4.2}$$

The state-space equation for the geometric center of the vehicle is given by

$$\begin{bmatrix} \dot{x} \\ \dot{y} \\ \dot{\theta} \end{bmatrix} = \begin{bmatrix} \cos(\theta) & -d\sin(\theta) \\ \sin(\theta) & d\cos(\theta) \\ 0 & 1 \end{bmatrix} \begin{bmatrix} v \\ \omega \end{bmatrix}. \tag{4.3}$$

Both state-space equations (4.1) and (4.3) are driftless nonlinear systems which rely entirely on control inputs to define their motion. They are particularly useful for autonomous driving research, as they simplify the analysis and synthesis of control laws without the need to counteract drift dynamics. Moreover, the simplicity of the unicycle model largely simplifies the design of Control Lyapunov and Control Barrier Functions by streamlining the calculations.

### 4.2.1.2 Single-track lateral vehicle dynamic model

To comprehensively evaluate the effectiveness and robustness of the proposed control framework in practical driving scenarios, it is essential to integrate more advanced vehicle models. Figure 4.2 illustrates the plane lateral motion of this single-track vehicle, which serves as the foundation for the subsequent controller design. The vehicle is moving forward with a constant speed. For realistic simulation, the wheel side slip angle (the angle between the direction of the wheel's travel and the actual path) is considered in this model. The complete derivation of this vehicle model can be found in Chapter 2 of [33]. Equation (4.4) demonstrates the state space equation of this vehicle lateral dynamics where $\beta$ and $r$ represent vehicle side slip angle and vehicle yaw rate, respectively, and form the state of the model. $\delta_f$ and $\delta_r$ are vehicle front and rear steering angles and are the inputs of the model, $M_{zd}$ is yaw moment disturbance which serves as the external disturbance. Equations (4.5) and (4.6) are used to calculate the vehicle's position based on $\beta$ and $r$. $\psi$ is the vehicle yaw angle. Table 4.1 provides a detailed explanation of the parameters in the vehicle model. The vehicle model parameters used are taken from reference [41] which has applied deep reinforcement learning control to safety of vulnerable road users in their interaction with autonomous vehicles.

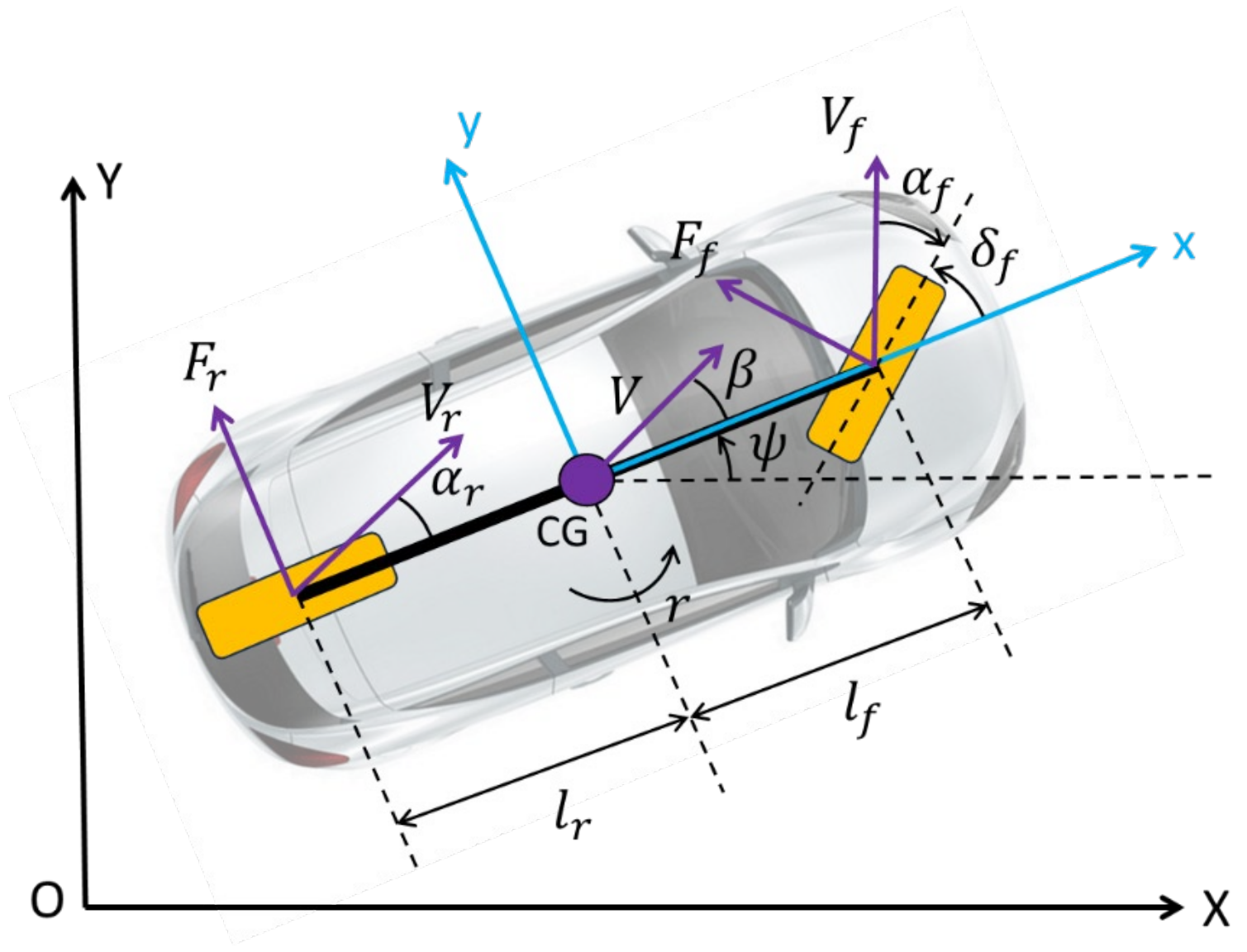

**Figure 4.2.** Linear single-track lateral vehicle dynamic model.

$$\begin{bmatrix} \dot{\beta} \\ \dot{r} \end{bmatrix} = \begin{bmatrix} \frac{-C_f - C_r}{mV} & -1 + \frac{C_r l_r - C_f l_f}{mV^2} \\ \frac{C_r l_r - C_f l_f}{I_z} & \frac{-C_f {l_f}^2 - C_r {l_r}^2}{I_z V} \end{bmatrix} \begin{bmatrix} \beta \\ r \end{bmatrix} + \begin{bmatrix} \frac{C_f}{mV} & \frac{C_r}{mV} \\ \frac{C_f l_f}{I_z} & \frac{C_r l_r}{I_z} \end{bmatrix} \begin{bmatrix} \delta_f \\ \delta_r \end{bmatrix} + \begin{bmatrix} 0 \\ \frac{1}{I_z} \end{bmatrix} M_{zd} \tag{4.4}$$

$$\Delta x = \int_0^{t_f} v cos(\beta + \psi)\, dt \tag{4.5}$$

$$\Delta y = \int_0^{t_f} v sin(\beta + \psi)\, dt \tag{4.6}$$

**Table 4.1.** Lateral model parameters [84].

| Symbol | Parameter |
|---|---|
| $X, Y$ | Earth-fixed frame coordinate |
| $x, y$ | Vehicle-fixed frame coordinate |
| $V$ | Center-of-gravity (CG) velocity |
| m | Mass |
| $I_z$ | Yaw moment of inertia |
| $\beta$ | Side-slip angle |
| $\psi$ | Yaw angle |
| r | Yaw rate |
| $M_{zd}$ | Yaw disturbance moment |
| $\delta_f, \delta_r$ | Front & rear wheel steer angle |
| $\alpha_f, \alpha_r$ | Front & rear tire slip angle |
| $C_f, C_r$ | Front & rear tire cornering stiffness |
| $l_f, l_r$ | Distance between CG and front & rear axle |
| $V_f, V_r$ | Front & rear axle velocity |
| $F_f, F_r$ | Front & rear lateral tire force |

By assuming a front-wheel-steering vehicle ( $\delta_r = 0$ ) and neglecting yaw moment disturbances ($M_{zd} = 0$), Equations (4.4) – (4.6) can be combined to construct a five-degree-of-freedom (5-DOF) lateral vehicle dynamic model for simulation purposes. The state-space equation for the proposed 5-DOF vehicle lateral dynamic model is given by

$$\begin{bmatrix} \dot{\beta} \\ \dot{r} \\ \dot{x} \\ \dot{y} \\ \dot{\psi} \end{bmatrix} = \begin{bmatrix} A_{11} * \beta + A_{12} * r \\ A_{21} * \beta + A_{22} * r \\ v * cos\,(\beta + \psi) \\ v * sin\,(\beta + \psi) \\ r \end{bmatrix} + \begin{bmatrix} B_1 \\ B_2 \\ 0 \\ 0 \\ 0 \end{bmatrix} \delta_f. \tag{4.7}$$

where $A_{11} = \frac{-C_f - C_r}{mv}$, $A_{12} = -1 + \left(\frac{C_r l_r - C_f l_f}{mv^2}\right)$, $A_{21} = \frac{C_r l_r - C_f l_f}{I_z}$, $A_{22} = \frac{-C_f l_f^2 - C_r l_r^2}{I_z V}$, $B_1 = \frac{C_f}{mv}$ and $B_2 = \frac{C_f l_f}{I_z}$. Compared to simplified lateral vehicle dynamic models like Equation (4.1) that only consider internal vehicle states such as sideslip angle and yaw rate, the proposed model further incorporates the vehicle's position and orientation $(x, y, \psi)$. This augmentation enables a clearer geometric interpretation of the vehicle state, which is particularly beneficial for the subsequent design of HOCLF-HOCBF-based controllers where reference tracking and obstacle avoidance can be formulated in the global coordinate frame.

### 4.2.2 Low-level CLF-CBF-QP Based Controller Design

#### 4.2.2.1 CLF and CBF Definition

The CLF is usually used as the constraint of an optimization problem to ensure the stability of dynamic systems by defining a scalar function that decreases over time as the system evolves. By encoding stability criteria into a function, the CLF allows the system to converge to the desired state with reasonable speed despite dynamic and environmental uncertainties. In autonomous driving, a CLF is employed to design a path-tracking controller that guides the vehicle towards the desired target position. This approach is robust in real-time applications where stability must be maintained even in the presence of disturbances, ensuring that the autonomous vehicle operates reliably and efficiently. Through CLF-based control strategies, vehicles can achieve precise trajectory tracking even in complex driving scenarios, which makes it a foundational element in advanced autonomous driving algorithms.

Let us first introduce the basic principle of CLF. Consider the control affine system given by

$$\dot{x} = f(x) + g(x)u \tag{4.8}$$

where $x$ is the state vector, $u$ is the control input vector, $f$ and $g$ are Lipschitz continuous in $x$. $f$ is the uncontrolled part, which represents the system's natural dynamics, while $g$ is the control distribution matrix which determines how the control inputs influence the system's dynamics. Let $V(x)$: $R^n \rightarrow R$ be a continuously differentiable function, a CLF $V(x)$ satisfies:

(1) Positive definiteness:

$$V(x) > 0 \text{ for all } x \neq x_e, \text{ and } V(x_e) = 0 \tag{4.9}$$

where $x_e$ is the equilibrium point.

(2) Sublevel set boundedness: For a given constant $c > 0$, the sublevel set, $\Omega_c = \{x \epsilon R^n: V(x) \leq c\}$ is bounded. This ensures that $V(x)$ defines a meaningful region of attraction (ROA) around $x_e$.

(3) Stability: There exists a control input $u \epsilon R^n$ such that the derivative of $V(x)$ along the trajectory of the system satisfies:

$$\min_{u \epsilon U} \dot{V}(x, u) = \min_{u \epsilon U} [\nabla V(x) \cdot (f(x) + g(x)u)] < 0, \quad \forall x \epsilon \Omega_c \backslash \{x_e\} \tag{4.10}$$

The Control Barrier Function (CBF) is usually used as a constraint of an optimization problem to ensure the safety of dynamic systems. A CBF defines a safe set which is a region in the state space where the system can operate without violating safety conditions. By incorporating CBF constraints into optimization-based controllers, the controller ensures that the system remains within the defined safe set while allowing flexibility for other objectives such as stability or performance. In the context of autonomous driving, CBFs are usually employed to guarantee collision avoidance, maintain lane adherence, and respect speed limits. In this section, we use CBF to enforce minimum distance constraints between the autonomous vehicle and obstacles, ensuring safe operation in dynamic environments. By integrating CBFs with other constraints such as CLF, optimization-based control strategy can be designed for autonomous vehicles. This approach enables simultaneous achievement of safety, efficient path tracking, and effective collision avoidance.

For CBF, let $h(x): R^n \rightarrow R$ be a continuously differentiable function that defines the safe set

$$C = \{x \in R^n: h(x) \geq 0\} \tag{4.11}$$

where $h(x) \geq 0$ represents the safe region, and $h(x) < 0$ represents the unsafe region. The function $h(x)$ is considered a CBF if there exists a control input $u \in R^n$ such that the following condition holds for all $x \in C$.

$$\sup_{u \in U}[\frac{\partial h(x)}{\partial x} \cdot ((f(x) + g(x)u)] \geq -\alpha(h(x)) \tag{4.12}$$

where $\alpha$ is a class-$\kappa$ function, which specifies the rate at which the system can approach the boundary of the safe set.

#### 4.2.2.2 HOCLF and HOCBF Definition

For complex systems where safety constraints depend on higher-order derivatives of the position states, standard CLFs and CBFs become insufficient. Therefore, we need to introduce HOCLF and HOCBF. The relative degree of HOCLF and HOCBF is the number of times we need to differentiate it along the dynamics of the system until the control input $u$ explicitly shows.

The definition of the HOCLF is presented as follows. Consider a $d$th-order continuously differentiable function $V(x): R^n \rightarrow R$. We let $\phi_0(x) = V(x)$ and a sequence of functions $\phi_i(x): R^n \rightarrow R, i \in \{1, \dots, d\}$ :

$$\phi_i(x) = \dot{\phi}_{i-1}(x) + \alpha_i(\phi_{i-1}(x)), i \in \{1, \dots, d\} \tag{4.13}$$

where $\alpha_i$, $i \in \{1, \dots, d\}$ are class-$\kappa$ functions. If there exist $\alpha_d$ such that for $\forall x \neq 0_n$,

$$\inf_{u \epsilon U}[L_f^d V(x) + L_g L_f^{d-1} V(x) u + S(h(x)) + \alpha_d(\phi_{d-1}(x))] \leq 0 \tag{4.14}$$

where $L_f$ and $L_g$ denote Lie derivatives along $f(x)$ and $g(x)$, then, $V(x)$ is a HOCLF for the system which can guarantee global and exponential stabilization.

Similarly, the definition of HOCBF is presented as follows. Consider an $r$th-order continuously differentiable function $h(x): R^n \to R$. We let $\Psi_0(x) = h(x)$ and a sequence of functions $\Psi_i(x): R^n \to R$, $i \in \{1, \dots, r\}$:

$$\Psi_i(x) = \dot{\Psi}_{i-1}(x) + \beta_i(\Psi_{i-1}(x)), i \in \{1, \dots, r\} \tag{4.15}$$

where $\beta_i$, $i \in \{1, \dots, r\}$ are class-$\kappa$ functions. We also define a sequence of sets $C_i$, $i \in \{1, \dots, r\}$:

$$\mathrm{C}_i(x) = \{x \in \mathbb{R}^n : \Psi_{i-1}(x) \geq 0\}, i \in \{1, \dots, r\} \tag{4.16}$$

If there exists $\beta_r$ and a control input $u \epsilon R^n$ such that the following condition holds for $\forall x \epsilon C_1 \cap C_2 \cap \dots \cap C_i$

$$\sup_{u \epsilon U}\left[L_f^m h(x) + L_g L_f^{m-1} h(x) u + S(h(x)) + \alpha_i(\Psi_{i-1}(x))\right] \geq 0 \tag{4.17}$$

where $L_f$ and $L_g$ denote Lie derivatives along $f(x)$ and $g(x)$, then the function $h(x)$ is considered a HOCBF for the system which can guarantee safety. The detailed derivation and proof of HOCLF and HOCBF can be found in [60], [85].

#### 4.2.2.3 CLF-CBF-QP Controller Design for Unicycle Model

For the unicycle model, the CLF is applied to design the path-tracking controller, ensuring the vehicle accurately follows a predefined trajectory. Equations (4.18) and (4.19) demonstrate the design of a Lyapunov function and its derivative where $e$ is the path tracking error. The path tracking error can be calculated using the position of the vehicle and the position of the tracking point on the path, $e = p - p_d(\gamma)$, where $p_d(\gamma): \mathbb{R} \to \mathbb{R}^2$ represents the planar parameterized path, generated by $B$-Spline fitting for example, and $\gamma \in \mathbb{R}$ is a time dependent parameter for position along the path. Equation (4.20) demonstrates the dynamics of the path where desired path speed is $\gamma_d$ and $g(e)$ is used to slow down the tracking point when the path tracking error is too large. The CLF incorporates path-tracking error, quantifying the deviation of the vehicle from the desired path. By integrating the CLF as a constraint within the quadratic programming (QP) optimization problem, the controller ensures that the system minimizes the path-tracking error at every time step. This formulation guarantees stability by driving the CLF to decrease over time, ultimately forcing the vehicle to converge to the predefined path. The detailed design of CLF is illustrated in Equation (4.21) where $\epsilon$ is the relaxing term which indicates that the vehicle can temporarily deviate from the path when necessary.

$$V(e) = \frac{1}{2}\|e\|^2 \tag{4.18}$$

$$\dot{V}(e) = e^T(\begin{bmatrix} \cos(\theta) & -d * \sin(\theta) \\ \sin(\theta) & d * \cos(\theta) \end{bmatrix} u - \frac{\partial p_d}{\partial \gamma}\dot{\gamma}) \tag{4.19}$$

$$\dot{\gamma} = \gamma_d + g(e) \tag{4.20}$$

$$e^T\left(\begin{bmatrix} \cos(\theta) & -d * \sin(\theta) \\ \sin(\theta) & d * \cos(\theta) \end{bmatrix} u - \frac{\partial p_d}{\partial \gamma}\dot{\gamma}\right) + \frac{\alpha}{2}\|e\|^2 \leq \epsilon \tag{4.21}$$

For the unicycle model, the CBF is utilized to design the collision avoidance controller, ensuring the vehicle's safety in the presence of nearby obstacles. To simplify the problem and enable effective mathematical formulation, we assume that both vehicle and obstacles have elliptical geometries. This assumption allows the use of an ellipse-based barrier function which defines a safe region by ensuring that the vehicle maintains an appropriate distance from obstacles. By assuming that both the vehicle and the obstacles have elliptical geometries, we imply that they can be approximated or enclosed by one or more elliptical shapes, depending on the complexity of their structures. Once these elliptical boundaries are established, elliptical CBF constraints can be applied, with each boundary corresponding to a specific CBF constraint, to formulate the QP problem. This approach enables efficient calculation of optimal control.

The primary purpose of utilizing CBFs is to ensure collision avoidance and maintain safety during path tracking. In our simulation studies, both the vehicle and obstacles are encircled with elliptical boundaries. CBFs are employed to guarantee that the vehicle's boundary does not overlap with the obstacle boundaries, effectively preventing collisions. If there is more than one obstacle, each obstacle/boundary corresponds to a specific CBF constraint. While it is possible to impose input constraints by introducing additional constraints into the QP formulation, this approach significantly increases the problem's complexity and computational time. To address this, our method solves the CLF-CBF-QP problem without input constraints and applies saturation directly during the execution of the calculated optimal input. This approach maintains computational efficiency while ensuring practical feasibility.

Equation (4.22) is the barrier function of the elliptical region with $H(\theta) = R(\theta)\Lambda R(\theta)^T$ and $\Lambda = \mathrm{diag}\{1/\mathrm{a}^2,\ 1/\mathrm{b}^2\}$. Where $R(\theta)$ is the 2D rotational matrix. Constant $a$ and $b$ are longest and shortest radiii of the ellipse. The center of this elliptical region is located at $p_c$,the orientation of the region is $\theta$, the position of a random point is $\delta$. This equation is used to determine whether a given point lies within the elliptical region. Equation (4.23) represents the boundary of the aforementioned elliptical region where $\rho$ is the rotation angle that is between 0 to $2\pi$. Equation (4.24) demonstrates the design of the barrier function between two arbitrary elliptical regions $i$ and $j$ where $h_i$ represents the barrier function of elliptical region $i$, $\mathcal{E}_j(\rho)$ represents the boundary function of the elliptical region $j$, and $\xi_{ci}, \xi_{cj}$ represent the position and orientation of the two elliptical regions. By incorporating this barrier function into the control framework, the vehicle can dynamically adjust its trajectory to avoid collisions while preserving stability and operational efficiency. The detailed design of CBF is illustrated in Equation (4.25).

$$h(\delta) = \frac{1}{2}(\delta - p_c)H(\delta - p_c) - \frac{1}{2} \tag{4.22}$$

$$\mathcal{E}(\rho) = \begin{bmatrix} a\cos(\rho)\cos(\theta) - bsin(\rho)\sin(\theta) + x_c \\ a\cos(\rho)\sin(\theta) + bsin(\rho)\cos(\theta) + y_c \end{bmatrix} \tag{4.23}$$

$$h_{ij}(\xi_{ci}, \xi_{cj}) = \min_{\rho\epsilon\mathbb{R}} h_i(\mathcal{E}_j(\rho)) \tag{4.24}$$

$$\frac{\partial h_{ij}}{\partial \xi_{ci}} g_c(\xi_{ci})u_i + \frac{\partial h_{ij}}{\partial \xi_{cj}} g_c(\xi_{cj})u_j + \beta h_{ij}(\xi_{ci}, \xi_{cj}) \geq 0 \tag{4.25}$$

After designing the appropriate CLF and CBF constraints for the optimization-based controller, the next step is to formulate the CLF-CBF-QP framework. The complete formulation of the CLF-CBF-QP is presented in Equation (4.26).

$$u^* = arg\min_{u,\epsilon} \|u\|^2 + q\epsilon^2 \tag{4.26}$$

$$s.t \ \ \dot{V}(\xi_i, u_i) + \alpha V(\xi_i) \leq \epsilon \text{ and}$$

$$\dot{h}_{ij}(\xi_{ci}, \xi_{cj}, u_i, u_j) + \beta h_{ij}(\xi_{ci}, \xi_{cj}) \geq 0, for\ j = 1..N$$

where $\xi_{ci}, \xi_{cj}$ represent the position and orientation of the elliptical regions corresponding to the vehicle and obstacles, respectively. The index $i$ refers to individual vehicles, while $j$ corresponds to individual obstacles. Constants $\alpha$ and $\beta$ are designed to ensure that the system converges to the optimal control solution at an exponential rate. $\epsilon$ is the relaxing term which indicates that the vehicle can temporarily deviate from the path when necessary. Additionally, $q$ is a positive constant introduced to penalize the relaxation of the CLF constraint, encouraging adherence to the desired trajectory. By solving this optimization problem, the autonomous vehicle can effectively avoid potential collisions with obstacles, achieve accurate path tracking to the greatest extent possible, and minimize control effort while ensuring efficient and safe navigation. Next, we demonstrate how to design the CLF-CBF-QP for a complex vehicle dynamic model.

#### 4.2.2.4 HOCLF-HOCBF-QP Controller Design for Vehicle Dynamic Model

In this section, the design of the proposed low-level controller is presented. The key objective of this controller is to compute safe and effective control inputs that allow the vehicle to follow a desired path from the start point to the destination while avoiding collisions with surrounding obstacles (VRUs here). In order to achieve this, we first design the path following part using HOCLF.

$$V(x) = \left(x - x_g\right)^2 + \left(y - y_g\right)^2 \tag{4.27}$$

$$\begin{aligned} \dot{V}(x,u) &= \mathcal{L}_f V(x) + \mathcal{L}_g V(x) u \\ &= 2v\cos\,(\beta + \psi)\left(x - x_g\right) + 2v\sin(\beta + \psi)\left(y - y_g\right) \end{aligned} \tag{4.28}$$

$$\begin{aligned} \mathcal{L}_f^2 V(x) &= \nabla\left(\mathcal{L}_f V(x)\right)^T f(x) \\ &= -2v\sin(\beta + \psi)\left(x - x_g\right)(A_{11}\beta + A_{12}r) + \\ &\quad 2v\cos(\beta + \psi)\left(y - y_g\right)(A_{11}\beta + A_{12}r) + \\ &\quad 2v^2 - 2v\sin(\beta + \psi)\left(x - x_g\right)r + 2v\cos(\beta + \psi)\left(y - y_g\right)r \end{aligned} \tag{4.29}$$

$$\begin{aligned} \mathcal{L}_g\mathcal{L}_f V(x)u = \nabla\left(\mathcal{L}_f V(x)\right)^T g(x)u &= -2v\sin(\beta + \psi)\left(x - x_g\right)B_1\delta_f + \\ &\quad 2v\cos(\beta + \psi)\left(y - y_g\right)B_1\delta_f \end{aligned} \tag{4.30}$$

Equations (4.27)-(4.30) demonstrate the design of the HOCLF and its derivative where $[x, y]$ represent the current coordinates of the vehicle, and $[x_g, y_g]$ are coordinates of the destination. The coefficients $A_{11}$, $A_{12}$, $A_{21}$, $A_{22}$, $B_1$, $B_2$ were defined in the vehicle dynamics model described earlier. The purpose of introducing the HOCLF in this design is to ensure the stability of the system, thereby enabling the vehicle to converge toward the target destination or tracking point on the desired path. The design idea is straightforward. We want the vehicle's position $[x, y]$ to eventually coincide with the destination coordinates $[x_g, y_g]$. To achieve this, we construct a Lyapunov candidate function $V(x)$, which quantifies the squared distance between the current position and the goal. The inequality condition in equation (4.31) is the HOCLF constraint

$$\mathcal{L}_f^2 V(x) + \mathcal{L}_g\mathcal{L}_f V(x)u + \alpha_1\left(\dot{V}(x,u)\right) + \alpha_2\left(V(x)\right) \le \delta \tag{4.31}$$

where $\alpha_1$ and $\alpha_2$ are class-$\kappa$ functions and $\delta$ is a slack variable. In practice, we implement these as positive constant gains. This constraint ensures that the control input $u$ consistently drives the system towards the goal in a stable and controlled manner.

Similarly, we can then design the collision avoidance part using HOCBF as

$$h(x) = (x - x_o)^2 + (y - y_o)^2 - r_o^2 \tag{4.32}$$

$$\begin{aligned} \dot{h}(x,u) &= \mathcal{L}_f h(x) + \mathcal{L}_g h(x)u = \mathcal{L}_f h(x) \\ &= 2v\cos\,(\beta + \psi)(x - x_o) + 2v\sin(\beta + \psi)(y - y_o) \end{aligned} \tag{4.33}$$

$$\mathcal{L}_f^2 h(x) = \nabla\left(\mathcal{L}_f h(x)\right)^T f(x)$$
$$= -2v sin(\beta+\psi)(x-x_o)(A_{11}\beta + A_{12}r) +$$
$$2v\, cos(\beta+\psi)\,(y-y_o)(A_{11}\beta + A_{12}r) + \tag{4.34}$$
$$2v^2 - 2v sin(\beta+\psi)(x-x_o)r + 2v\, cos(\beta+\psi)\,(y-y_o)r$$

$$\mathcal{L}_g\mathcal{L}_f h(x)u = \nabla\left(\mathcal{L}_f h(x)\right)^T g(x)u = -2v sin(\beta+\psi)(x-x_o)B_1\delta_f + \tag{4.35}$$
$$2v\, cos(\beta+\psi)\,(y-y_o)B_1\delta_f$$

Equations (4.32)-(4.35) demonstrate the design of the HOCBF and its derivative where $[x, y]$ represent the current coordinates of the vehicle, and $[x_o, y_o]$ are coordinates of the obstacles. The coefficients $A_{11}$, $A_{12}$, $A_{21}$, $A_{22}$, $B_1$, $B_2$ were defined in the vehicle dynamics model described earlier. The purpose of introducing the HOCBF in this design is to enforce safety by ensuring that the vehicle avoids potential collisions with surrounding obstacles. The design idea is straightforward. We want to prevent the vehicle's position $[x, y]$ from entering a circular danger zone of radius $r_o$ centered at $[x_o, y_o]$. To achieve this, we construct a barrier candidate function $h(x)$ which quantifies whether the center of the vehicle is entering the dangerous zone or not. The inequality condition in equation (4.36) is the HOCBF constraint. For each obstacle, we need a unique HOCBF constraint

$$\mathcal{L}_f^2 h(x) + \mathcal{L}_g\mathcal{L}_f h(x)u + \alpha_3\left(\mathcal{L}_f h(x)\right) + \alpha_4\big(h(x)\big) \geq 0 \tag{4.36}$$

where $\alpha_3$ and $\alpha_4$ are class-$\kappa$ functions. In practice, we implement these as positive constant gains. This constraint ensures that the control input $u$ cannot drive the system towards the unsafe region with obstacles.

To combine the previous design of HOCLF and HOCBF within a single control framework, we need to formulate a QP that incorporates both HOCLF and HOCBF as inequality constraints. This optimization-based approach enables the controller to compute control inputs that not only drive the system towards the desired destination but also maintain safety by avoiding the entering of unsafe regions with obstacles.

The formulation of the QP is presented in equation (4.37).

$$u^* = arg\min_{u,\delta}\left\|u - u_{ref}\right\|^2 + q\delta^2 \tag{4.37}$$
$$s.t\ \ \mathcal{L}_f^2 V(x) + \mathcal{L}_g\mathcal{L}_f V(x)u + \alpha_1\left(\dot{V}(x,u)\right) + \alpha_2\big(V(x)\big) \leq \delta$$
$$\text{and}$$
$$\mathcal{L}_f^2 h(x) + \mathcal{L}_g\mathcal{L}_f h(x)u + \alpha_3\left(\mathcal{L}_f h(x)\right) + \alpha_4\big(h(x)\big) \geq 0, for\ each\ obstacle$$

where $u$ is the control input, $u_{ref}$ is a nominal reference input, which can be set to zero for minimizing control effort. $\delta$ is a slack variable of the HOCLF constraint, allowing temporary relaxation of the HOCLF constraint when it is in conflict with HOCBF. The penalty weight $q > 0$ balances the performance and constraint violation

The parameters in the HOCLF-HOCBF-QP controller include the CLF constraint coefficients, CBF constraint coefficients, and CLF relaxing terms $\delta$ in QP formulations. The CLF constraint coefficients affect tracking aggressiveness. Higher values improve convergence speed but may lead to more abrupt control actions. Moreover, if the CLF constraint coefficients have an excessively large value, that will impose overly strict convergence requirements, which may lead to an infeasible optimization problem. Increasing the CBF constraint coefficients makes the safety constraint less strict, allowing the vehicle to follow more efficient trajectories but reducing the safety margin. Conversely, reducing the CBF constraint coefficients enforces stricter safety constraints, forcing the vehicle to maintain larger distances from obstacles. The CLF relaxing terms $\delta$ influences the trade-off between strict constraint satisfaction and feasible control effort. Compared to other parameters, the CLF constraint coefficients have a more direct and sensitive impact on the feasibility and effectiveness of the overall control performance.

### 4.2.3 High-level DRL Based Collision Avoidance Controller Design

Markov Decision Process (MDP) is a framework used in modeling sequential decision-making problems. In an MDP, the system evolves over time by selecting actions based on the current state. The goal is to find an optimal policy that can maximize long-term cumulative reward. Autonomous driving naturally fits into this framework. At each moment, driving decisions such as whether to change lanes, accelerate, or slow down are made based on the current vehicle state and traffic environment. Therefore, in this section, we formulate the high-level autonomous driving task as an MDP and try to find the optimal driving strategy.

Once the autonomous diving problem is formulated using MDP, reinforcement learning (RL) can then be employed to optimize the decision-making process and find an optimal policy. References [86-87] present different applications of reinforcement learning. Deep Reinforcement Learning (DRL), a subset of RL, is particularly effective for autonomous driving applications and can be broadly categorized into value-based and policy-based methods. Value-based DRL, inspired by Q-learning [88], estimates action values to determine the best possible decision. Techniques such as Deep Q-Networks (DQN), Double Deep Q-Networks (DDQN) and their successors [89-92] improve learning efficiency in environments with large state spaces and discrete action spaces. Policy-based DRL [93], on the other hand, directly learns a mapping from states to actions without explicitly estimating value functions. Methods such as Policy Gradient (PG), Actor-Critic (A2C [94], A3C [95]), Proximal Policy Optimization (PPO) [96], and their successors [97-99] refine decision-making by adjusting policy parameters to maximize expected rewards.

In this section, we propose using the DQN framework to design the high-level control for the unicycle model and propose using DDQN to design the high-level control for the vehicle dynamic model. Both DQN and DDQN, as value-based reinforcement learning algorithms, offer several advantages over policy-based algorithms. They are computationally efficient, particularly for discrete action spaces, and can achieve stable convergence with techniques like experience replay and target networks. These properties make them ideal choices for handling high-level decision-making in structured environments.

#### 4.2.3.1 DDQN High-Level Decision-Making Agent for Unicycle Model

For the unicycle model, we convert the driving environment into a grid map, where the map is divided into discrete grids. Each grid represents a specific location, with grids containing the destination and obstacles being distinctly marked. At each time step, the DQN framework determines a high-level decision, guiding the vehicle to move to a nearby grid based on its learned policy. The CLF-CBF controller then executes these high-level decisions, refining the vehicle's trajectory and ensuring stability, safety, and efficiency throughout the process. This combination of DQN for decision-making and CLF-CBF for control execution ensures a robust and effective approach to autonomous driving in dynamic environments.

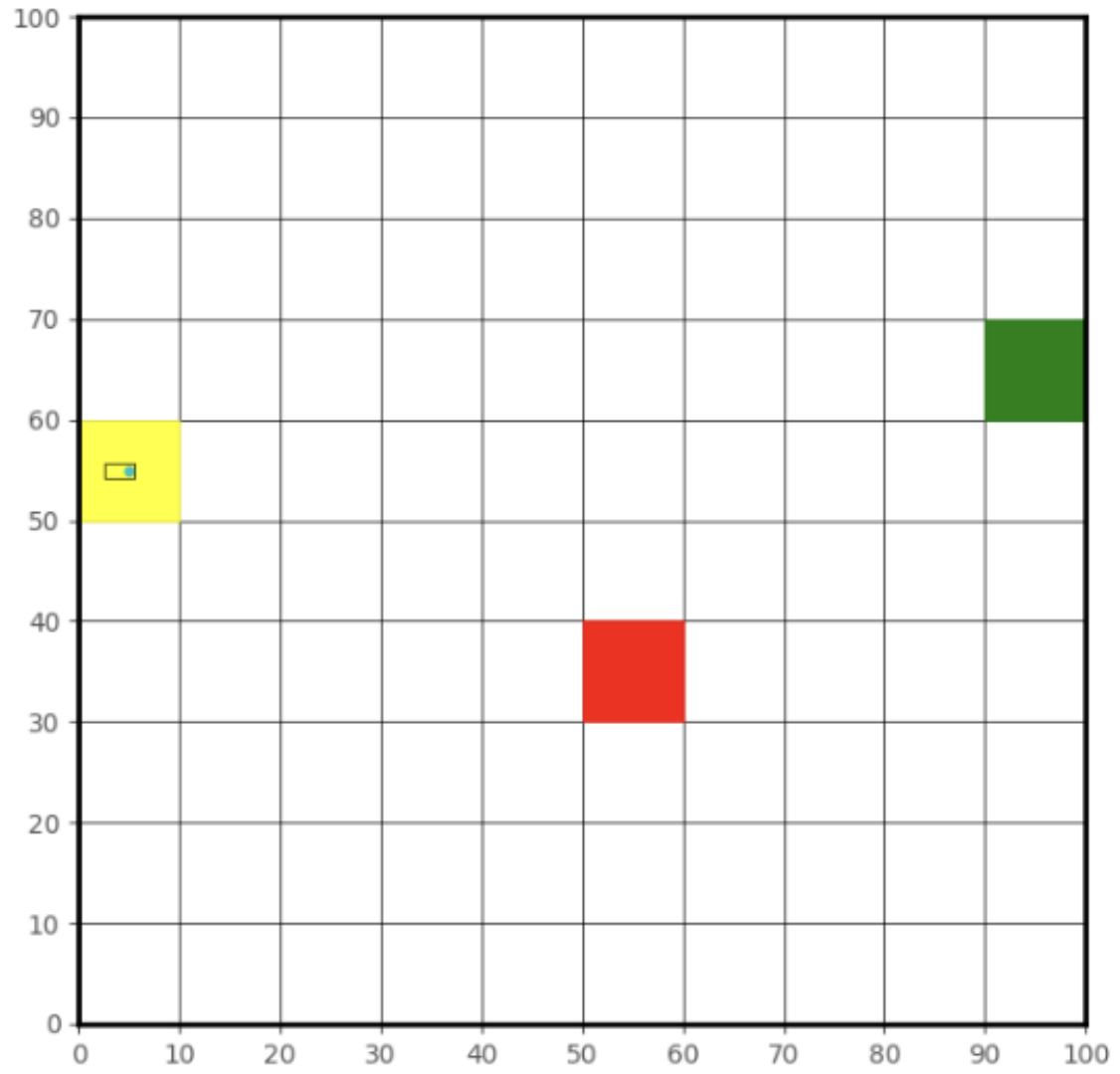

**Figure 4.3.** Deep reinforcement learning traffic environment setting.

Figure 4.3 illustrates an example of the traffic environment which can be used to train a DRL based high level decision-making agent. The yellow grid represents the vehicle's current position, while the red grid indicates the presence of a dynamic obstacle. The green grid marks the destination. The vehicle's objective is to navigate around the dynamic obstacle and reach the destination as quickly as possible. The DRL-based high-level decision-making agent is responsible for generating basic navigation commands, such as moving to the grid above or below the current position. These commands serve as guidance for the overall trajectory planning. The CLF-CBF-QP-based low-level controller, in turn, interprets and executes these commands with precision, ensuring smooth and safe motion while adhering to vehicle dynamics and avoiding collisions. This hierarchical structure enables the seamless integration of high-level strategic decision-making with low-level control execution, providing both flexibility and robustness in navigation.

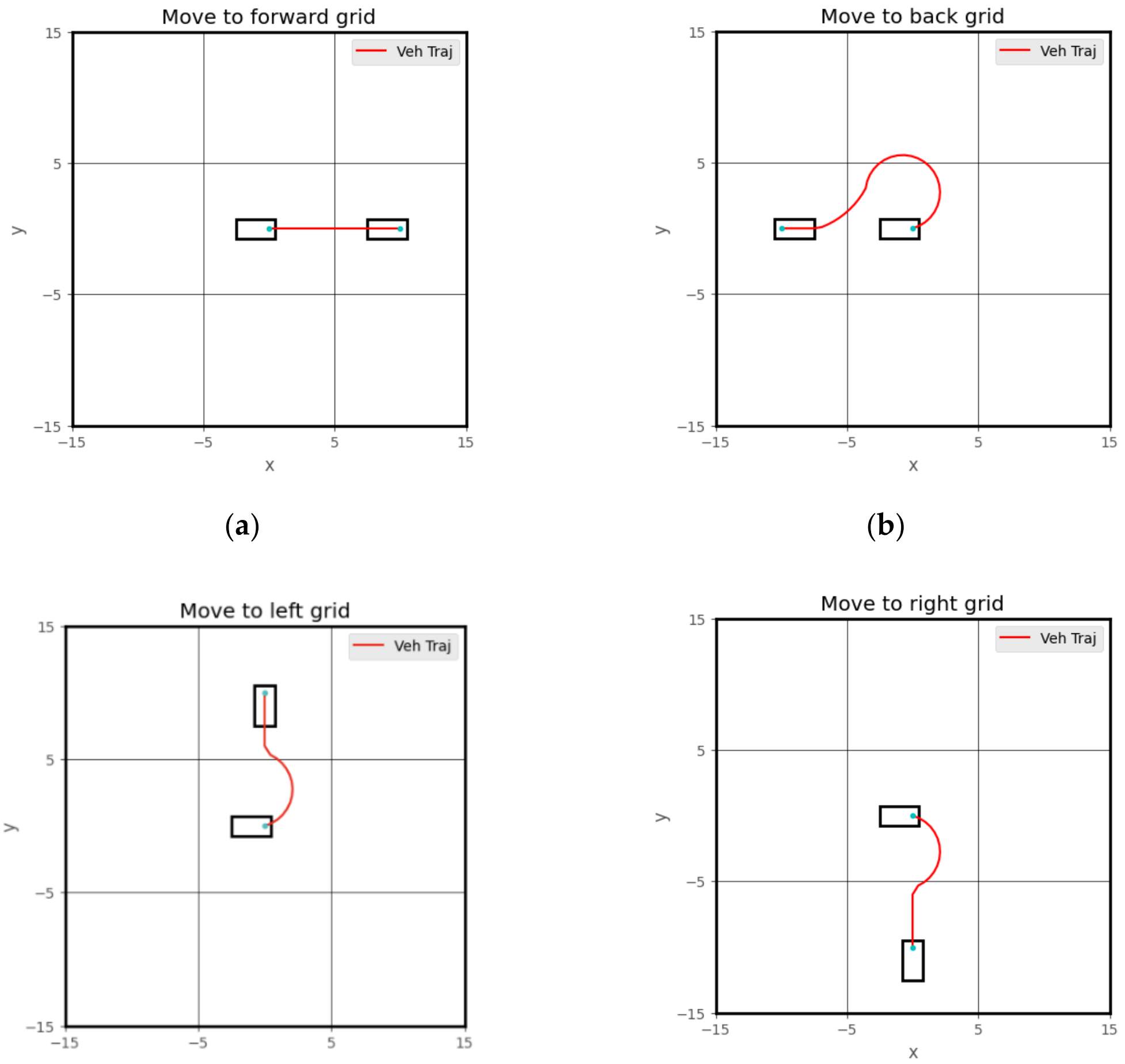

(**a**) (**b**)

(**c**) (**d**)

**Figure 4.4.** Sample low level control steps: (**a**) move to forward grid; (**b**) move to back grid; (**c**) move to left grid; (**d**) move to right grid.

Figure 4.4 demonstrates how to use the CLF-CBF controller to execute high-level steps generated by the DRL agent. Each high-level decision corresponds to a unique trajectory. After the DRL-based high-level controller decides, the CLF-CBF controller ensures that the vehicle moves from the center of the current grid to the center of the next grid. Notably, there are two possible ways to move to the grid behind the vehicle: either a left turn or a right turn, both of which are feasible. To handle such scenarios, an additional rule-based algorithm can be incorporated to select the optimal path based on specific conditions. Furthermore, the DRL controller may occasionally decide that the vehicle should remain stationary. If the DRL chooses not to move, the low-level controller will maintain the vehicle's position until the next step.

**Algorithm 4.1.** DQN algorithm flowchart

**Algorithm 4.1: DQN algorithm flowchart**

Initialize replay memory $D$
Initialize target network $\hat{Q}$ and Online Network $Q$ with random weights $\theta$
**for** each episode **do**
  Initialize traffic environment
  **for** t = 1 to T **do**
    With probability $\epsilon$ select a random action $a_t$
    Otherwise select $a_t = \max_a Q^*(s_t, a; \theta)$
    Execute $a_t$ in CARLA and extract reward $r_t$ and next state $s_{t+1}$
    Store transition ($s_t$, $a_t$, $r_t$, $s_{t+1}$) in $D$
    **if** t mod training frequency == 0 **then**
      Sample random minibatch of transitions ($s_j$, $a_j$, $r_j$, $s_{j+1}$)) from D
      Set $y_j = r_j + \gamma \max_{a'} \hat{Q}(s_{j+1}, a'); \theta)$
      for non-terminal $s_{j+1}$
      or $y_j = r_j$ for terminal $s_{j+1}$
      Perform a gradient descent step to update $\theta$
      Every N steps reset $\hat{Q} = Q$
    **end if**
    Set $s_{t+1} = s_t$
  **end for**
**end for**

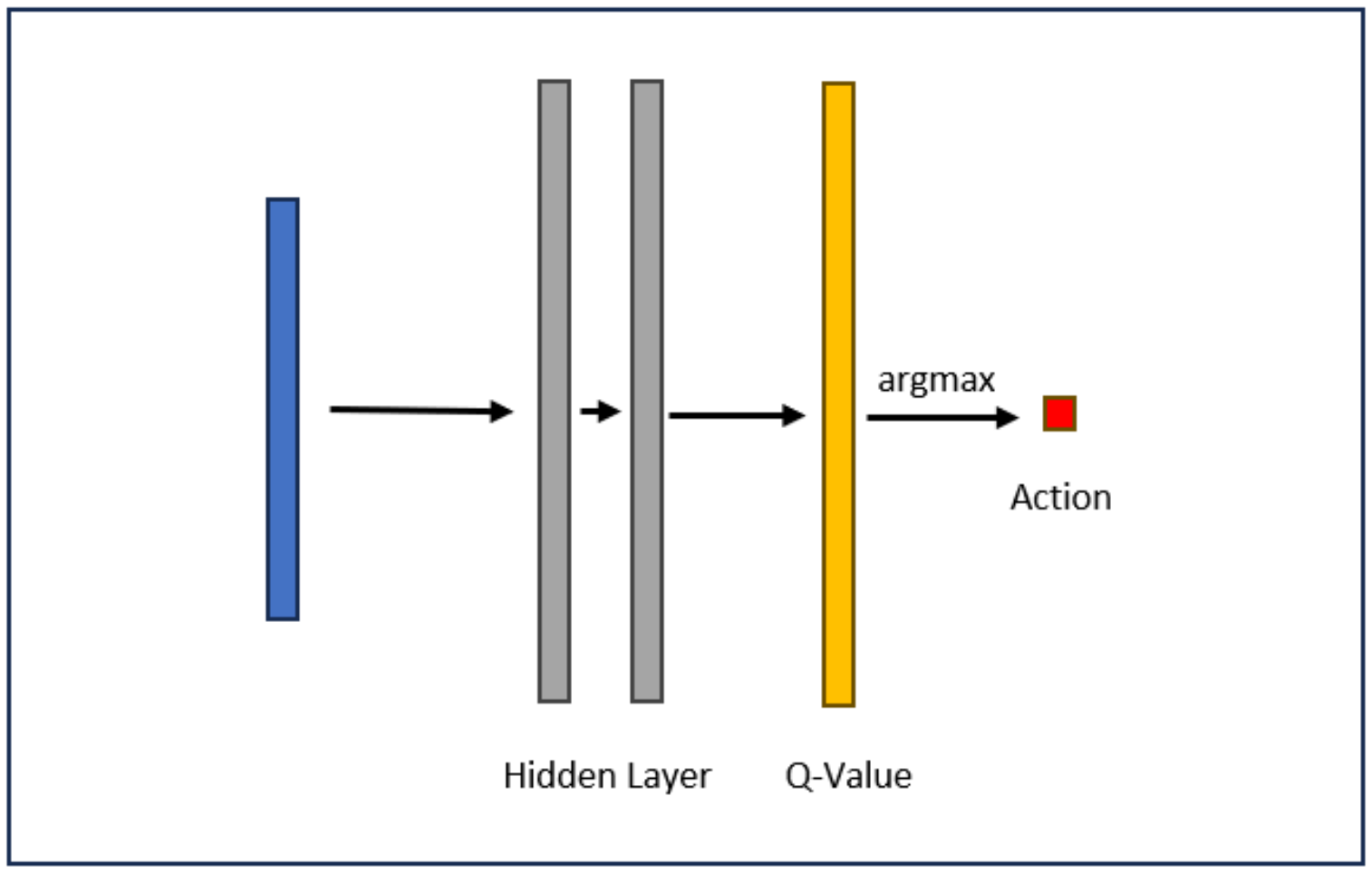

**Figure 4.5.** DQN framework neural network structure

Algorithm 4.1 demonstrates the DQN framework that is used to train the autonomous driving high-level decision-making agent. The Q-value, also known as action-value, represents the expected cumulative reward that an agent can obtain starting from a specific state $s$ and then taking a specific action $a$. In DQN, a neural network approximates the Q-value function. The network takes the current state $s$ as input and outputs a vector of Q-values, one for each possible action in the action space. Figure 4.5 demonstrates the structure of the neural network used in the DQN framework. The neural networks used in this paper are feedforward neural networks. The proposed DQN agent contains two fully connected hidden layers that each contain 32 neurons. The corresponding activation function for the hidden layer is ReLU. The learning rate for training is set to 0.001, using the Adam optimizer, and the model evaluates training performance using the mean absolute error (MAE) metric. The agent undergoes 1,000 warm-up steps before learning and updates the target network with a rate of 0.01.

#### 4.2.3.2 DDQN High-Level Decision-Making Agent for Vehicle Dynamic Model

To design high-level decision-making agent for vehicle dynamic models, we use a different approach. Through observations of daily driving behavior, we find that most human drivers tend to perform lane-changing maneuvers to avoid obstacles, such as stationary vehicles or unexpected hazards. Inspired by this observation, we developed a DDQN high-level decision-making agent for autonomous vehicles which can automatically perform lane-changing maneuvers to avoid obstacles in multi-lane traffic environments. This approach is also applicable to suddenly appearing VRUs.

**Table 4.2.** DDQN loss function parameters.

| Symbol | Parameter |
| --- | --- |
| $s$ | State |
| $a$ | Action |
| $r$ | Immediate reward |
| $\theta_i^-$ | Target-network's parameter |
| $\theta_i$ | Online-network's parameter |
| $\gamma$ | Discount for future reward |

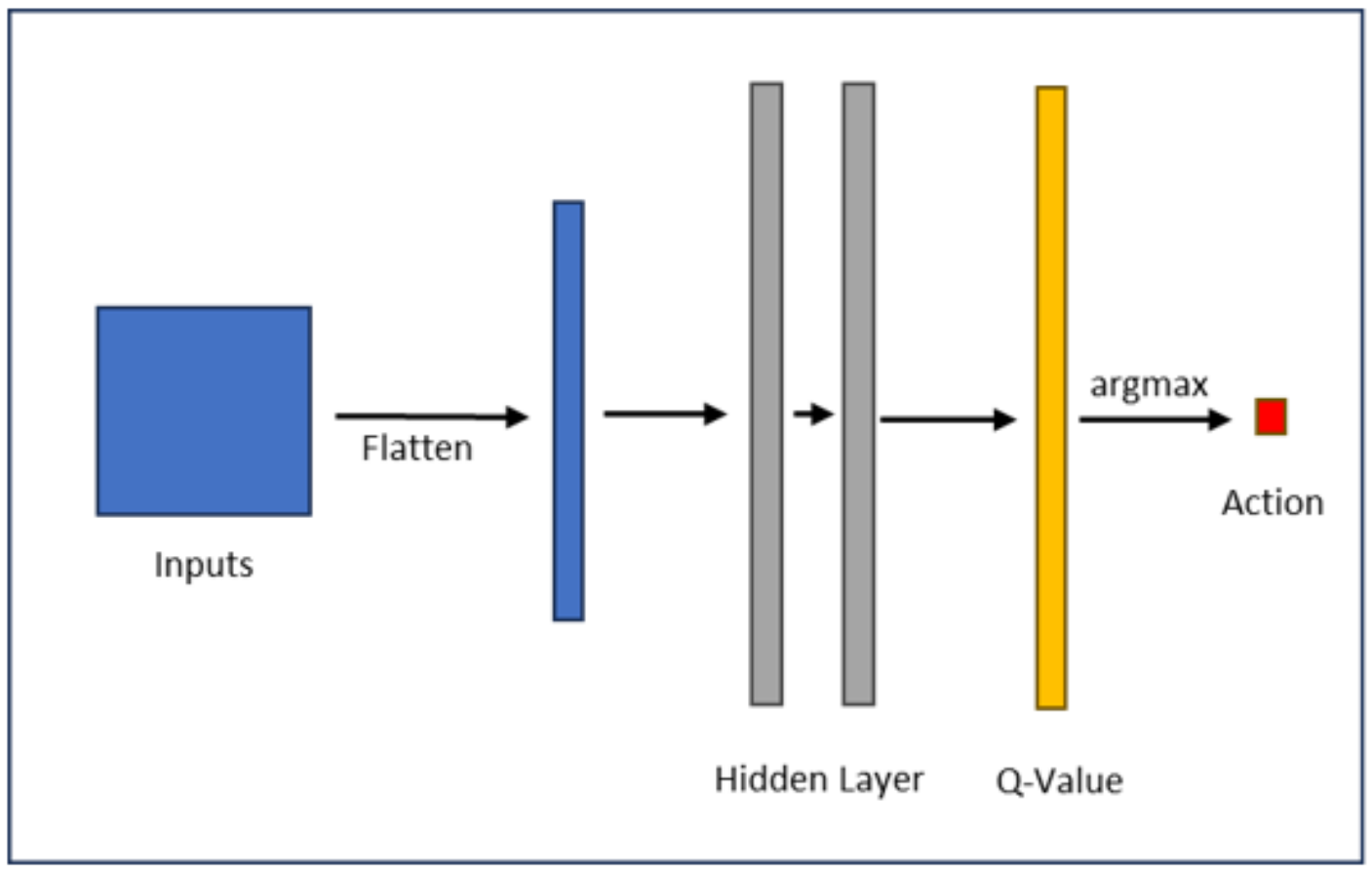

**Figure 4.6.** DDQN framework neural network structure

Since the high-level decision-making agent is required to generate discrete lane-level actions, a value-based reinforcement learning method is particularly well-suited. Compared with policy-based methods, value-based approaches like Q-learning and its variants tend to be more sample-efficient and easier to train in discrete action spaces. Given the relatively simple structure of our task and the need for stable and efficient learning, we adopt the DDQN algorithm to train the high-level agent. From the loss function of DDQN shown in equation (4.38), DDQN mitigates the overestimation bias found in standard DQN by decoupling action selection and action evaluation during Q-value updates (using $\max_{a_{t+1}} Q_{\theta_i^-}\left(s_{t+1}, arg\max_{a_{t+1}} Q_{\theta_i}(s_{t+1}, a_{t+1})\right)$ instead of $\max_{a_{t+1}} Q_{\theta_i^-}(s_{t+1}, a_{t+1})$ for update), leading to a more stable training process. Table 4.2 provides a detailed explanation of the parameters in loss functions. Moreover, while DDQN is used in this work due to its simplicity and effectiveness, the modular design of our hierarchical framework allows for flexibility. For complex driving tasks, more advanced algorithms such as SAC or DDPG can be applied for better performance.

$$L_i(\theta) = \mathbb{E}_{(s,a,r)}\left[\left(r + \gamma \max_{a_{t+1}} Q_{\theta_i^-}\left(s_{t+1}, arg\max_{a_{t+1}} Q_{\theta_i}(s_{t+1}, a_{t+1})\right) - Q_{\theta_i}(s_t, a_t)\right)^2\right] \tag{4.38}$$

**Algorithm 4.2.** Hierarchical DDQN algorithm flowchart

**Algorithm 4.2**

Initialize high-level DDQN agent
Initialize low-level HOCLF-HOCBF-QP controller
**for** each episode **do**
    Initialize and reset traffic environment
    **for** t = 1 to T **do**
        With probability $\epsilon$ select a random high-level action $a_t$
        Otherwise select high-level action $a_t = \max_a Q^*(s_t, a; \theta)$
        **for** each low-level control steps **do**
            Reset target tracking point according to current states and $a_t$
            Reset obstacle list according to current states
            Calculate optimal control $u^*$ using HOCLF-HOCBF-QP
            Execute $u^*$ for a control step
        **end for**
        Calculate reward $r_t$ and record next state $s_{t+1}$
        Store transition ($s_t$, $a_t$, $r_t$, $s_{t+1}$) in replay buffer $D$
        **if** t mod training frequency == 0 **then**
            Sample random minibatch of transitions ($s_j$, $a_j$, $r_j$, $s_{j+1}$)) from D
            Set $y_j = r_j + \gamma \max_{a_{j+1}} \hat{Q}(s_{j+1}, \text{argmax}_{a_{j+1}} Q(s_j, a_{j+1}; \theta); \theta)$
            for non-terminal $s_{j+1}$
            or $y_j = r_j$ for terminal $s_{j+1}$
            Perform a gradient descent step to update $\theta$
            Every N steps reset $\hat{Q} = Q$
        **end if**
        Set $s_{t+1} = s_t$
    **end for**

26: **end for**

Figure 4.6 illustrates the neural network structure used in the DDQN framework and Algorithm 4.2 demonstrates the pseudocode of hierarchical DDQN implementation. The DDQN employs a fully connected feedforward neural network consisting of two hidden layers, each with 128 neurons using ReLU as activation functions. The input layer has 25 units (flattened from 5×5 including ego vehicle's information and traffic environment information), and the output layer has 3 units corresponding to three lane-level actions. The training process uses a standard replay buffer with a size of 100,000 and a mini-batch size of 64. The Q-network is updated using the mean squared error (MSE) loss between the predicted Q-values and the target Q-values. The Adam optimizer is used with a learning rate of 0.001. To stabilize training and mitigate overestimation bias, a target network is maintained and synchronized with the main Q-network every 100 learning steps. An $\varepsilon$-greedy exploration strategy is adopted, where $\varepsilon$ linearly decays from 1.0 to 0.05 over 200,000 steps. This linear decay strategy allows the agent to fully explore the environment instead of converging to a local optimal policy.

## 4.3 Results

### 4.3.1 Unicycle Model Testing Results

In this section, we present the simulation results of the proposed controller for the unicycle model to evaluate its performance and robustness. First, the CLF-based path-tracking simulation results are shown, highlighting the controller's ability to achieve precise path tracking under normal conditions. Next, the simulation results of the CLF-CBF-based autonomous driving controller are shown for scenarios involving both static and dynamic obstacles. These results illustrate the vehicle's capability to maintain precise path tracking during normal traffic conditions and effectively execute collision avoidance maneuvers in emergency situations. Finally, we present the simulation results of the hybrid framework, where the DQN-based high-level decision-making agent is combined with the CLF-CBF-based low-level controller. These results demonstrate how integrating traditional optimization-based controllers with deep reinforcement learning can significantly enhance the autonomous driving capabilities of vehicles, enabling better decision-making and improved driving safety in complex traffic environments. Moreover, all the test conditions are initially evaluated using simulations within a Python-based environment. These simulations are then followed by Simulink-based real-time model-in-the-loop (MIL) and hardware-in-the-loop (HIL) simulations, the latter validating the real-time capabilities of the proposed controller. This two-stage testing process ensures both the feasibility and practicality of the controller in dynamic and real-time scenarios.

Figure 4.7 shows the real-time MIL simulation results of the CLF based path tracking controller. In the figure, x and y coordinates represent the bird's-eye view map of the testing traffic conditions, with the units for both the x- and y-axes being meters. The figure illustrates that the vehicle follows the desired path with high precision under most conditions, demonstrating the controller's effectiveness in path tracking tasks. However, a slight deviation from the original path is observed in regions with higher curvature, where tracking accuracy decreases marginally. This minor deviation could be attributed to the dynamic complexity introduced by the path's curvature, which challenges the controller's ability to maintain perfect alignment. Despite these small discrepancies, the overall performance, the CLF-based path-tracking controller is robust, showing its capability to handle real-time scenarios effectively while maintaining accurate path tracking. Further refinements or adjustments may help address the minor tracking offsets in curved sections to improve performance further. Figure 4.8 shows the steering angle response $\theta$ of the vehicle over time during the simulation. The plot indicates that the controller maintains stable steering behavior throughout the scenario.

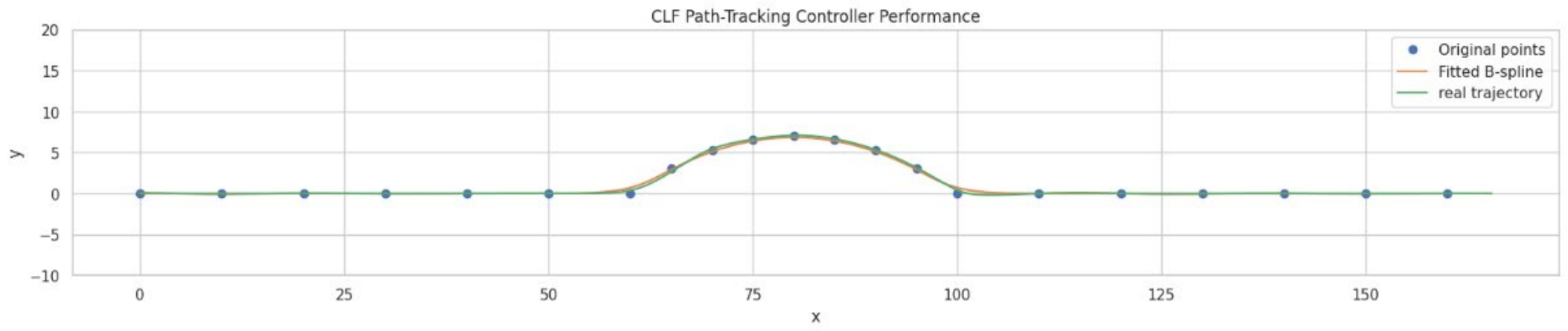

**Figure 4.7.** CLF Path Tracking Controller MIL results. x and y coordinates are in m.

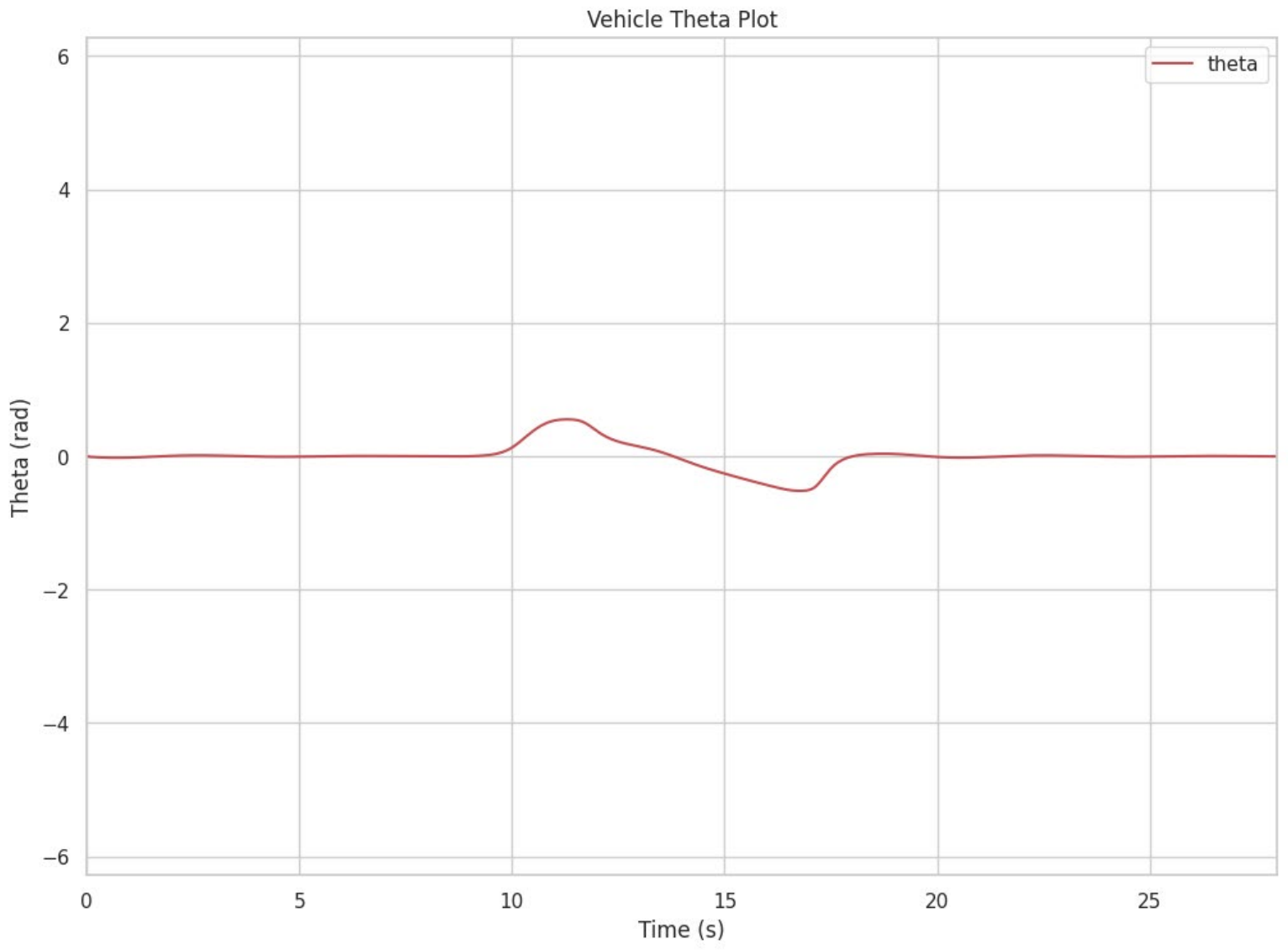

**Figure 4.8.** CLF Path Tracking Controller Theta Plots.

Figure 4.9 presents the simulation results of the CLF-CBF-based autonomous driving controller in an environment with dynamic obstacles. In the figure, x and y coordinates represent the bird's-eye view map of the testing traffic conditions, with the units for both the x- and y-axes being meters. The original path waypoints are marked with blue dots, the fitted B-spline trajectory in orange, and the vehicle's actual trajectory in green. A dynamic obstacle, represented as a red circle, moves in a circular pattern. When the obstacle moves to the top and blocks a section of the pre-planned trajectory, the vehicle is required to deviate from the original path to avoid a collision. The simulation results demonstrate the controller's ability to adjust the vehicle's trajectory dynamically, ensuring safe navigation around the obstacle. The smooth transitions and minimal deviation from the intended path highlight the robustness of the CLF-CBF framework in handling dynamic obstacles effectively. Additionally, Figure 4.10 illustrates the vehicle's steering angle response $\theta$ over time during the simulation. The plot reveals stable steering behavior throughout the scenario, with minor oscillations in the middle phase as the vehicle adjusts to the desired trajectory while avoiding the obstacle. The demo video link of this scenario is attached at the end of the section.

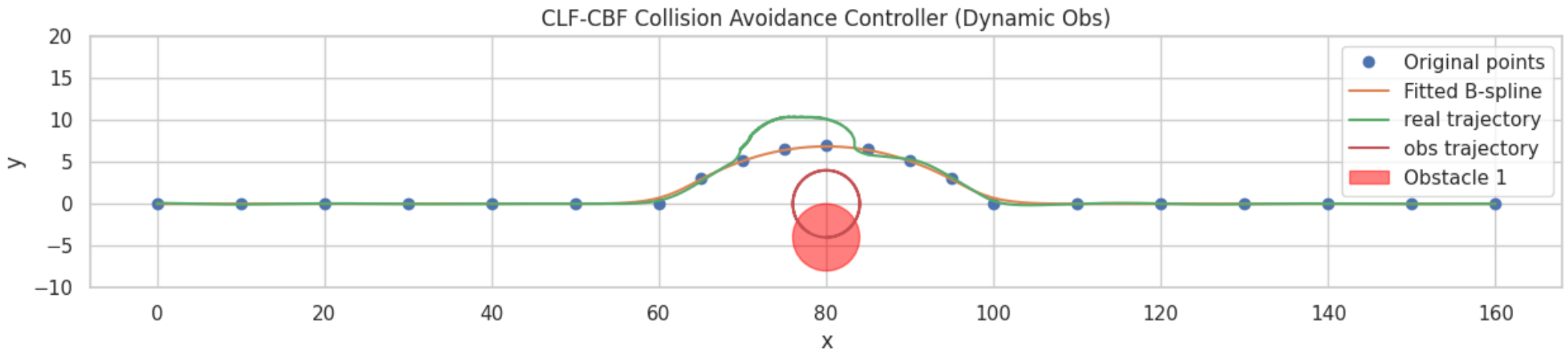

**Figure 4.9.** CLF-CBF based Autonomous Driving Controller for Dynamic Obstacle. x and y coordinates are in m.

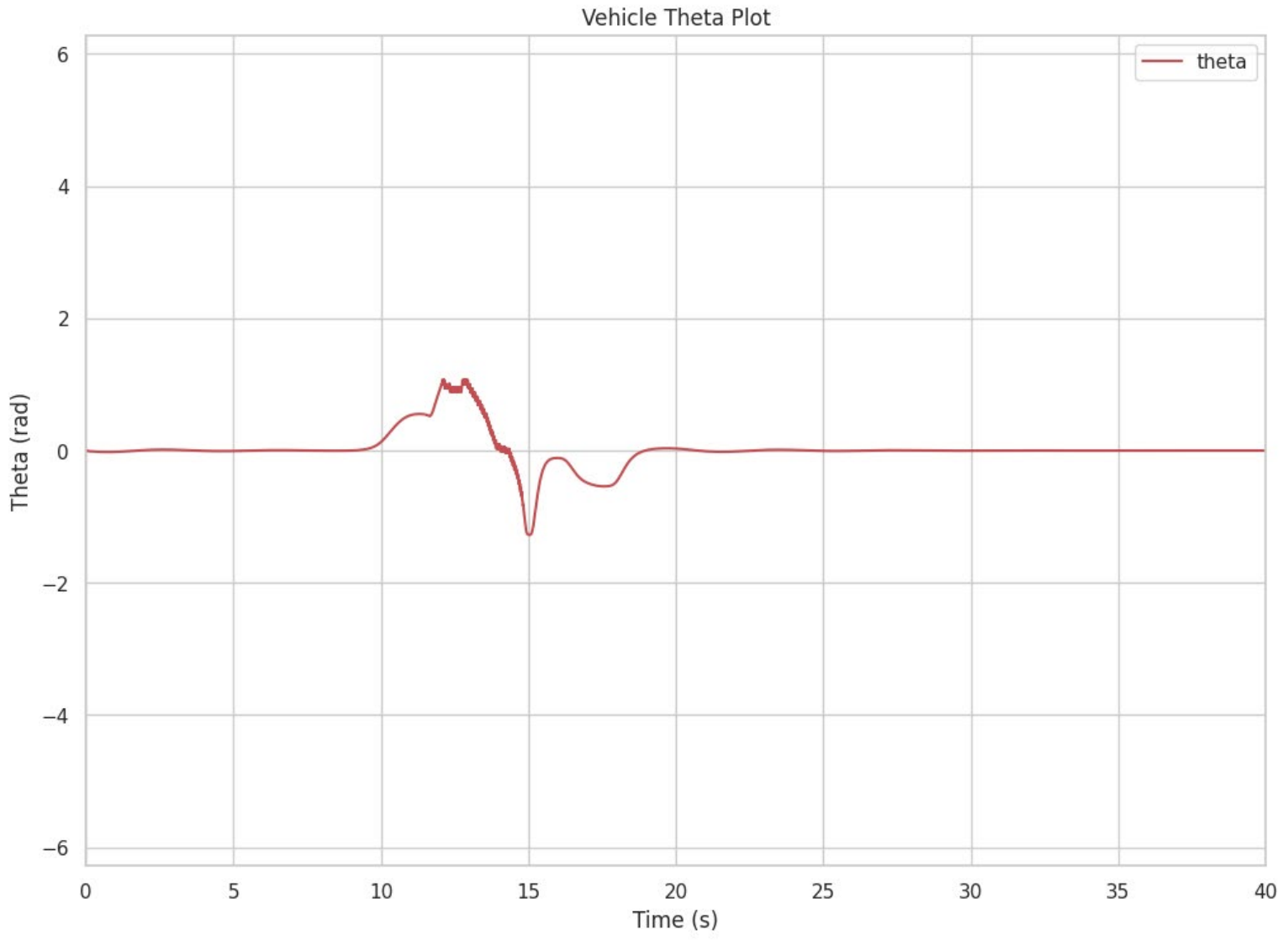

**Figure 4.10.** CLF-CBF based Controller Steering Plot for Dynamic Obstacle.

Then, we present simulations results of the hybrid DRL and CLF-CBF-based autonomous driving controller. The vehicle starts at a designated initial position and needs to navigate to a specified destination, with a freely moving obstacle present in the environment. There is no pre-calculated path between the initial position and the destination. Instead, the high-level DRL agent dynamically calculates a rough path based on the vehicle's status and the surrounding traffic conditions, while the low-level CLF-CBF-based controller executes the agent's decisions. The objective of the simulation is to evaluate whether the controller can effectively perform collision avoidance and ensure safe navigation in a dynamic environment.

In this experiment, the state representation includes the distance between the vehicle and the obstacle, as well as the distance between the vehicle and the destination. The action space consists of four possible movements: forward, backward, left, and right. A positive reward of +25 is assigned when the vehicle successfully reaches the destination grid. Conversely, a large negative reward of -300 is imposed if the vehicle collides with an obstacle by entering its grid. Additionally, a small negative reward of -1 is applied for every move to encourage the vehicle to reach the destination as quickly as possible, promoting efficient navigation.

Figure 4.11 demonstrates the training process of the proposed DRL high-level decision-making agent, which indicates significant improvements in the agent's performance over time. The reward plot shows a sharp increase during the early stages, starting from a highly negative value -6.2 and stabilizing near 0.5 after 300,000 steps, indicating that the agent quickly learned a basic policy and continued to refine it. The loss decreases rapidly from an initial high value 106 to a more stable range after 100,000 steps, with minor fluctuations throughout until the end, which reflects that the agent can effectively minimize prediction error. Similarly, the mean Q-values show a notable increase, transitioning from early negative values to a stable value range around 20 after 100,000 steps. This demonstrates improved confidence in action-value estimations. Together, these three plots indicate the agent's ability to effectively learn and optimize its policy, balancing exploration and exploitation to achieve improved performance as training progresses. The minor fluctuations in loss and reward suggest that further fine-tuning may still enhance stability and performance.

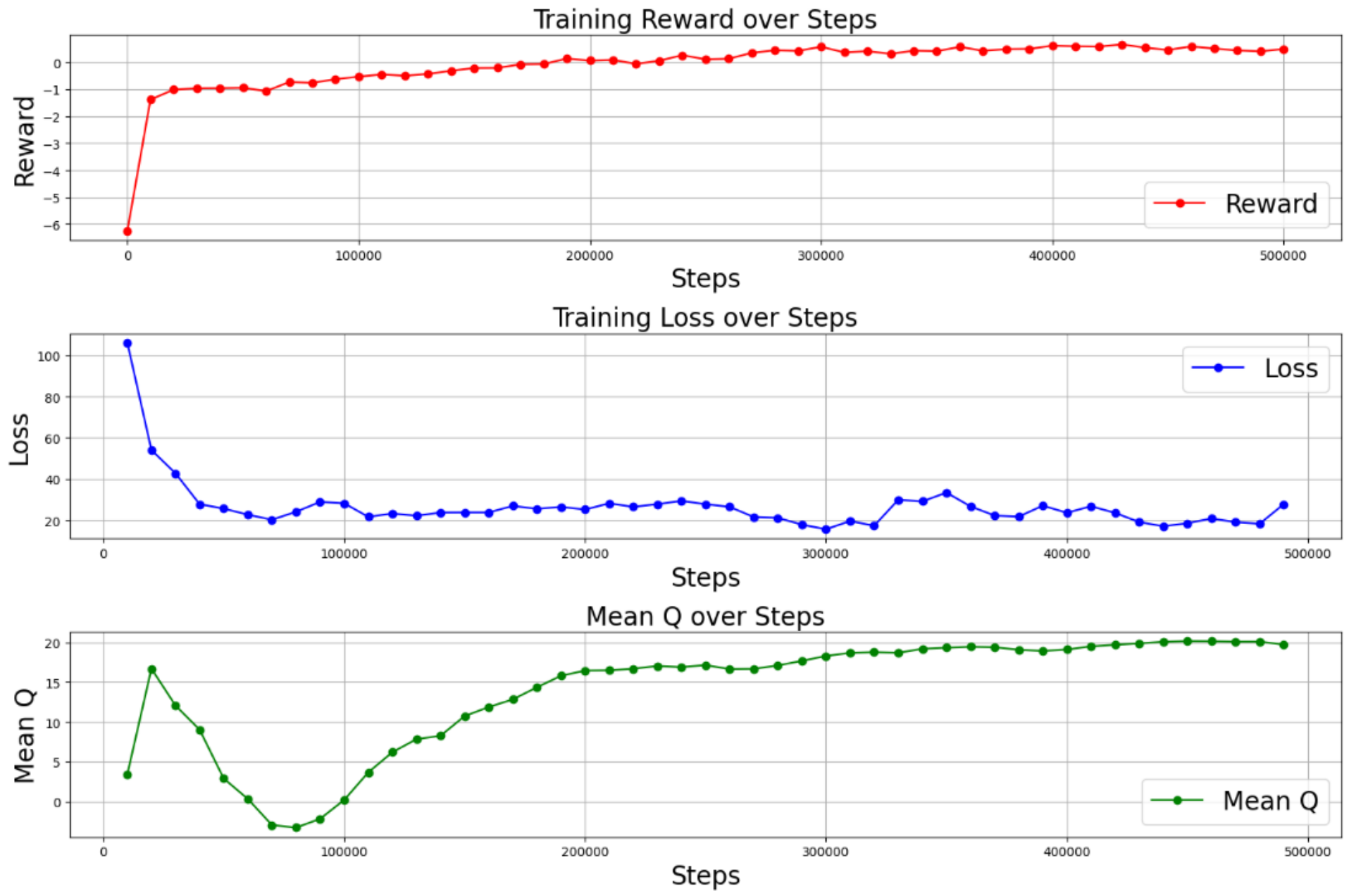

**Figure 4.11.** Deep reinforcement learning training progress.

Figure 4.12 demonstrates an example of the proposed DRL high-level decision-making agent. In the figure, x and y coordinates represent the bird's-eye view map of the testing traffic conditions, with the units for both the x- and y-axes being meters. The blue line indicates the trajectory of the proposed agent, and the red line indicates the motion of obstacles. It is shown that the agent can navigate through the grid while avoiding obstacles. The agent begins from the starting position on the left and successfully reaches the goal area on the right. The smooth progression of the blue line highlights the agent's ability to make efficient decisions to circumvent the moving obstacles while maintaining a clear path towards the target. In contrast, the red trajectory depicts the dynamic movement of obstacles, adding complexity to the environment. This example demonstrates the agent's effective decision-making capabilities in handling high-level planning and real-time obstacle avoidance. The demo video link of other scenarios is attached at the end of the section.

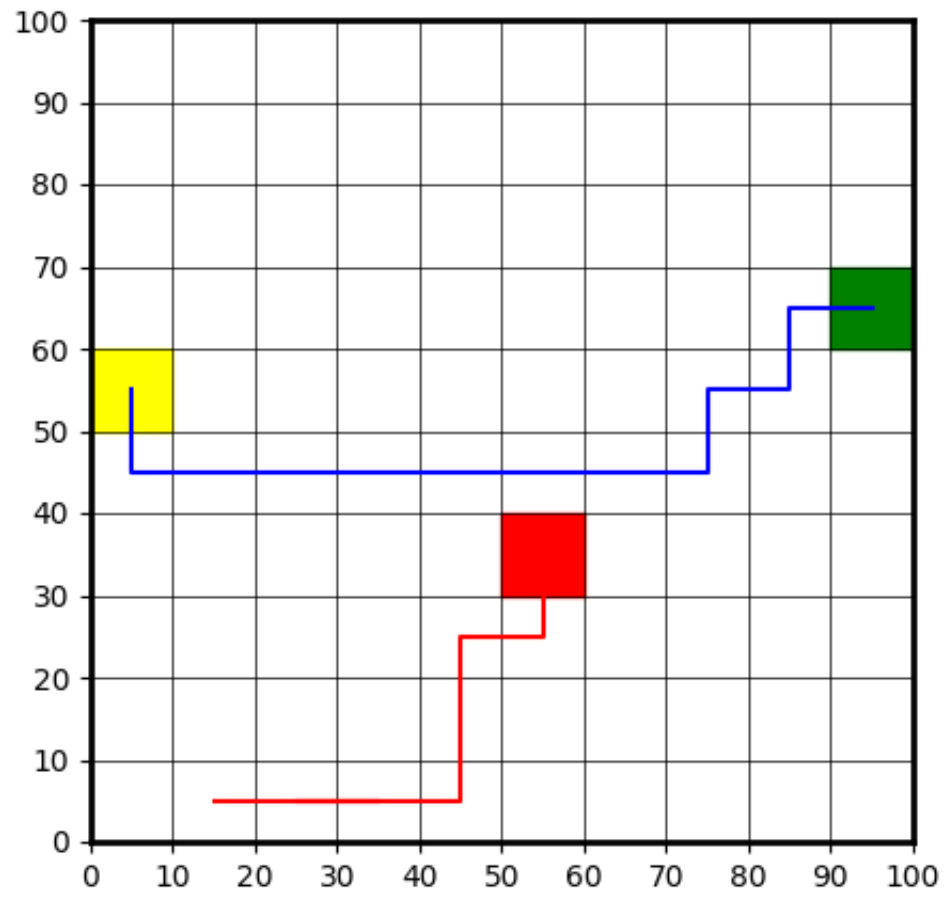

**Figure 4.12.** DRL High-Level Agent Demo. Coordinates units are in m. The trajectory of the obstacle is marked in red line.

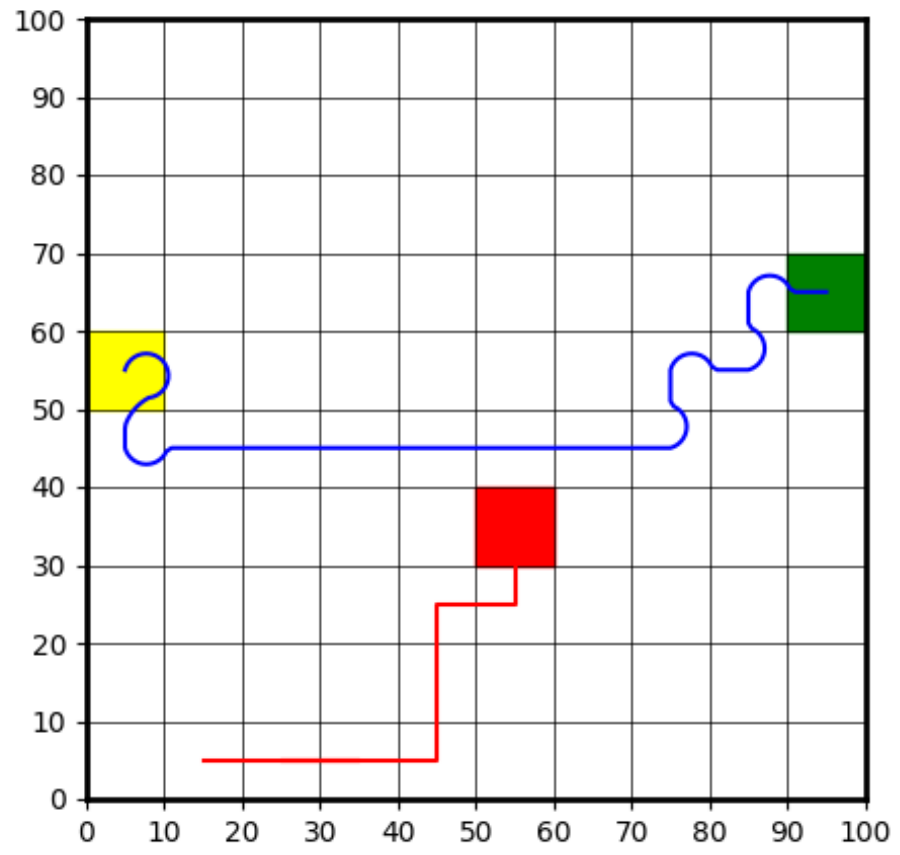

**Figure 4.13.** DRL High-Level Agent Trajectory for Vehicle. Coordinates units are in m. The trajectory of the obstacle is marked in red line.

Figure 4.13 demonstrates the trajectory generated by the proposed DRL high-level decision-making agent. The blue line indicates the real trajectory which can be tracked by the vehicle. From Figure 4.13, the rough sketch generated by the DRL high-level agent is successfully converted into a control feasible path that can be tracked by the vehicle. Combined with CLF-CBF-QP-based control, the autonomous vehicle can follow this path, navigating from the starting point to the endpoint without colliding with obstacles.

Figure 4.14 shows the control flow chart of proposed DRL and CLF-CBF-QP Based hybrid autonomous driving control. In this framework, the environment provides the DRL high-level decision-making agent with key information, such as the relative distance between the vehicle, obstacles, and the destination, which serves as input for decision-making. Simultaneously, the environment sends path tracking data to the CLF-CBF-QP-based low-level controller, ensuring precise trajectory tracking. The unicycle model, on the other hand, supplies the vehicle's position and orientation to both the environment and the low-level controller for updates and control command calculations. At each step, the DRL high-level decision-making agent determines the next grid to move to based on the vehicle's status and surrounding information. Once the next grid is selected, the CLF-CBF-QP-based low-level controller executes the decision by enabling the vehicle to track and follow a pre-designed trajectory. The unicycle model then carries out the control commands sent by the low-level controller and updates its status in real time, closing the feedback loop. This hierarchical structure ensures that the high-level agent focuses on strategic planning while the low-level controller handles precise execution, maintaining safety and efficiency.

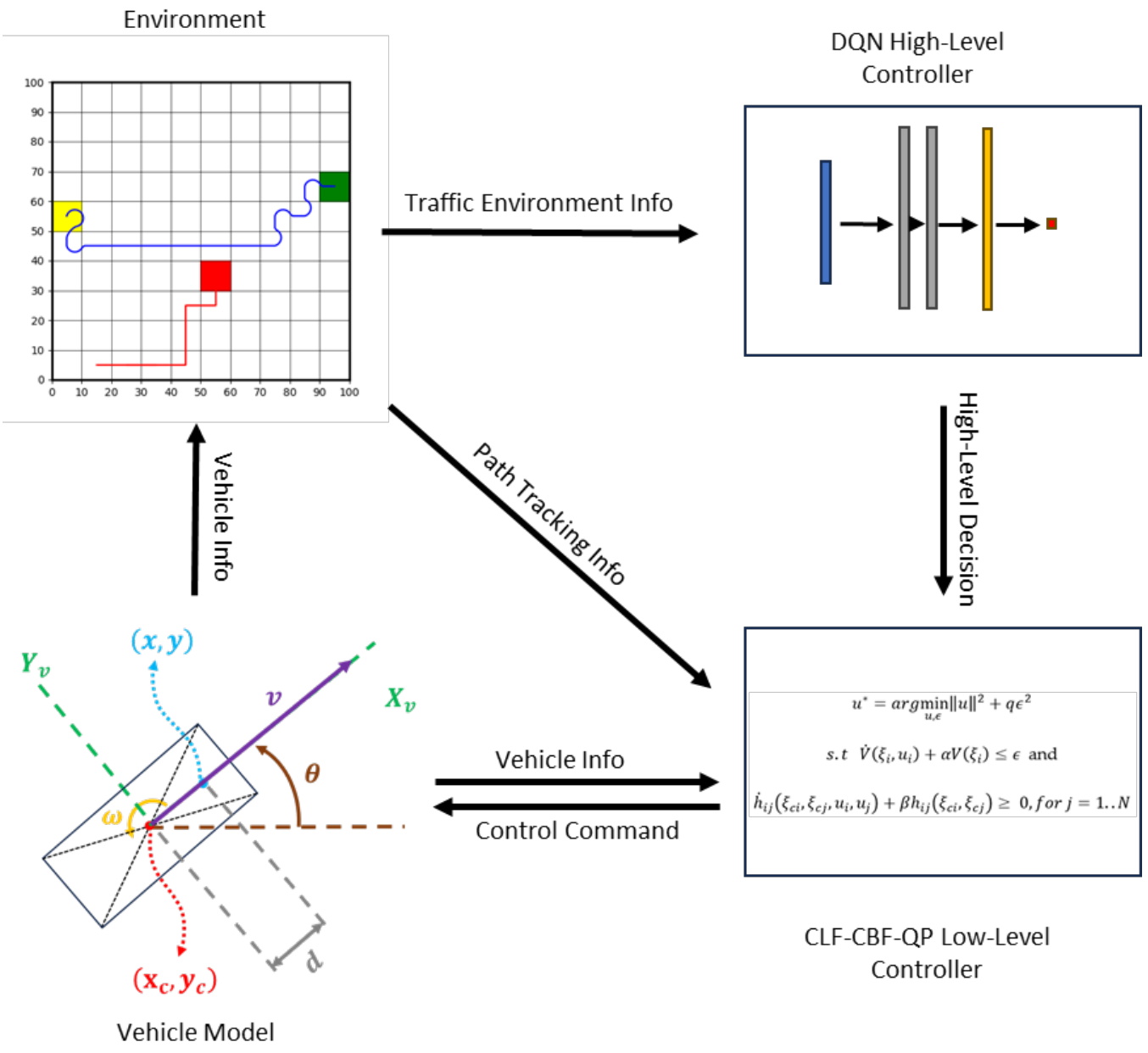

**Figure 4.14.** Proposed DRL and CLF-CBF-QP Based Hybrid Control Flow Chart

Overall, the proposed DRL high-level controller demonstrates its capability to find collision-free and optimal paths which can be further improved by increasing training episodes and fine tuning the hyperparameter.

The demo video for the simulation result using unicycle vehicle model is provided at: CLF-CBF for static obstacle, CLF-CBF for dynamic obstacle, DRL high-level

### 4.3.2 Vehicle Dynamic Model Testing Results

In this section, we present simulation results of the proposed hierarchical control framework using the vehicle dynamic model to evaluate its effectiveness and robustness. The results are organized into two parts to demonstrate both the individual performance of the low-level HOCLF-HOCBF-QP based controller and the overall system performance after combining with the high-level DDQN based decision making agent. Table 4.3 displays the values of the parameters used in the simulations.

**Table 4.3.** Value of parameters used in simulation.

| Symbol | Parameter | Value |
|---|---|---|
| $V$ | Center-of-gravity (CG) velocity | 5 m/s |
| m | Mass | 3000 kg |
| $I_z$ | Yaw moment of inertia | 5.113e3 kg*m^2 |
| $C_f$ | Front tire cornering stiffness | 3e5 N/rad |
| $C_r$ | Rear tire cornering stiffness | 3e5 N/rad |
| $l_f$ | Distance between CG and front axle | 2 m |
| $l_r$ | Distance between CG and rear axle | 2 m |
| $\delta_{fmax}$ | Maximum allowed front wheel steer angle | 0.7 rad |
| $\delta_{fmin}$ | Minimum allowed front wheel steer angle | -0.7 rad |

To prove the effectiveness of the proposed low-level controller, we conduct simulations to evaluate the performance of the HOCLF-HOCBF-QP controller in both path-tracking and collision avoidance. We first test the HOCLF-QP controller's tracking performance using a reference path tracking test case. In this test case, the vehicle is required to travel from a starting point to a destination while following a predefined reference trajectory. The results show that the proposed HOCLF design allows the vehicle to smoothly and accurately follow the predefined path in obstacle-free environments.

Then, we incorporate HOCBF constraints to the controller and evaluate the controller's collision avoidance capability. An obstacle is deliberately placed near the reference path to test whether the controller can successfully avoid potential collisions by temporarily deviating from the planned trajectory. Simulation results indicate successful navigation around the obstacle and smooth return to the reference path once the obstacle is safely passed. In addition, we also evaluate the controller capability to perform reference point tracking. In this test case, the vehicle starts from an arbitrary starting point and is required to reach a target reference point while avoiding surrounding obstacles. The simulation results demonstrate that the proposed HOCLF-HOCBF-QP controller can effectively perform path planning and ensure real-time safety in complex dynamic environments with multiple obstacles.

Figure 4.15 shows the tracking performance of the HOCLF-QP controller on a predefined single lane change reference trajectory. The red dashed line represents the desired reference path of the single lane change maneuver, while the blue solid line illustrates the actual trajectory of the ego vehicle under the HOCLF-QP control. As observed in the figure, the tracking performance is highly accurate throughout the entire path. The controller successfully leads the vehicle to follow the reference path with minimal lateral deviation, indicating the effectiveness of the HOCLF formulation in ensuring stability of the system. This result validates the controller's ability to serve as a reliable low-level trajectory tracking controller for use in the proposed hierarchical control framework.

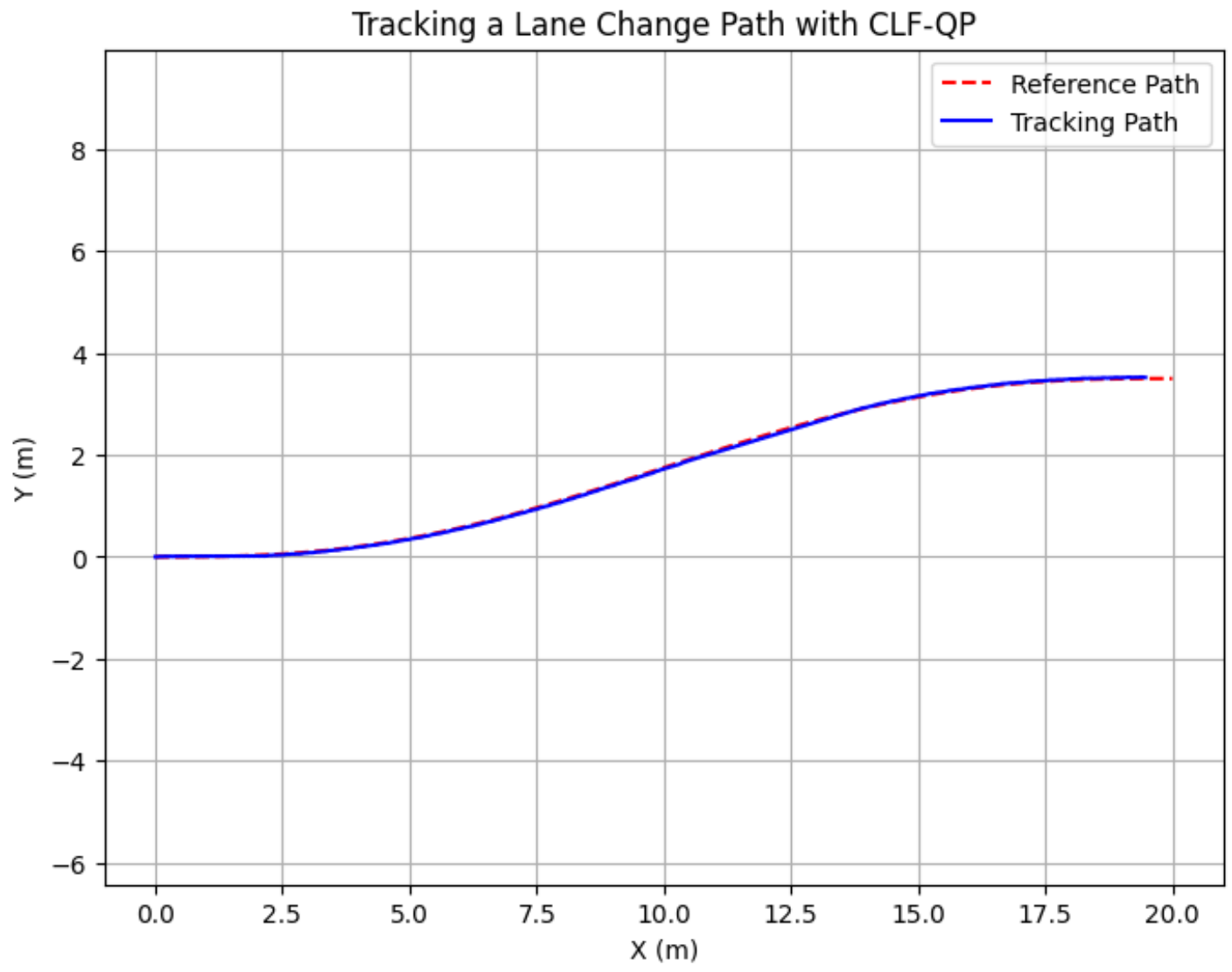

**Figure 4.15.** CLF Path Tracking Controller Performance.

Figure 4.16 shows the trajectory tracking result of the HOCLF-HOCBF-QP controller in the presence of a static obstacle. The red dashed line represents the original reference path of the single lane change maneuver while the orange circle represents the obstacle that is deliberately added near the predefined path. The blue line represents the actual trajectory of the ego vehicle under the HOCLF-HOCBF-QP control. Compared to the HOCLF-QP controller results, which focused only on trajectory tracking, demonstrated in the previous section, the HOCLF-HOCBF-QP controller successfully leads the vehicle to change its trajectory to avoid potential collision with obstacles while still tracking the desired path after passing the obstacle. The safety constraint is enforced by HOCBF, which ensures that the system state remains within a safe set even when the original reference would result in a potential collision. The deviation from the reference path is observed near the obstacle, which is an intentional and necessary result of the CBF-based safety intervention. Once the vehicle passes the obstacle, it smoothly returns to its reference trajectory, demonstrating the controller's ability to balance safety and stability.

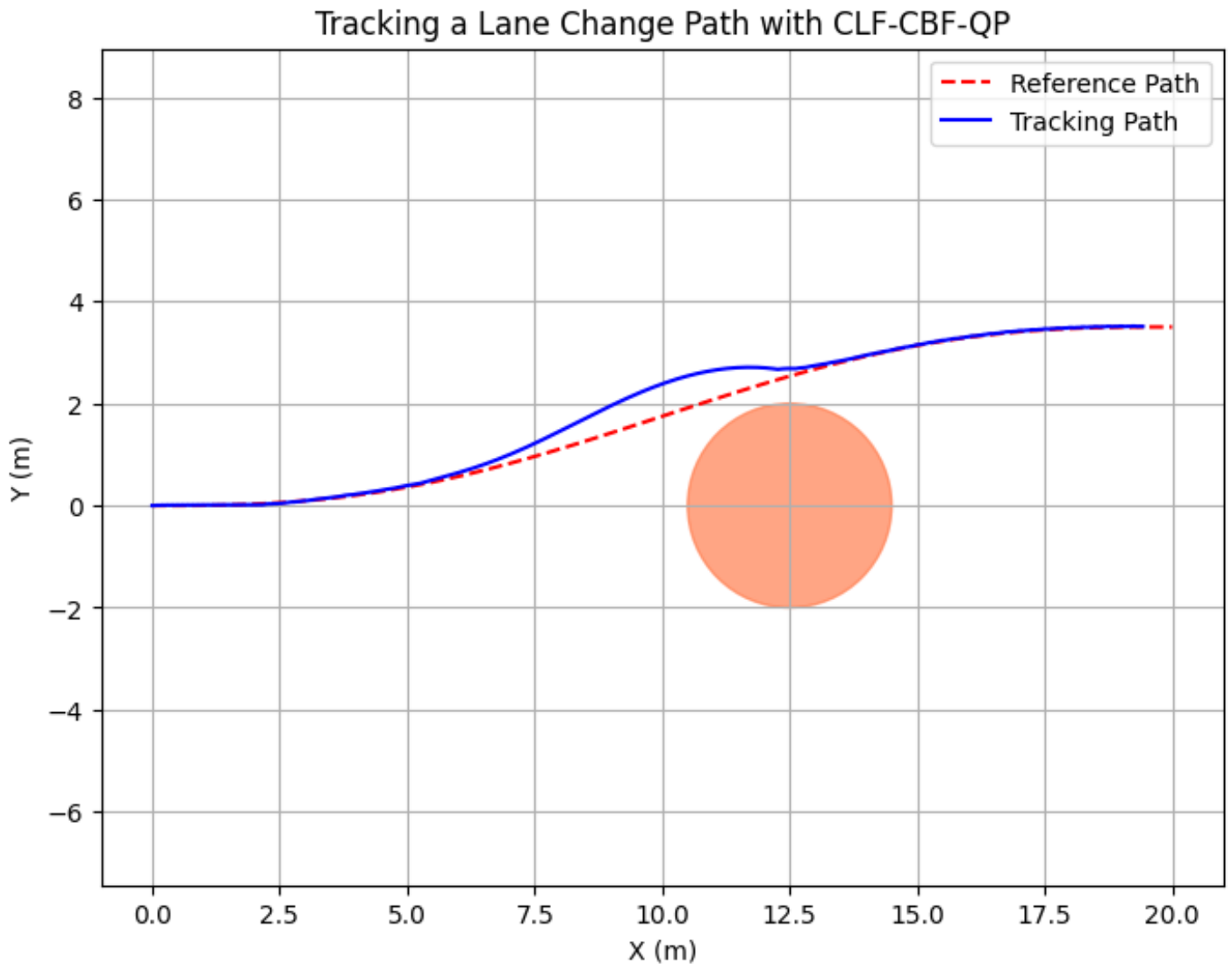

**Figure 4.16.** CLF-CBF based Autonomous Driving Controller for Static Obstacle, Path Tracking.

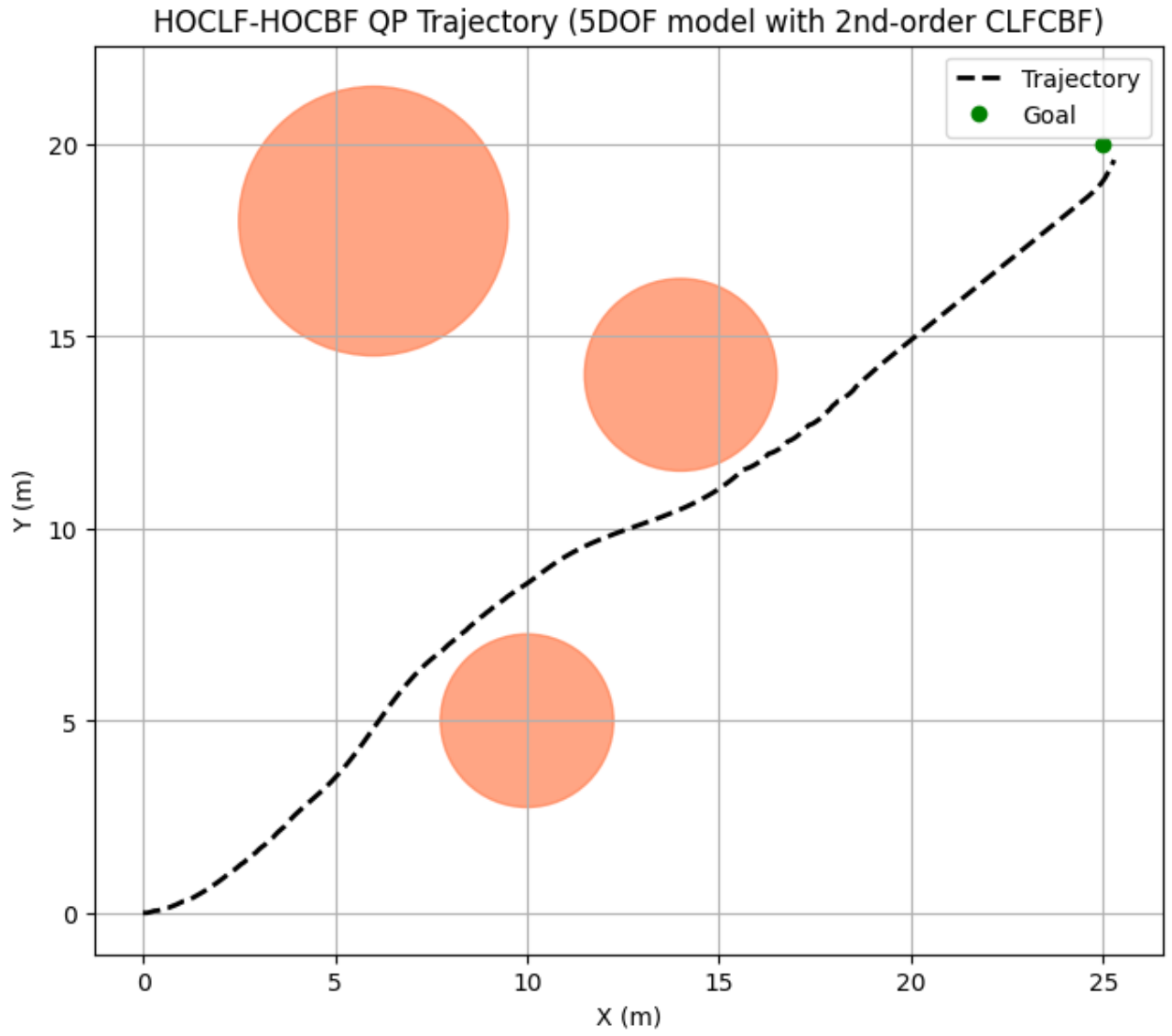

**Figure 4.17.** CLF-CBF based Autonomous Driving Controller for Static Obstacle, Reference Point Tracking.

Figure 4.17 illustrates the reference point tracking performance of the HOCLF-HOCBF-QP controller in a complex environment with multiple static obstacles. The vehicle starts from the origin and aims to reach a predefined destination (represented by the green marker), while avoiding collisions with the circular obstacles (represented by the orange circles). The dashed line shows the actual trajectory of the ego vehicle under the HOCLF-HOCBF-QP control, which demonstrates the controller's ability to balance target point tracking and obstacle avoidance. Notably, the trajectory deviates smoothly around all three obstacles, indicating that the HOCBF constraints are effectively preventing the system from entering unsafe sets. This test case proves the effectiveness of the HOCLF-HOCBF-QP controller in reference point tracking and collision avoidance tasks, particularly in complex environments with multiple obstacles. This result further validates the controller's ability to serve as a reliable low-level trajectory tracking module in the proposed hierarchical control framework.

To further evaluate the low-level HOCLF-HOCBF-QP controller's collision avoidance capability, we tested its performance in a traffic scenario involving a dynamic obstacle moving along a predefined singe lane changing trajectory. Figure 4.17 shows the trajectory tracking result of the HOCLF-HOCBF-QP controller in the presence of a dynamic obstacle. The ego vehicle successfully tracked the reference path before and after the interaction with obstacles, while performing a clear collision avoidance maneuver when approaching the obstacle. During the avoidance process, the vehicle deviated from the reference path to maintain safety but smoothly rejoined the original trajectory once the obstacle was passed.

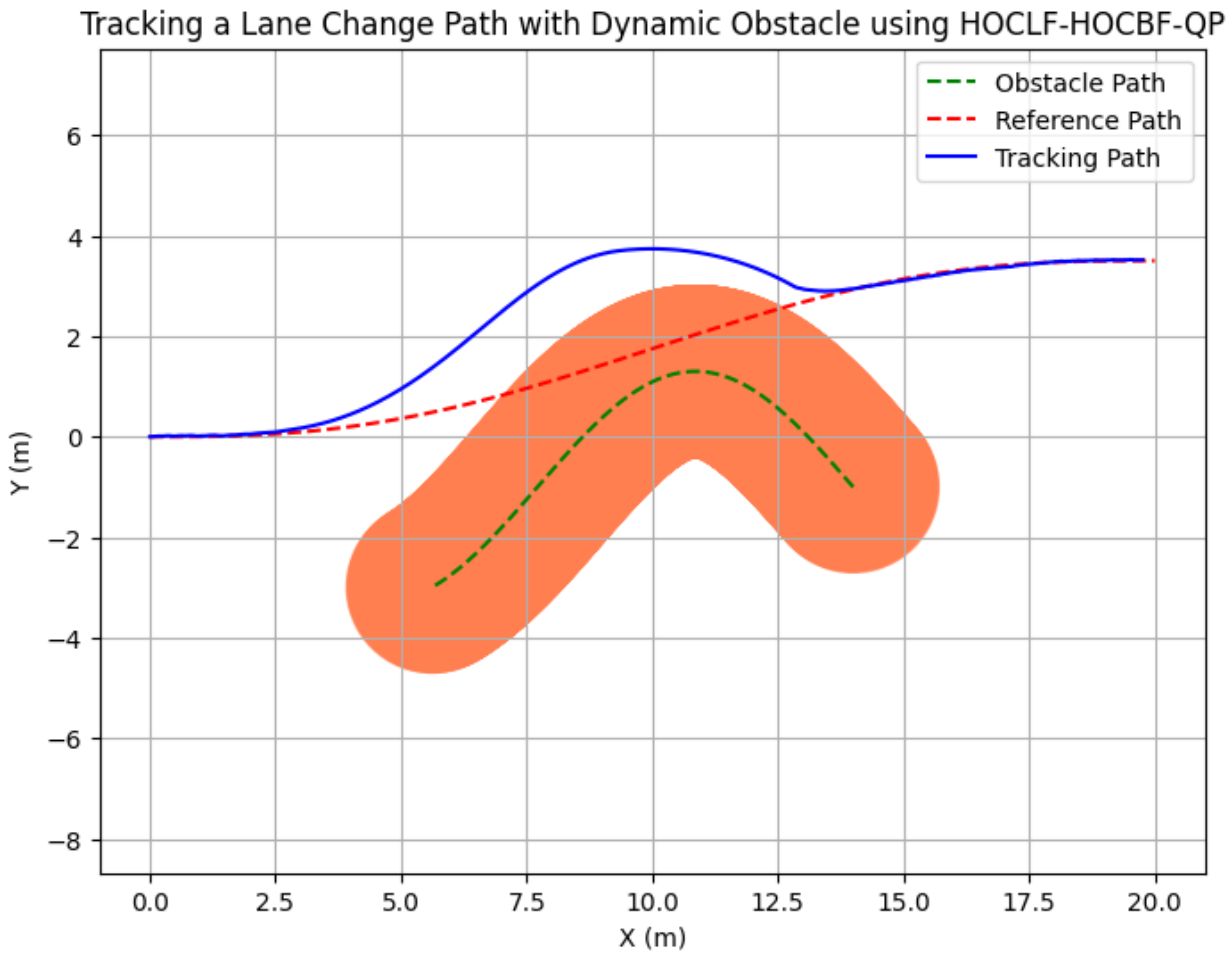

**Figure 4.18.** CLF-CBF based Autonomous Driving Controller for Dynamic Obstacle, Path Tracking.

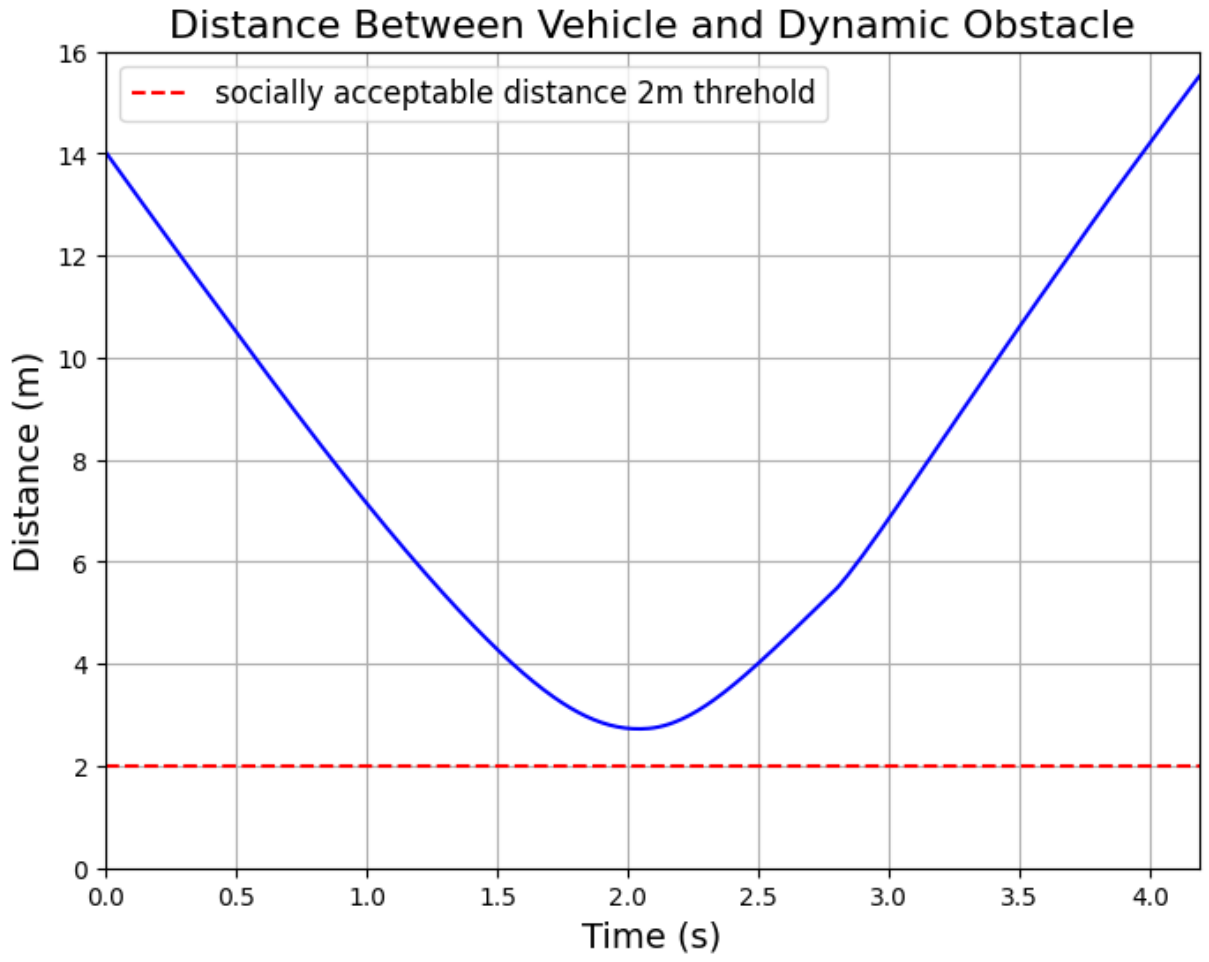

**Figure 4.19.** Real Time Distance Between Vehicle and Dynamic Obstacles

Figure 4.18 illustrates the real time distance between the vehicle and the dynamic obstacle over time. The minimum distance occurs at around 2.0 seconds, where the vehicle and the obstacle are at their closest. The minimum distance remains above the predefined safety threshold of 2 meters, which is generally considered a socially acceptable minimum safe distance between vehicles and surrounding objects (VRUs) in typical driving scenarios.

To further evaluate the low-level HOCLF-HOCBF-QP controller's collision avoidance capability, we use a low-level HOCLF-HOCBF-QP controller to replicate realistic FARS230 bicyclist crash scenarios which describe the traffic scenario where the motorist is trying to merge and overtake a bicyclist. Bicyclists are highly vulnerable road users due to their small size, lower visibility, and their frequent presence in vehicle blind spots. When a vehicle attempts to merge or change lanes without properly accounting for bicyclists, the likelihood of severe collisions increases dramatically.

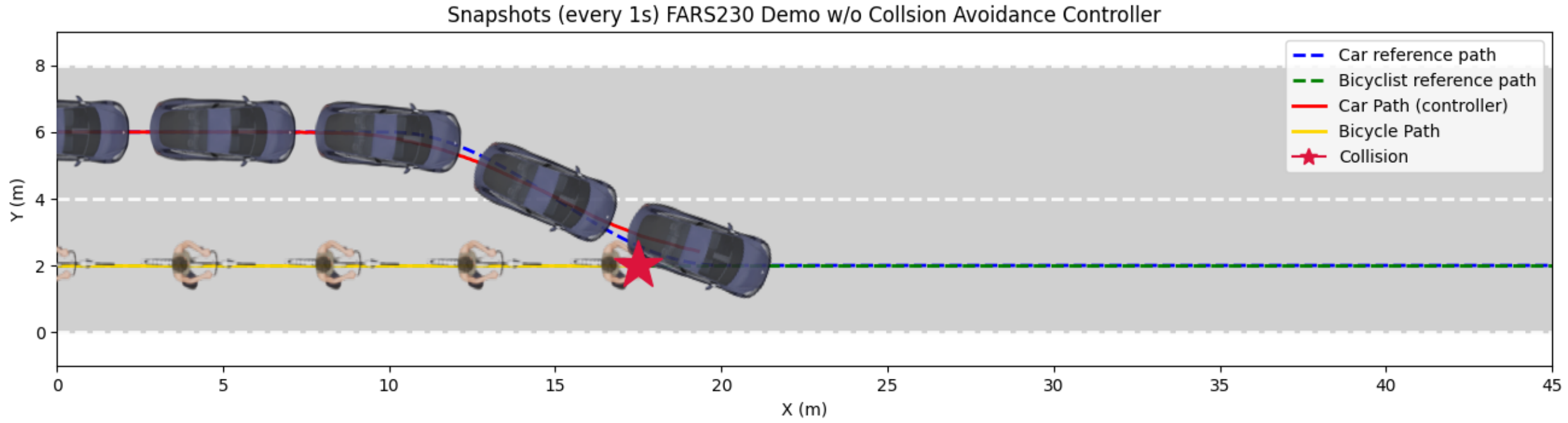

**Figure 4.20.** Snapshot of FARS230 Demo without Collision Avoidance Controller.

Figure 4.20 illustrates a representative scenario where the high-level planner makes an incorrect decision, causing the ego vehicle to merge into a lane occupied by a bicyclist. If the low-level controller is restricted to pure path-following (without any safety constraints), it simply follows the pre-defined trajectory and fails to react to the bicyclist's presence, leading to a collision. In this figure, the green dashed line and yellow solid line represent the bicyclist's reference path and actual trajectory, respectively, while the blue dashed line represents the vehicle's pre-calculated trajectory. It is evident that when the high-level decision-making agent makes an incorrect choice, the vehicle attempts to merge at the wrong position, ultimately resulting in a crash represented by a red star. In this experiment, we use the HOCLF-based path-following controller, which demonstrates excellent tracking performance and can accurately follow the pre-defined trajectory. However, this example highlights a critical limitation: if the low-level controller focuses only on path-following and lacks collision avoidance mechanisms, it will blindly execute the unsafe commands from the high-level planner, even if they lead directly into dangerous situations such as colliding with a bicyclist. This underscores the necessity of integrating HOCBF constraints to enhance the low-level controller's ability to handle dynamic obstacles.

To further investigate this, in the following simulation experiments, we enable the full HOCLF-HOCBF-QP controller, which not only maintains accurate path tracking but also actively adjusts the trajectory when the safety constraints between the vehicle and bicyclist are violated. For example, when the ego vehicle approaches a bicyclist during a lane merge, the controller dynamically modifies the steering input to avoid the collision while minimizing deviation from the intended path. Such behavior demonstrates the controller's capability to ensure safety even under erroneous high-level decisions.

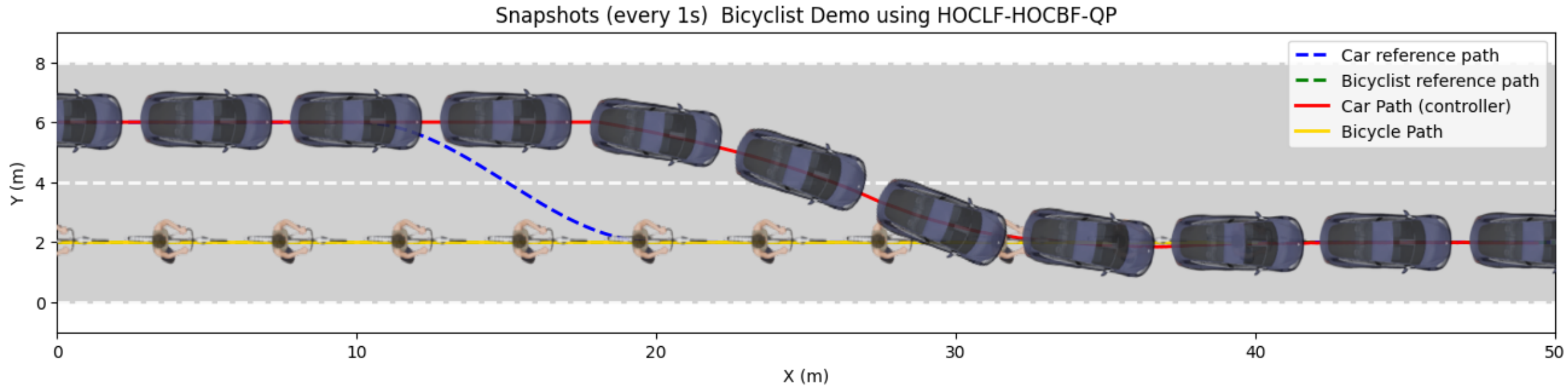

**Figure 4.21.** Snapshot of FARS230 Demo with HOCLF-HOCBF-QP Collision Avoidance Controller

Figure 4.21 demonstrates the behavior of the ego vehicle when the HOCLF-HOCBF-QP low-level controller is enabled, which allows both path-following and collision avoidance. In this scenario, the high-level planner still attempts a lane merger that would lead to a potential collision with the bicyclist. However, unlike in Figure 4.20, the low-level controller now incorporates HOCBF safety constraints, allowing it to dynamically adjust the steering and lateral position to avoid the bicyclist while maintaining close adherence to the original reference path.

In the figure, the blue dashed line and green dashed line represent the reference paths for the vehicle and the bicyclist, respectively. The red solid line shows the actual trajectory of the ego vehicle, which initially follows the reference path but then deviates laterally when approaching the bicyclist to ensure a safe passing distance. The yellow solid line corresponds to the bicyclist's actual motion. The snapshots of the car show how it smoothly changes lanes around the bicyclist and then gradually returns to a safer trajectory without a collision.

This example highlights the effectiveness of integrating HOCLF (for accurate path tracking) with HOCBF (for safety guarantees) in the QP framework. Even when the high-level planner generates suboptimal or unsafe commands, the low-level HOCLF-HOCBF-QP controller autonomously enforces safety while minimizing path deviation, demonstrating its capability to handle real-world scenarios of close proximity interactions with vulnerable road users (VRUs).

Next, we present the overall performance of the proposed hierarchical control framework for vehicle dynamic models, which integrates a high-level DDQN based decision-making agent and a low-level HOCLF-HOCBF-QP based trajectory tracking controller. After validating the effectiveness of the low-level controller in the previous section, we integrate a high-level planner to handle more complex decision-making tasks. To evaluate the overall performance of the hierarchical control framework, we design a simple and complex test case within a multi-lane traffic environment. In both cases, the ego vehicle must travel from a starting point to a predefined destination, navigating through a roadway populated with multiple obstacles. The vehicle is required to dynamically avoid obstacles by performing appropriate lane changes. The test cases are constructed using a highway-environment simulator [100]. The environment settings, vehicle dynamics, and the low-level control strategy are modified to match our problem setup. Unlike traditional trajectory planning setups, no predefined global path is provided. Instead, the DDQN-based high-level agent generates discrete lane-level decisions (e.g., idle meaning stay in lane, left lane change, right lane change) based on observation of the ego vehicle's state and the traffic environment. At the same time, the HOCLF-HOCBF-QP low-level controller ensures that each decision is executed safely and accurately in real time.

In this experiment, the environment state is represented as a 5×5 array that encodes information about the ego vehicle and its surrounding obstacles. Each row corresponds to an entity (either the ego vehicle or an obstacle) and includes the following attributes: [presence, $x$, $y$, $v_x$, $v_y$], where presence is a binary indicator of whether the object exists in the current frame. The action space is discrete and consists of three possible maneuvers: idle (no lane change), left lane change, and right lane change. The reward function is designed to encourage goal-reaching behavior while penalizing unsafe or inefficient actions. A positive reward of +50 is assigned when the vehicle successfully reaches its destination. A large negative reward of −100 is applied in the event of a collision with any obstacle. Additionally, at each timestep, the agent receives a dense reward proportional to its forward progress ($\Delta x$), calculated as $r_{progress} = c * (x_{current} - x_{prev})$, where $c$ is a positive coefficient, which encourages faster navigation. A small penalty of −0.5 is applied when the agent executes a lane change (either left or right), to discourage unnecessary lateral movement and to promote trajectory stability.

Simulation results from both test cases show that the proposed framework can make the vehicle successfully navigate going from the start point to its destination while effectively avoiding all obstacles using the appropriate lane changing maneuver(s).

We first present the training progress of the high-level DDQN based decision-making agent in the proposed hierarchical control framework. Table 4.4 summarizes the key hyperparameters used in the DDQN training process and Figure 4.22 illustrates the training progress of the DDQN high-level decision-making agent in the three-lane complex autonomous driving environment. A low-pass filter is applied to smooth both the episode reward and step count curves for better interpretability. Initially, the agent exhibits poor performance due to high probability of random explorations, with total reward remaining negative and step count relatively low, indicating frequent collisions and early episode terminations. After approximately 2500 episodes, a significant improvement is observed: the total reward begins to rise rapidly, and the average number of steps per episode increases concurrently. This trend suggests that the agent gradually learns an effective lane-changing strategy to avoid obstacles and to extend its episode longevity. After about 3500 episodes, both rewards and steps per episode are maintained at a relatively high level, indicating convergence to a stable policy. Some fluctuations are still present in the reward curve, likely due to exploration behavior or occasional difficult scenarios.

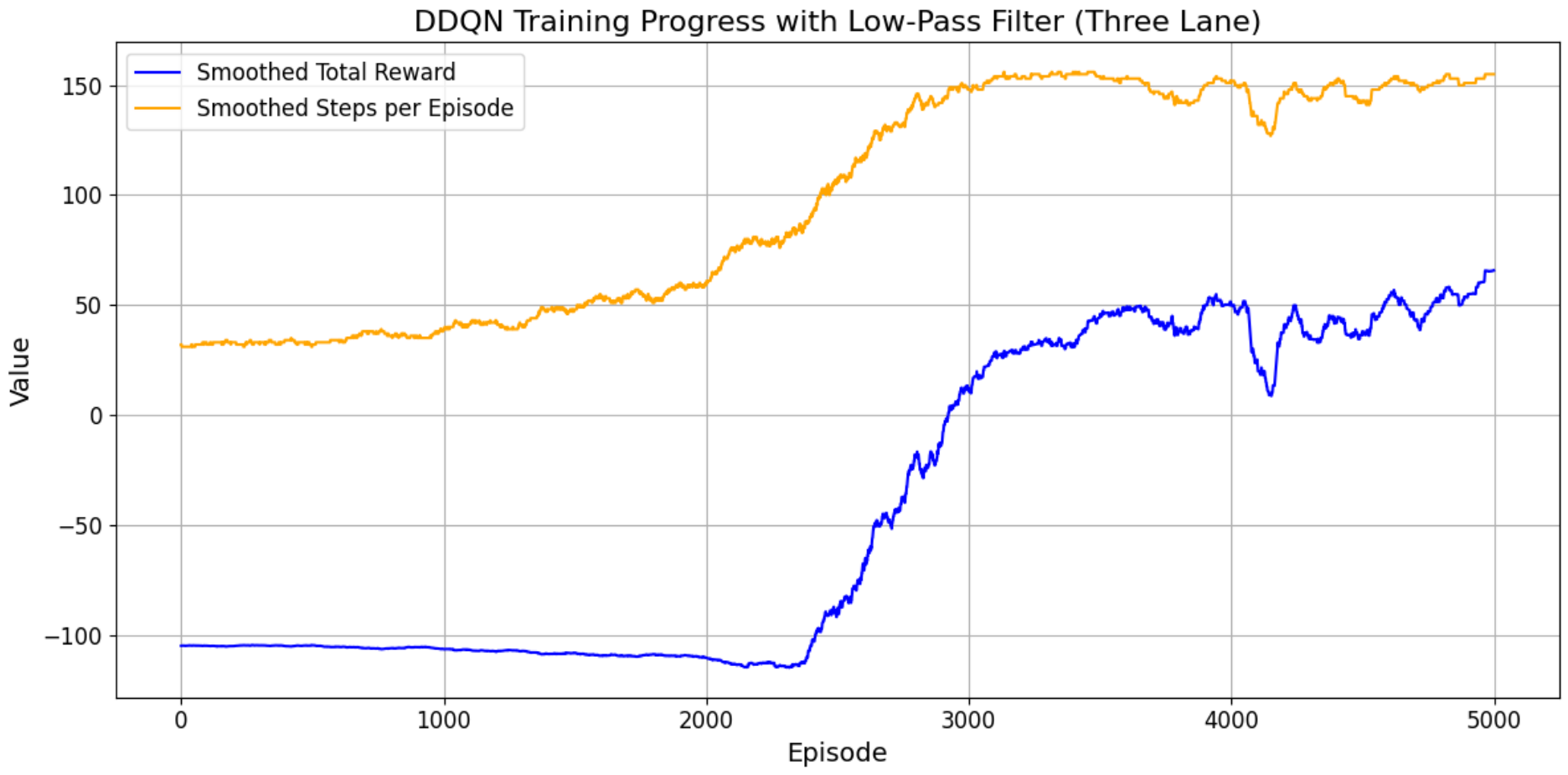

**Figure 4.22.** Deep reinforcement learning training progress.

**Table 4.4.** Hyperparameters and training settings.

| Hyperparameter | Value |
|---|---|
| Replay buffer size | 100,000 |
| Mini-batch size | 64 |
| Learning rate | 0.001 |
| Target network update frequency | Every 100 steps |
| Exploration strategy (ε decay) | 1.0 → 0.05 over 200,000 steps |
| Discount factor ($\gamma$) | 0.99 |
| Optimizer | Adam |
| Loss function | Mean Squared Error (MSE) |

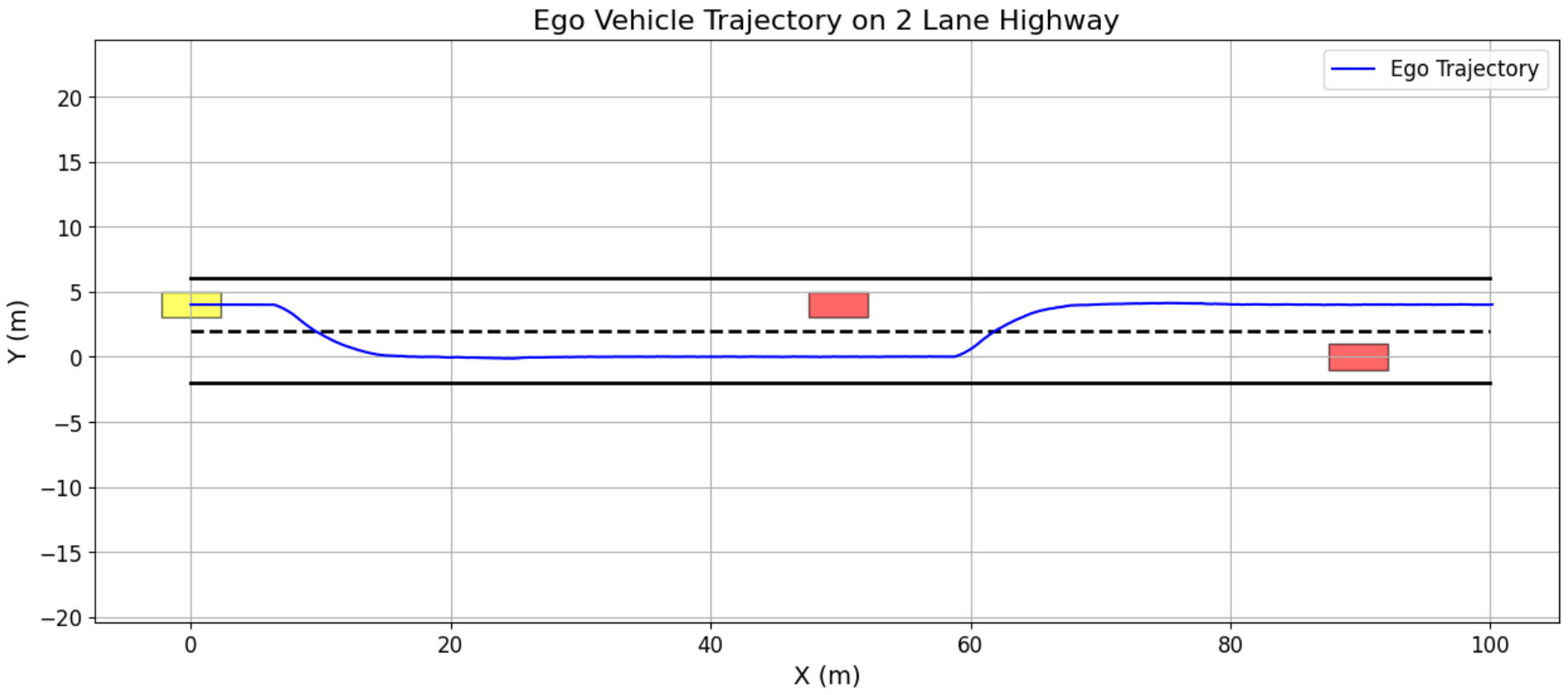

**Figure 4.23.** Hierarchical controller overall performance in simple test case

Figure 4.23 illustrates the ego vehicle's trajectory in a two-lane highway test case using the proposed hierarchical controller. The road is divided into two lanes with clearly marked boundaries and centerline. The ego vehicle (represented by yellow rectangle) starts in the upper lane and initially follows a straight path before encountering an obstacle (represented by red rectangle) positioned ahead in the same lane. The vehicle performs a lane-change maneuver in front of the obstacle by smoothly transitioning into the lower lane. After passing the obstacle, the vehicle performs another lane-change maneuver and returns to the upper lane, to avoid collision with the second obstacle.

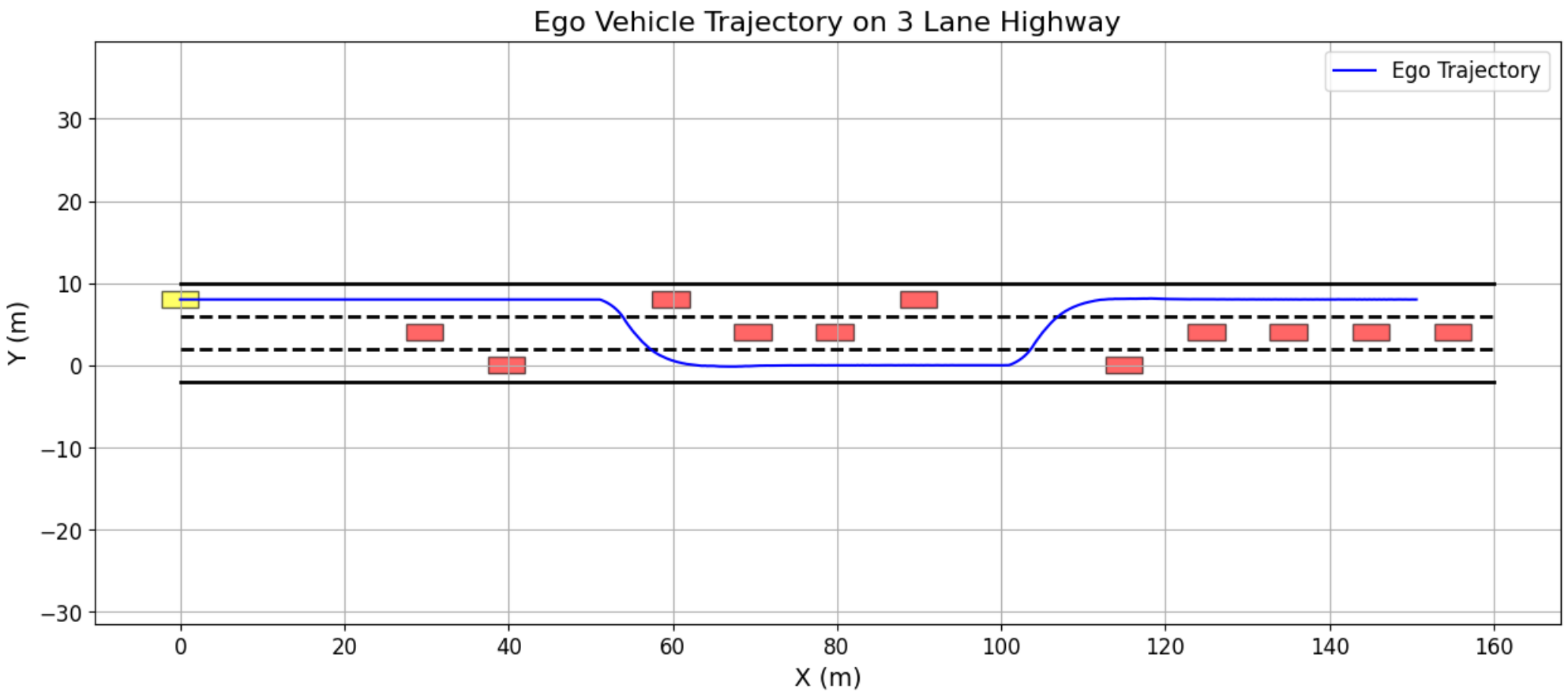

**Figure 4.24.** Hierarchical controller overall performance in complex test case

Figure 4.24 illustrates the ego vehicle's trajectory in a complex three-lane highway test case using the proposed hierarchical controller. From the plot, we notice that this environment is much more complex compared to the simple two-lane test case demonstrated before. The ego vehicle (represented by the yellow rectangle) starts in the upper lane and initially follows a straight path before encountering an obstacle (represented by the red rectangle) positioned ahead in the same lane. The vehicle performs two consecutive lane-change maneuvers to transition smoothly into the lower lane in response to a series of obstacles. After passing the obstacles, the vehicle performs another two consecutive lane-change maneuvers and returns to the upper lane, to avoid collision with other obstacles.

The two trajectory plots in Figures 4.23 and 4.24 demonstrate the effectiveness of the proposed hierarchical control system. The high-level DDQN agent can correctly generate lane-change decisions based on obstacle positions, while the low-level HOCLF-HOCBF-QP controller ensures smooth and safe high-level decision execution. The trajectory remains continuous and collision-free throughout the simulation, indicating successful integration of decision-making and control components. Moreover, the vehicle consistently follows a straight path in obstacle-free regions, without performing unnecessary lane-change behaviors. This suggests that the penalty design for lane changes successfully discourages unnecessary lane-changing actions, which makes the decision-making process more efficient.

To evaluate the computational efficiency of the proposed hierarchical control framework, we conducted computational cost tests based on the aforementioned simulation test case. In complex test case settings, the low-level control simulation frequency is 100 Hz and high-level policy simulation frequency is 5 Hz. The low-level HOCLF-HOCBF-QP controller requires an average solve time of approximately 0.66 ms per step using the Gurobi optimizer while the high-level DDQN decision-making agent requires an average computational time of approximately 0.60 ms per step. These tests were performed in a Google Colab CPU environment which is using a single-threaded Intel Xeon processor without GPU acceleration. Even under this relatively limited computational setting, the solve times accounts for only 6.6% of the low-level control cycle (10 ms) and 0.3% of the high-level decision-making cycle (200 ms), respectively. These results demonstrate that the proposed framework has good real-time capability. While it is expected that the computational time, particularly for the HOCLF-HOCBF-QP controller, will increase in more complex traffic scenarios with higher numbers of obstacles and constraints, our current results suggest that the proposed framework still provides sufficient real-time capability for typical autonomous driving tasks. In future work, we plan to further evaluate its effectiveness and real time capability by conducting tests using the Hardware-in-the-Loop (HIL) approach and Vehicle-in-Virtual-Environment (VVE) approach, to ensure its real-world feasibility and practical applicability in different driving scenarios.

The demo video for the simulation result using the vehicle dynamics model is provided at: Two Lane Test Case Demo, Three Lane Test Case Demo, FARS230 Demo with HOCLF-HOCBF-QP Controller

## 4.4 Conclusion

Path planning and collision avoidance are critical challenges in the development of reliable autonomous driving systems, particularly in dynamic traffic environments with obstacles. Traditional rule-based planners often struggle to handle such complexities. To address these limitations, we proposed a hierarchical decision-making and control framework that enables autonomous vehicles to automatically avoid obstacles through lane-changing maneuvers. The proposed system integrates a high-level DDQN based decision-making agent with a low-level CLF-CBF-QP-based controller. The high-level DRL agent generates discrete high-level decisions based on ego vehicle's information and traffic environment information, while the low-level CLF-CBF-QP controller ensures safe and efficient high-level decision refinement and execution. To comprehensively evaluate the proposed hierarchical collision avoidance framework, we implemented it on two distinct vehicle models. Initial feasibility was verified using a simplified unicycle model, followed by a deployment on a more realistic vehicle dynamics model. Simulation results validated the effectiveness of the proposed hierarchical control framework in autonomous driving tasks.

Furthermore, the implementation of our framework is highly modular and flexible, allowing it to adapt to different vehicle models and driving environments. Both the high-level agent and the low-level controller can be customized to incorporate task-specific constraints, system dynamics, and safety requirements. This adaptability endows the framework with strong generalizability and makes it suitable for a wide range of autonomous driving applications.

Compared to traditional optimization-based methods such as model predictive control (MPC), the proposed framework significantly simplifies the online optimization process. Traditional approaches typically solve large-scale optimal control problems at each step, which introduces high computational complexity and often requires accurate environmental modeling. Also, these approaches sometimes may struggle to find optimal paths in complex traffic conditions. In contrast, our approach takes advantage of the high-level DRL based decision-making agent and further enhances the system's capability to navigate various complex traffic environments. In addition, compared to traditional reinforcement learning methods, including end-to-end approaches, the key advantage of our proposed framework lies in the explicit integration of low-level controller which ensures hard-coded safety rules. Traditional RL methods rely entirely on reward design and learned policies to avoid collisions, which can still lead to unsafe behaviors during training or in unforeseen situations, as they lack hard safety guarantees. In contrast, our approach separates decision-making and low-level control. The high-level DDQN agent focuses on discrete lane-level maneuver selection, leveraging the adaptability and learning capability of RL to handle diverse traffic conditions. Meanwhile, the low-level HOCLF-HOCBF-QP controller ensures safety by introducing the CBF constraint, providing a hard-coded safety layer.

There are still some limitations in this study. First, the two vehicle models used in this work, despite representing different levels of modeling fidelity, are still relatively simple. The unicycle model provides a basic abstraction for preliminary validation, while the vehicle dynamics model introduces more realism but remains a simplified representation compared to full-scale commercial vehicle systems, which limits the realism of driving behavior. Future work will consider integrating a longitudinal dynamic model to capture the vehicle's longitudinal motions, hence letting the model become a full vehicle dynamic model. Second, although DQN and DDQN perform well in the current setup, more advanced DRL algorithms such as SAC or DDPG could potentially improve learning efficiency and policy robustness in more complex or uncertain environments and their use will also be investigated in future work. Finally, we plan to conduct more detailed sensitivity analyses and ablation studies to further understand the impact of key parameters and the contribution of each module, especially in more complex dynamic environments with multiple road users.

# Chapter 5: Vehicle-in-Virtual-Environment (VVE) Based Testing Pipeline

## 5.1 Introduction

Traditional testing pipelines, such as Model-in-the-Loop (MIL) [101] and Hardware-in-the-Loop (HIL) [102-103] simulations followed by deployment on public roads have significant limitations in terms of safety, cost, and efficiency. These methods often require exposing other road users to autonomous vehicles equipped with unverified or experimental driving functions, raising substantial safety and ethical concerns. Moreover, the dependence on physical road testing is both expensive and time-consuming, which can slow down the iterative development and refinement of autonomous driving technologies.

To address these challenges, a novel approach called Vehicle-in-Virtual-Environment (VVE) was proposed [46], [104]. The VVE method integrates real vehicles into highly realistic virtual environments, enabling comprehensive and resource-efficient testing without the need for public road exposure. This approach not only enhances safety by eliminating risks to other road users but also significantly reduces testing costs and accelerates the development process. Additionally, VVE is particularly well-suited for training and evaluating Deep Reinforcement Learning (DRL) based autonomous driving agents. By providing a controlled yet dynamic simulation environment, VVE allows DRL agents to interact with diverse traffic scenarios and adapt to complex, real-world-like conditions safely and efficiently. This integration facilitates the thorough validation of autonomous driving systems, ensuring robust performance and safety before any public road deployment. Consequently, the VVE method offers a safer, more efficient, and cost-effective solution for advancing autonomous driving technology.

By integrating traditional methods with the novel VVE method, this section proposes a novel testing pipeline for evaluating autonomous driving decision-making and control algorithms. This comprehensive testing pipeline integrates Model-in-the-Loop (MIL), Hardware-in-the-Loop (HIL), Vehicle-in-Virtual-Environment (VVE) testing, and public road testing to thoroughly validate the algorithms. Initially, the algorithm is developed and tested using the MIL approach. A comprehensive and detailed vehicle dynamics model is created using Simulink, incorporating traditional longitudinal and lateral dynamics, tire rotation, and tire force models. By adjusting parameters, this model can simulate the vehicle's dynamics with high accuracy, allowing for effective preliminary development and testing. Moreover, deep learning or reinforcement learning based autonomous driving algorithms can also be trained and validated using this MIL testing method. Subsequently, the developed algorithm is evaluated under HIL settings. HIL testing integrates real hardware components with the simulation environment, enabling us to simulate signal transmission delays and eliminate unrealistic extreme inputs, such as abrupt steering changes from left to right. This step brings the algorithm closer to real-world conditions by accounting for hardware-related factors that can affect performance. The VVE testing method is then employed to further evaluate the algorithm. The VVE method synchronizes the real-world motions of an actual vehicle with its virtual counterpart in a simulated environment, allowing for the creation of various virtual traffic scenarios for safe and resource-efficient testing. Because real vehicles and pedestrians participate in the simulation process, the dynamics captured are more authentic, enhancing the credibility of the simulation results. VVE significantly reduces testing costs and time, improving testing efficiency by enabling extensive scenario testing without the risks associated with public road exposure. In addition, the VVE method possesses multi-actor capabilities, enabling it to be applied to various scenarios such as Vehicle-to-Pedestrian (V2P) tests [105]. After successfully passing through MIL, HIL, and VVE testing phases, the autonomous driving algorithm attains the capability to operate in the real world with a higher degree of confidence. This rigorous testing pipeline ensures that the algorithm is robust and ready for the final phase of validation through public road testing.

## 5.2 Methodology

### 5.2.1 Model-in-Loop and Vehicle Dynamic Model

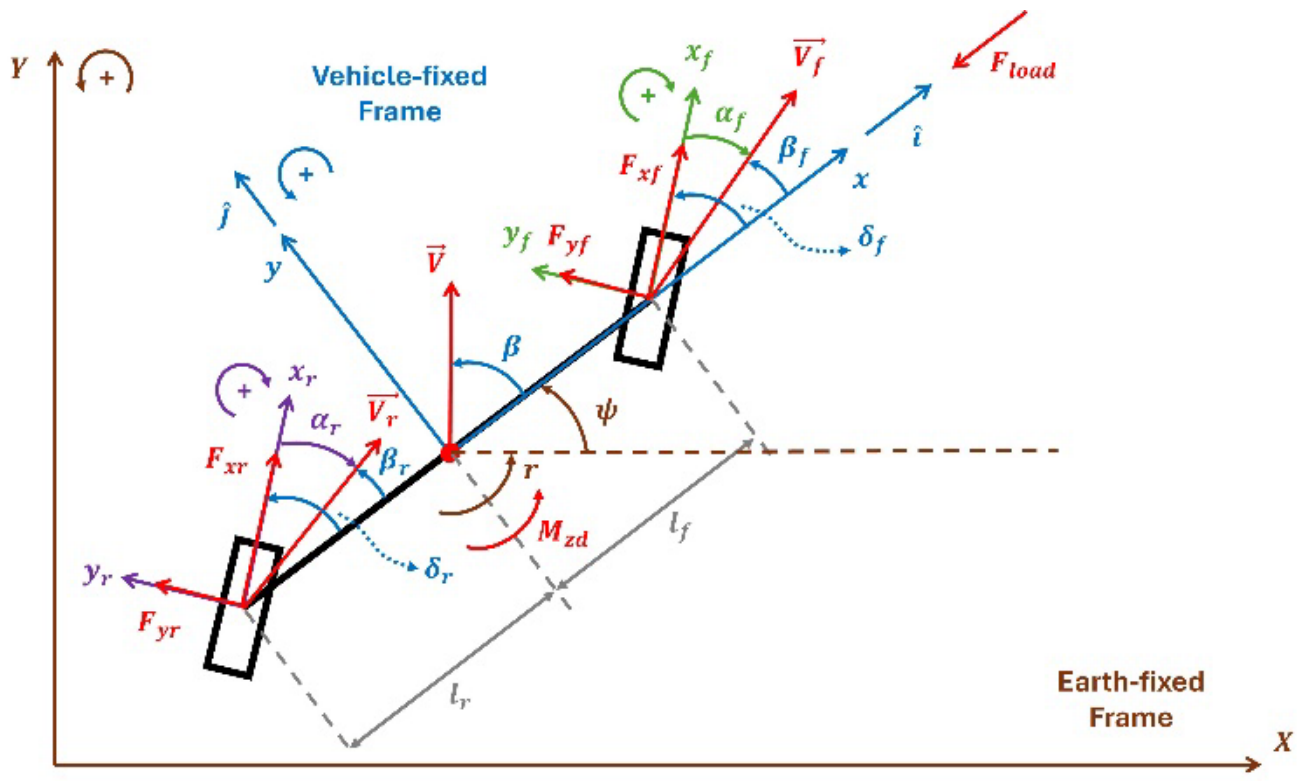

**Figure 5.1.** Extended lateral dynamic model [106]

The first stage of the proposed testing process is Model-in-the-Loop (MIL) study, where a simulation environment is used to develop a suitable ADAS system for the proposed use case. This would require a vehicle model that can represent real vehicle behaviors in a reasonably accurate manner. In this section, an extended single-track vehicle model that captures both the longitudinal and lateral dynamic behaviors is introduced. In general, the proposed vehicle model contains three major components: an extended single-track lateral model to capture the longitudinal and lateral dynamics of the vehicle, a Modified Dugoff coupled tire model to provide the tire forces that feed into the single-track model and a wheel rotation model to aid the calculation of tire forces.

The overall model configuration for the extended lateral model is illustrated in Figure 5.1, which is adapted from [33]. Table 5.1 lists the parameters used in this model. It can be observed that this model is an extension of the simple single-track lateral model that only makes use of lateral tire forces. Instead, this extended model includes both longitudinal and lateral tire models as well as longitudinal road loads to account for the vehicle dynamic behaviors in both longitudinal and lateral directions.

**Table 5.1.** Extended lateral dynamic model parameters.

| Parameters | Descriptions |
|---|---|
| $m$ | Vehicle mass [kg] |
| $l_f$ | Distance between front axle & CG [m] |
| $l_r$ | Distance between rear axle & CG [m] |
| $I_z$ | Vehicle yaw moment of inertia |
| $V, V_f, V_r$ | Vehicle CG, front & rear axle velocity [m/sec] |
| $\beta, \beta_f, \beta_r$ | Vehicle CG, front & rear axle side slip angle [rad] |
| $\delta_f, \delta_r$ | Front & rear steer angle [rad] |
| $\alpha_f, \alpha_r$ | Front & rear tire side slip angle [rad] |
| $F_{xf}, F_{xr}$ | Front & rear tire longitudinal force [N] |
| $F_{yf}, F_{yr}$ | Front & rear tire lateral force [N] |
| $\psi$ | Vehicle yaw angle [rad] |
| $r$ | Vehicle yaw rate [rad/sec] |
| $M_{zd}$ | Vehicle yaw moment disturbance [Nm] |
| $F_{load}$ | Longitudinal road load [N] |

From Figure 5.1, it is possible to write down the dynamic equations of vehicle side slip angle ($\beta$), vehicle speed ($V$) and vehicle yaw rate ($r$) as described in Equation (5.1).

$$\begin{bmatrix} \dot{\beta} \\ \dot{V} \\ \dot{r} \end{bmatrix} = \begin{bmatrix} -\frac{\sin(\beta)}{mV} & \frac{\cos(\beta)}{mV} & 0 \\ \frac{\cos(\beta)}{m} & \frac{\sin(\beta)}{m} & 0 \\ 0 & 0 & \frac{1}{I_z} \end{bmatrix} \begin{bmatrix} \Sigma F_x \\ \Sigma F_y \\ \Sigma M_z \end{bmatrix} - \begin{bmatrix} r \\ 0 \\ 0 \end{bmatrix} \tag{5.1}$$

Accounting for the resultant forces and moments in the vehicle-fixed frame based on the slightly simplified Figure 5.2, one can write Equation (5.2).

$$\begin{bmatrix} \Sigma F_x \\ \Sigma F_y \\ \Sigma M_z \end{bmatrix} = \begin{bmatrix} \cos(\delta_f) & \cos(\delta_r) & -\sin(\delta_f) & -\sin(\delta_r) \\ \sin(\delta_f) & \sin(\delta_r) & \cos(\delta_f) & \cos(\delta_r) \\ l_f \sin(\delta_f) & -l_r \sin(\delta_r) & l_f \cos(\delta_f) & -l_r \cos(\delta_r) \end{bmatrix} \begin{bmatrix} F_{xf} \\ F_{xr} \\ F_{yf} \\ F_{yr} \end{bmatrix} + \begin{bmatrix} -F_{load} \\ 0 \\ M_{zd} \end{bmatrix} \tag{5.2}$$

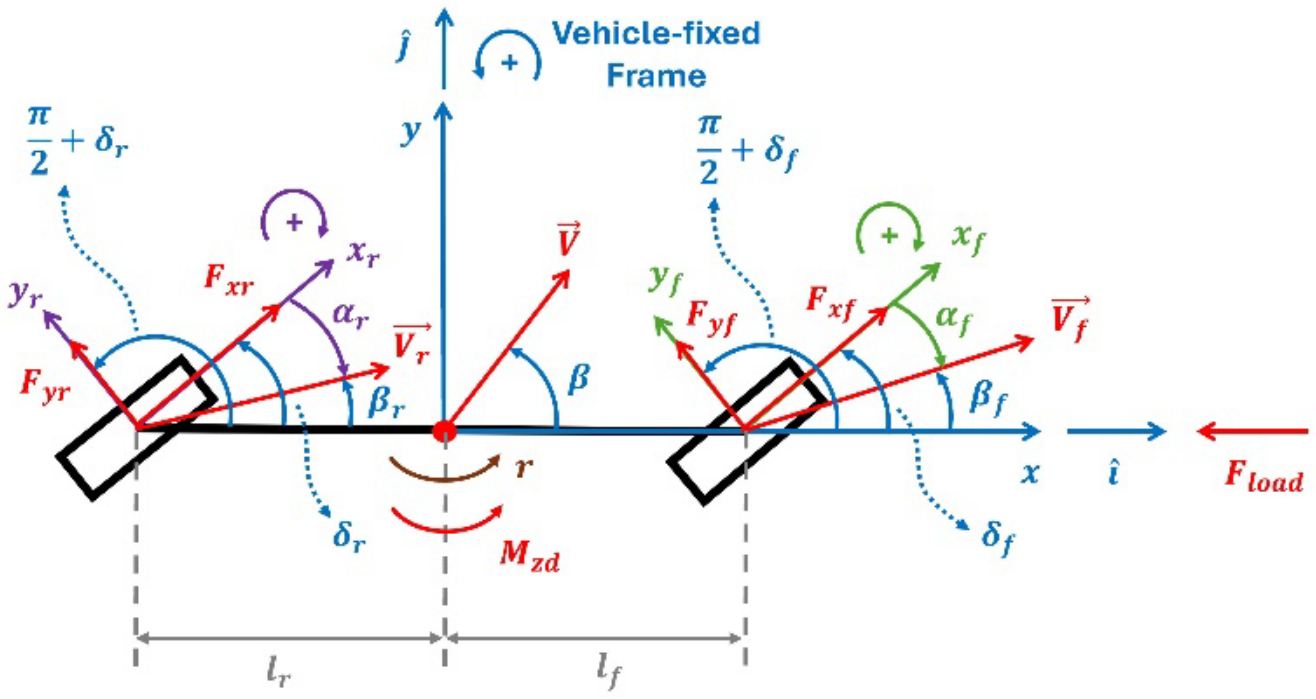

**Figure 5.2.** Extended lateral dynamic model represented in vehicle-fixed frame

Combining Equation (5.1) and Equation (5.2), one can arrive at the extended model equations of motion (EOM) as illustrated in Equation (5.3). It should be remarked that this model is essentially a nonlinear longitudinal and lateral single-track dynamic model. Also note that in this model, $(\beta, V, r)$ are regarded as system states, $(\delta_f, \delta_r, M_{zd})$ are treated as system inputs, and tire forces $(F_{xf}, F_{xr}, F_{yf}, F_{yr})$ are to be calculated from the tire model before being fed into this model.

$$\begin{bmatrix}\dot{\beta}\\ \dot{V}\\ \dot{r}\end{bmatrix} = A\begin{bmatrix}F_{xf}\\ F_{xr}\\ F_{yf}\\ F_{yr}\end{bmatrix} + B\begin{bmatrix}-F_{load}\\ 0\\ M_{zd}\end{bmatrix} - \begin{bmatrix}r\\ 0\\ 0\end{bmatrix} \tag{5.3}$$

$$\text{where:}\begin{cases} A = \begin{bmatrix} \frac{\sin(\delta_f-\beta)}{mV} & \frac{\sin(\delta_r-\beta)}{mV} & \frac{\cos(\delta_f-\beta)}{mV} & \frac{\cos(\delta_r-\beta)}{mV} \\ \frac{\cos(\delta_f-\beta)}{m} & \frac{\cos(\delta_r-\beta)}{m} & \frac{-\sin(\delta_f-\beta)}{m} & \frac{-\sin(\delta_r-\beta)}{m} \\ \frac{l_f\sin(\delta_f)}{I_z} & \frac{-l_r\sin(\delta_r)}{I_z} & \frac{l_f\cos(\delta_f)}{I_z} & \frac{-l_r\cos(\delta_r)}{I_z} \end{bmatrix} \\ B = \begin{bmatrix} \frac{-\sin(\beta)}{mV} & \frac{\cos(\beta)}{mV} & 0 \\ \frac{\cos(\beta)}{m} & \frac{\sin(\beta)}{m} & 0 \\ 0 & 0 & \frac{1}{I_z} \end{bmatrix} \end{cases}.$$

In order to provide the above-mentioned single-track model with tire forces, an appropriate tire model is needed. In this section, the Modified Dugoff model is used due to its dependency on only a small number of parameters as well as its capability to capture longitudinal and lateral tire force coupling effects. Equation (5.4) describes the equations for longitudinal and lateral tire forces. Both equations are adapted from [107]. Table 2 details the parameters used in the tire model.

$$\begin{cases} F_x = \frac{C_x s}{1-s} f(Z) g_x \\ F_y = \frac{C_y \tan(\alpha)}{1-s} f(Z) g_y \end{cases} \tag{5.4}$$

$$\text{where:} \begin{cases} Z = \frac{\mu F_z(1-s)}{2\sqrt{(C_x s)^2 + (C_y \tan(\alpha))^2}} \\ f(Z) = \begin{cases} Z(2-Z) \; if \; Z < 1 \\ 1 \; if \; Z \geq 1 \end{cases} \\ g_x = (1.15 - 0.75\mu)s^2 - (1.63 - 0.75\mu)s + 1.5 \\ g_y = (\mu - 1.6)\tan(\alpha) + 1.5 \end{cases}.$$

**Table 5.2.** Modified Dugoff tire model parameters

| Parameters | Descriptions |
|---|---|
| $C_x$ | Longitudinal tire stiffness [N] |
| $C_y$ | Lateral tire stiffness [N/rad] |
| $s$ | Tire longitudinal slip, $s\epsilon[-1,1]$ |
| $\alpha$ | Tire side slip angle [rad] |
| $F_x$ | Longitudinal tire force |
| $F_y$ | Lateral tire force |
| $\mu$ | Road friction coefficient |
| $F_z$ | Tire vertical load [N] |

It should be remarked that the inputs to the tire model are tire longitudinal slip ($s$) and lateral side slip angle ($\alpha$). While the tire side slip angle can be obtained by applying post-processing on the outputs of the extended lateral model mentioned above, the calculation of tire longitudinal slip requires an additional model describing the wheel rotational dynamics. This wheel rotational model is illustrated in Figure 5.3 and its dynamic equations are listed in Equation (5.5) and Equation (5.6). Table 5.3 lists the necessary parameters for this model. Note that the deviated angular velocities of the front and rear tire ($\Delta\omega_f, \Delta\omega_r$) are introduced to achieve wheel-vehicle speed synchronization and to avoid small amplitude oscillations in the longitudinal tire forces for undriven wheels, which can cause numeric issues during simulation, especially at low speed.

$$\begin{cases} I_f \Delta\dot{\omega}_f = M_f - F_{xf}R_f \\ I_r \Delta\dot{\omega}_r = M_r - F_{xr}R_r \end{cases} \quad (5.5)$$

where: $\Delta\omega_{f0} = \Delta\omega_{r0} = 0$

$$\begin{cases} \omega_f = \Delta\omega_f + \frac{V_{fxf}}{R_f} \\ \omega_r = \Delta\omega_r + \frac{V_{rxr}}{R_r} \end{cases} \quad (5.6)$$

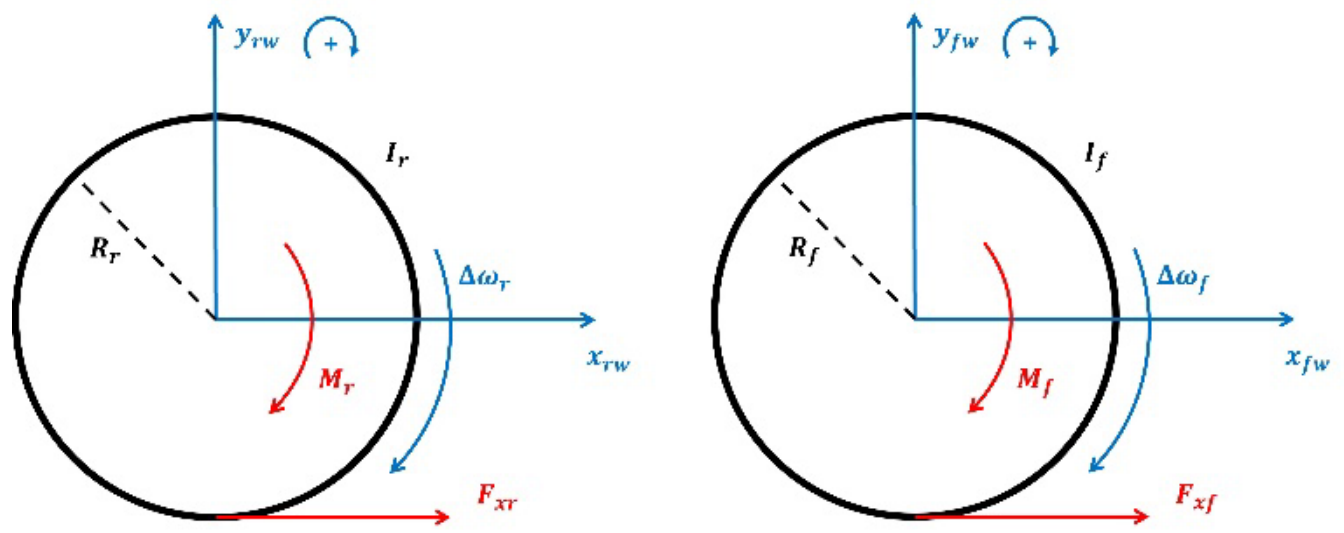

**Figure 5.3.** Wheel rotation model

**Table 5.3.** Wheel rotation model parameters

| Parameters | Descriptions |
|---|---|
| $\omega_f, \omega_r$ | Front & rear tire angular velocity [rad/sec] |
| $\Delta\omega_f, \Delta\omega_r$ | Front & rear tire deviated angular velocity [rad/sec] |
| $R_f, R_r$ | Front & rear tire radius [m] |
| $I_f, I_r$ | Front & rear tire moment of inertia |
| $M_f, M_r$ | Front & rear tire driving torque [Nm] |
| $V_{fxf}$ | Front axle longitudinal velocity in front wheel frame [m/sec] |
| $V_{rxr}$ | Rear axle longitudinal velocity in rear wheel frame [m/sec] |

Combining all the above-mentioned components, one can construct the overall vehicle model as illustrated in Figure 5.4. It can be observed that vehicle position $(X, Y)$ and orientation/yaw angle ($\psi$) can be obtained through model output post-processing.

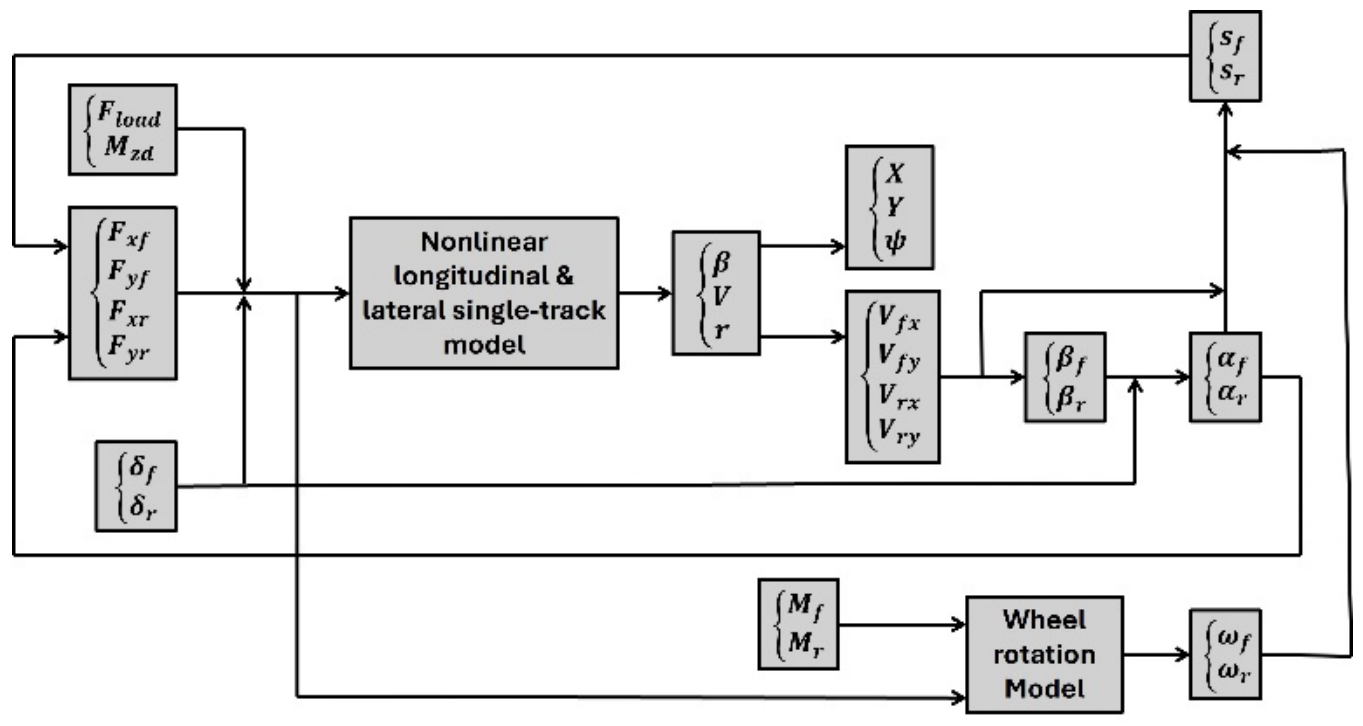

**Figure 5.4.** Overall vehicle model structure

## 5.2.2 Hardware-in-Loop Test

Hardware-in-the-Loop (HIL) testing is an exceptionally effective simulation and validation technique widely employed in the development and testing of autonomous driving functions. The core principle of HIL testing involves integrating actual hardware components in vehicles, such as autonomous driving controllers (MicroAutobox or MABX for low level control), GPS, and dedicated short-range communications (DSRC), etc., into the simulation environment. This is very close to real vehicle testing except the real vehicle is replaced by a vehicle model that can accurately simulate vehicle dynamics. This setup allows for the evaluation of hardware performance of autonomous driving functions under pre-set operating conditions in a laboratory. Additionally, HIL testing enables the evaluation of autonomous driving functions in real-time scenarios and under signal transmission delays, while also facilitating the detection of unrealistic extreme inputs that could affect system reliability. By conducting all experiments in the laboratory, HIL testing significantly reduces overall testing costs and avoids the risks associated with on-road trials. Furthermore, HIL testing serves as an indispensable precursor to public road testing, ensuring that autonomous driving systems are thoroughly evaluated for safety and performance before being deployed in real-world environments.

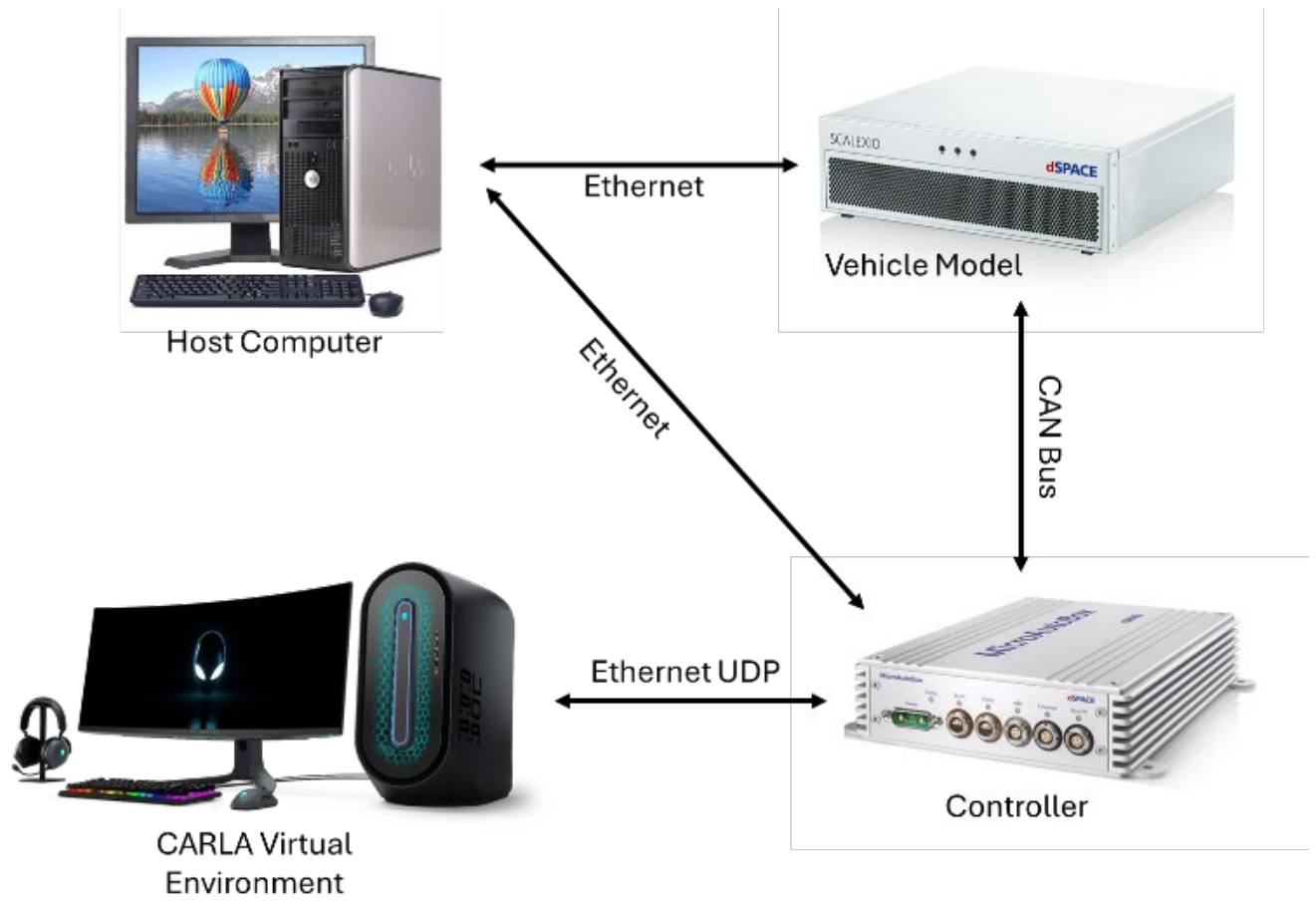

**Figure 5.5.** Hardware-in-the-Loop (HIL) test setup

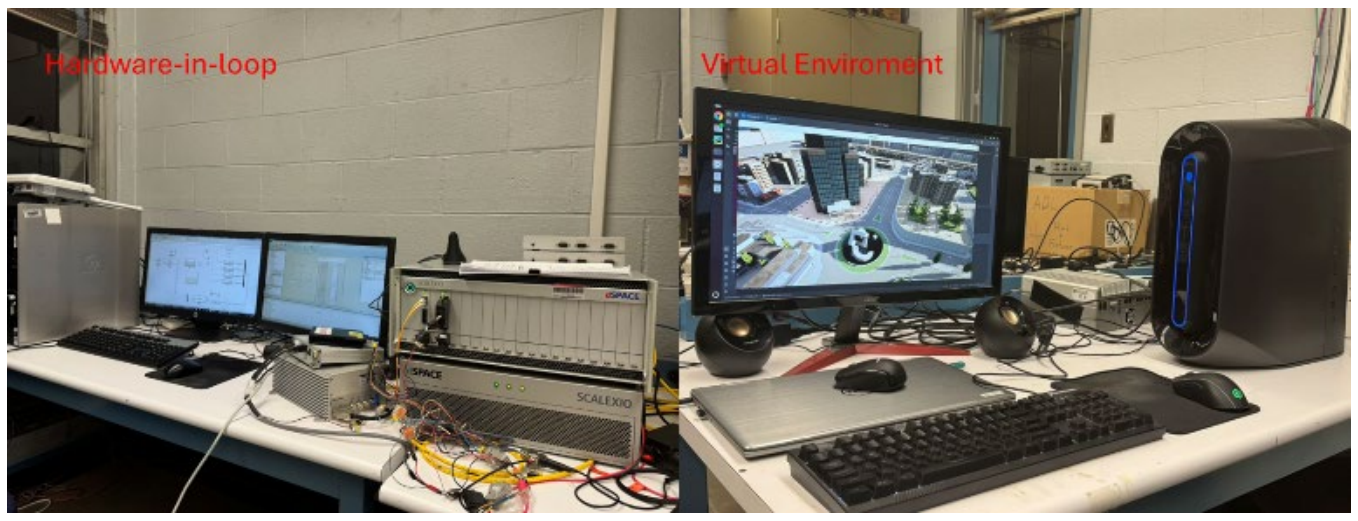

**Figure 5.6.** Hardware layout in the lab

Figure 5.5 demonstrates the Hardware-in-the-Loop (HIL) test setup employed in this section. Figure 5.6 illustrates our laboratory's hardware configuration, which includes a HIL simulation workstation, and a computer dedicated to managing the virtual environment. In our configuration, the aforementioned vehicle dynamic Simulink model operates on a Scalexio simulation computer in real time, continuously calculating the vehicle's position and state based on inputs such as throttle, brake, and steering. These vehicle states are transmitted via the CAN bus to the MicroAutobox (MABX) electronic control unit, which functions both as the vehicle's controller and as a communication hub between the vehicle model and the virtual environment.

Then, the MABX sends the current vehicle information back to the computer managing the virtual environment and receives environmental data in return. It processes this data to generate appropriate vehicle control commands. Communication between the MABX and the virtual environment computer is developed through Ethernet UDP. The virtual environment computer then uses data from the MABX to update the vehicle's status within the simulated environment, providing a visual representation of the experiment. Additionally, the Host Computer controls the operations of both the Scalexio and MABX systems, coordinating the overall simulation process.

This HIL setup establishes an integrated, real-time feedback loop where the vehicle dynamic model and corresponding virtual environment interact dynamically. This enables comprehensive testing of the vehicle control system under various simulated conditions without the need for physical road tests. As a result, HIL testing not only enhances safety by minimizing risks associated with on-road trials but also significantly reduces the time and costs involved in prototype development and verification.

Moreover, the HIL test setup is designed for seamless transition to a real vehicle including VVE testing. In real-world scenarios, the Scalexio simulation computer can be replaced with an actual vehicle, and the computer managing the virtual environment can be substituted with an in-vehicle PC capable of reading sensor data and analyzing the traffic environment. Additionally, the existing HIL architecture can be enhanced by integrating various other hardware components to increase the realism and reliability of the simulation. For example, DSRC or other communication modules can be added to simulate Vehicle-to-Vehicle (V2V) or Vehicle-to-VRU (V2VRU) communications, roadside units (RSU) can be incorporated to transmit real infrastructure data to the HIL system, electronic horizon modules can be included to obtain accurate map information and traffic control computers can be added to generate signal phase and timing (SPaT) and MAP messages. These enhancements allow for a more comprehensive and realistic testing environment, further bridging the gap between simulated and real-world autonomous driving scenarios. By incorporating these additional hardware elements, the HIL setup becomes a versatile and robust platform for both development and extensive validation of autonomous driving systems before their deployment on public roads.

### 5.2.3 Vehicle-In-Virtual-Environment (VVE) Test

Given that Hardware-in-the-Loop (HIL) employs all the real equipment necessary for ADAS testing except for the real vehicle, the natural next step would be to commence testing that involves the real vehicle, currently the most common approach of which is to perform public road testing. Carrying out the test on public roads, however, comes with several critical drawbacks. Firstly, other road users are involuntarily involved in the testing of the ADAS systems, which poses safety concerns, especially during extreme and edge cases. Secondly, the previously mentioned extreme and edge cases are typically rare occurrences, and would require significant mileage to encounter and test, limiting the overall efficiency of the approach. In that regard, the Vehicle-in-Virtual-Environment (VVE) was proposed as a novel approach to perform real-vehicle testing in a safe and efficient manner.

The overall architecture of the VVE approach is displayed in Figure 5.7. The real vehicle is operated in a safe and open testing space, where its motions are synchronized, via frame transformation, with those of a virtual vehicle operating in a highly realistic virtual environment [104]. Depending on the desired traffic scenario to be tested, the virtual environment can be edited with relative ease. Virtual onboard sensors of our choice can be fitted to the virtual vehicle and its data collected from the virtual environment can be fed into the onboard equipment of the real vehicle so that the real vehicle control unit can react to a virtual traffic scenario.

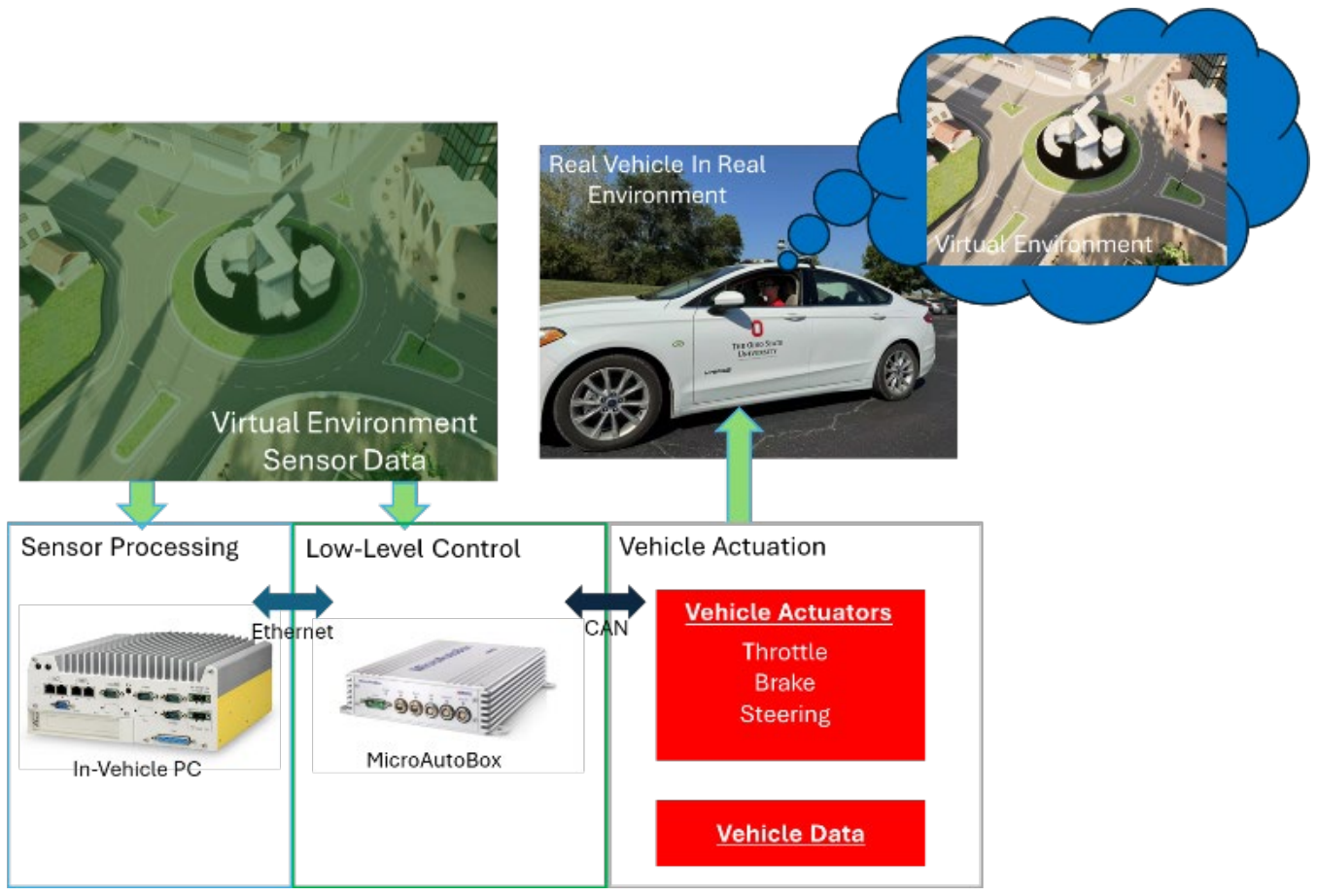

**Figure 5.7.** VVE Architecture

With the synchronization in place, difficult, rare and safety critical tests can be conducted by simply editing the virtual environment to create the respective traffic scenarios, reducing the cost of testing. This method is also significantly safer, as the vehicle operates in a separate open space and does not run the risk of a real traffic accident should the ADAS system being tested fail the safety critical experiment. The utilization of real vehicle dynamics is another benefit of the VVE approach, as this simulation-like approach combines the benefits of real vehicle testing without invoking any safety and cost drawbacks. This approach also provides the possibility of multi-actor experiments. Figure 5.8 provides an example for this type of test in the form of a Vehicle-to-Pedestrian (V2P) communication-based collision avoidance experiment. The real pedestrian and the real vehicle operate in separate spaces that are safe and open. The real pedestrian is equipped with a mobile phone that has IMU and GPS sensors installed as well as a mobile application that calculates the heading and position of the pedestrian, and this motion information is broadcast via Bluetooth low-energy (BLE) connection. This pedestrian motion data, together with the vehicle motion data, are synchronized through frame transformation into a virtual pedestrian and a virtual vehicle operating within the same virtual environment where virtual vehicle-to-pedestrian collision is possible. V2P-based collision avoidance experiment can hence be carried out under this setup, where the virtual vehicle feeds the virtual pedestrian motion data into the real vehicle so that the real vehicle can react to avoid collisions in the virtual world.

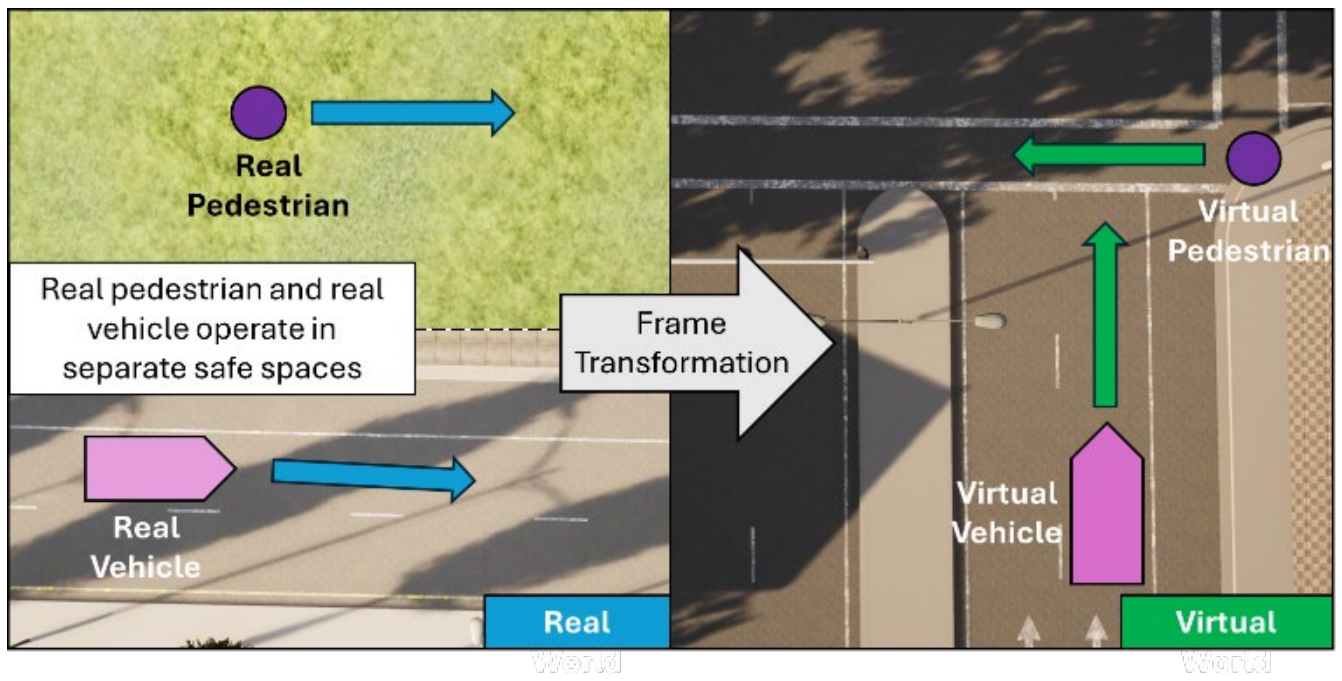

**Figure 5.8.** V2P test using VVE approach

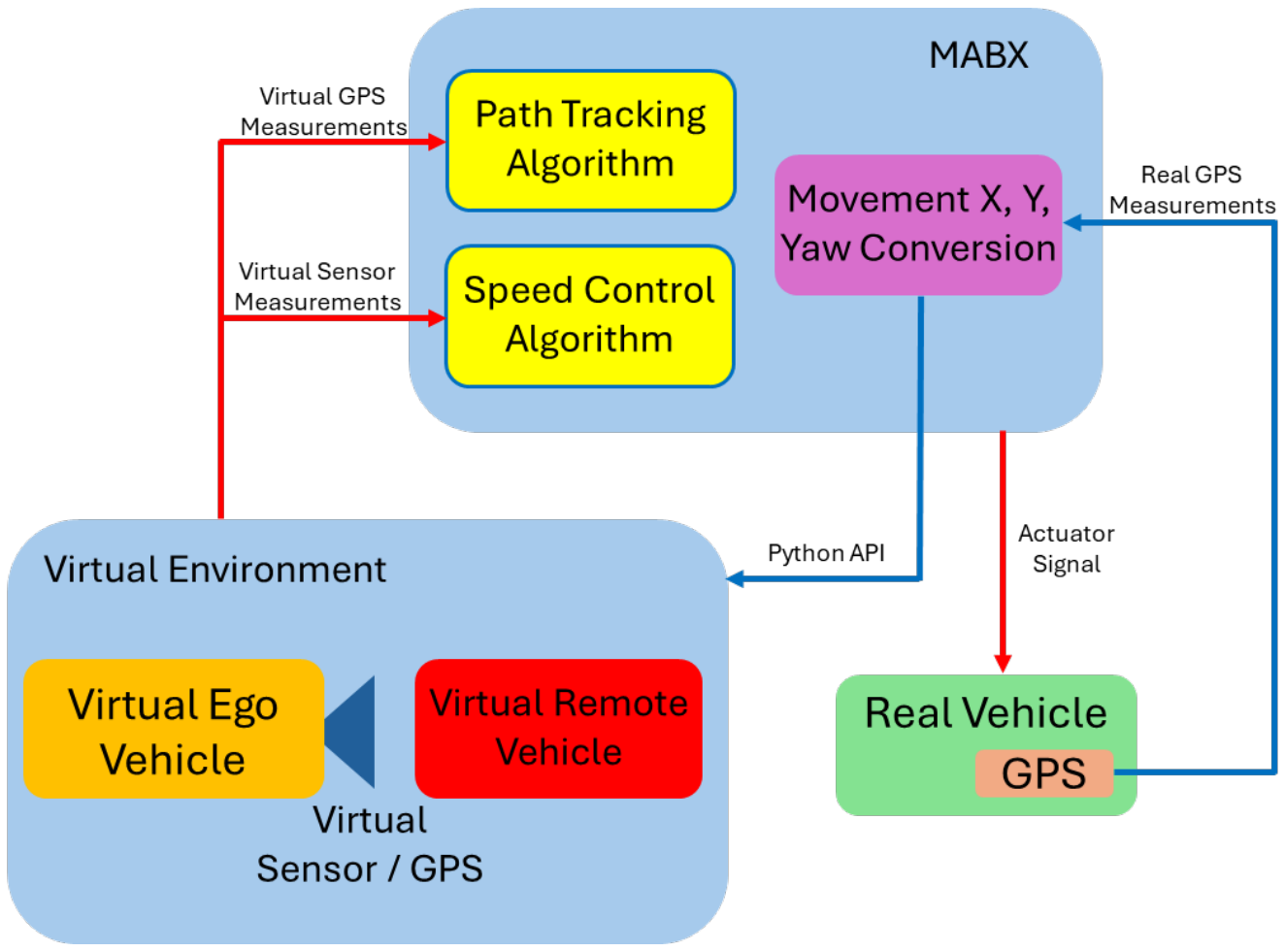

**Figure 5.9.** Implementation structure of VVE approach

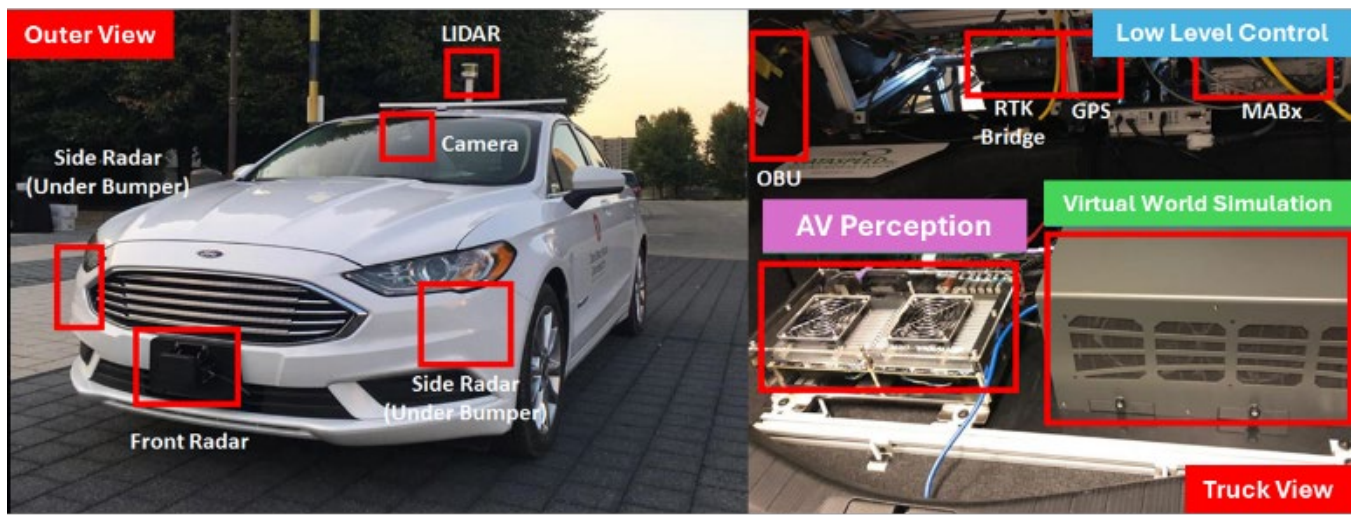

**Figure 5.10.** Test vehicle used for VVE approach

Figure 5.9 illustrates the current implementation structure of the corresponding VVE setup, and Figure 5.10 displays the equipment onboard our test vehicle. The real vehicle has an RTK GPS unit with differential antennas that provides us with both vehicle position and heading information, and this data is fed into the dSpace microautobox (MABX) unit, which is our onboard control unit, before being sent via Ethernet UDP protocol to an in-vehicle PC that runs an Unreal Engine-based CARLA virtual environment. The virtual environment applies the frame transformation routine to the received data to achieve the desired real-virtual synchronization. On the other hand, the virtual sensor data collected in the virtual environment is fed back into the MABX unit using Ethernet UDP protocol again.

## 5.3 Results

To demonstrate the effectiveness of the proposed VVE based testing pipeline, we designed a simple Double Deep Q-Network (DDQN) based emergency brake algorithm for vulnerable road users and evaluatde its performance using the proposed pipeline. The design details of this algorithm can be found in our published paper [106]. We conducted a step-by-step evaluation of each component within the pipeline to clearly demonstrate how each contributes to identifying potential weaknesses and for validating the robustness of the algorithm.

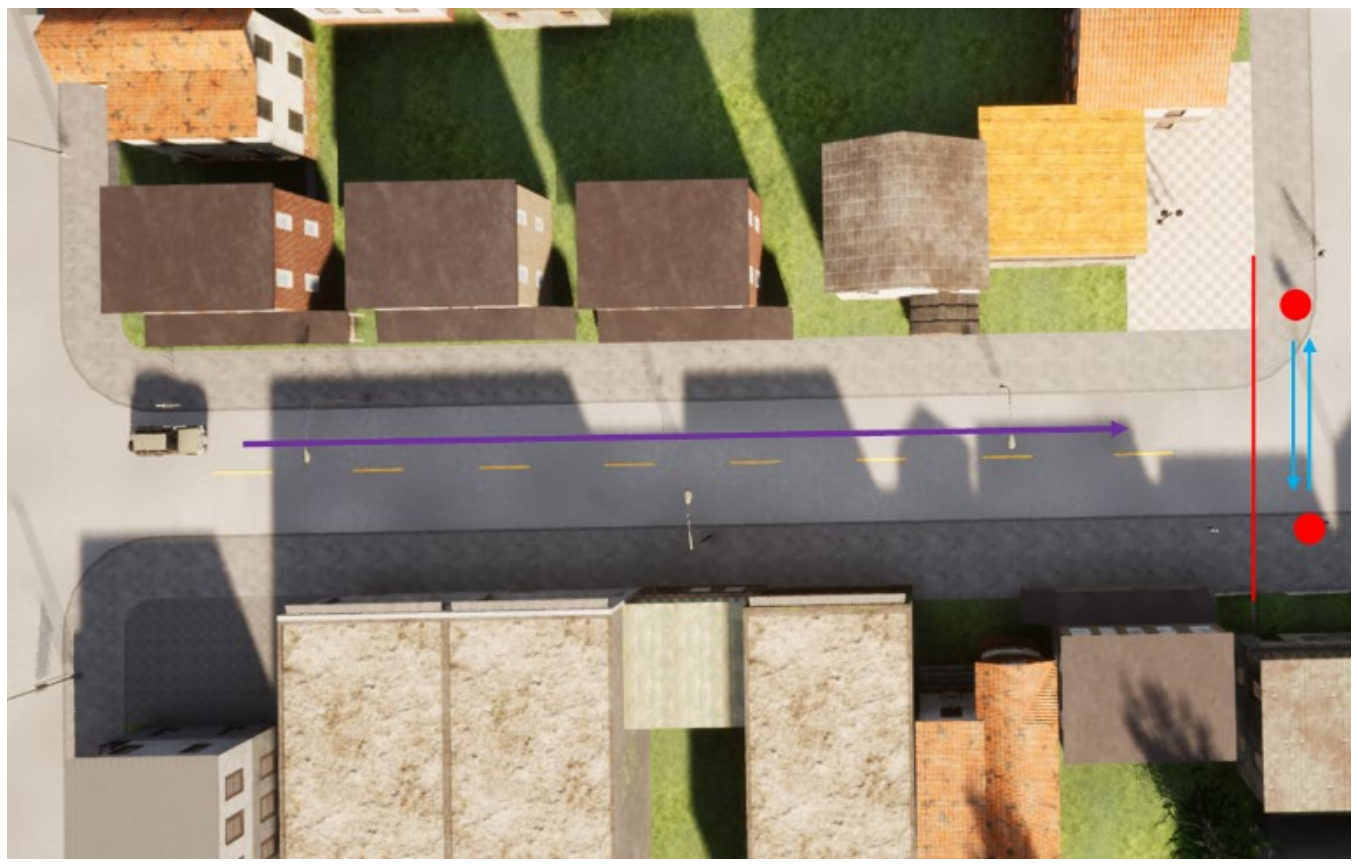

**Figure 5.11.** Traffic scenario used for testing.

Figure 5.11 demonstrates the traffic scenario within the virtual environment used to evaluate the overall performance of the DDQN-based autonomous driving agent. In this scenario, the vehicle navigates along a street while two pedestrians cross the crosswalk and walk back and forth. The vehicle is tasked with maintaining a comfortable speed and coming to a safe stop before the crosswalk when pedestrians are present. This test scenario is specifically designed to evaluate the agent's ability to perform emergency braking maneuvers effectively when encountering unexpected pedestrian movements. By simulating such dynamic and potentially hazardous conditions, this scenario ensures that the DDQN agent can reliably respond to emergencies, demonstrating its capability to maintain safety and control in various traffic situations.

### 5.3.1 Model-in-Loop Test Results

We begin by evaluating the effectiveness of the proposed emergency braking algorithm using a MIL setup. This initial stage allows us to validate the algorithm's core decision-making logic in a simulated environment before introducing hardware dependencies. The results of the MIL tests are presented below.

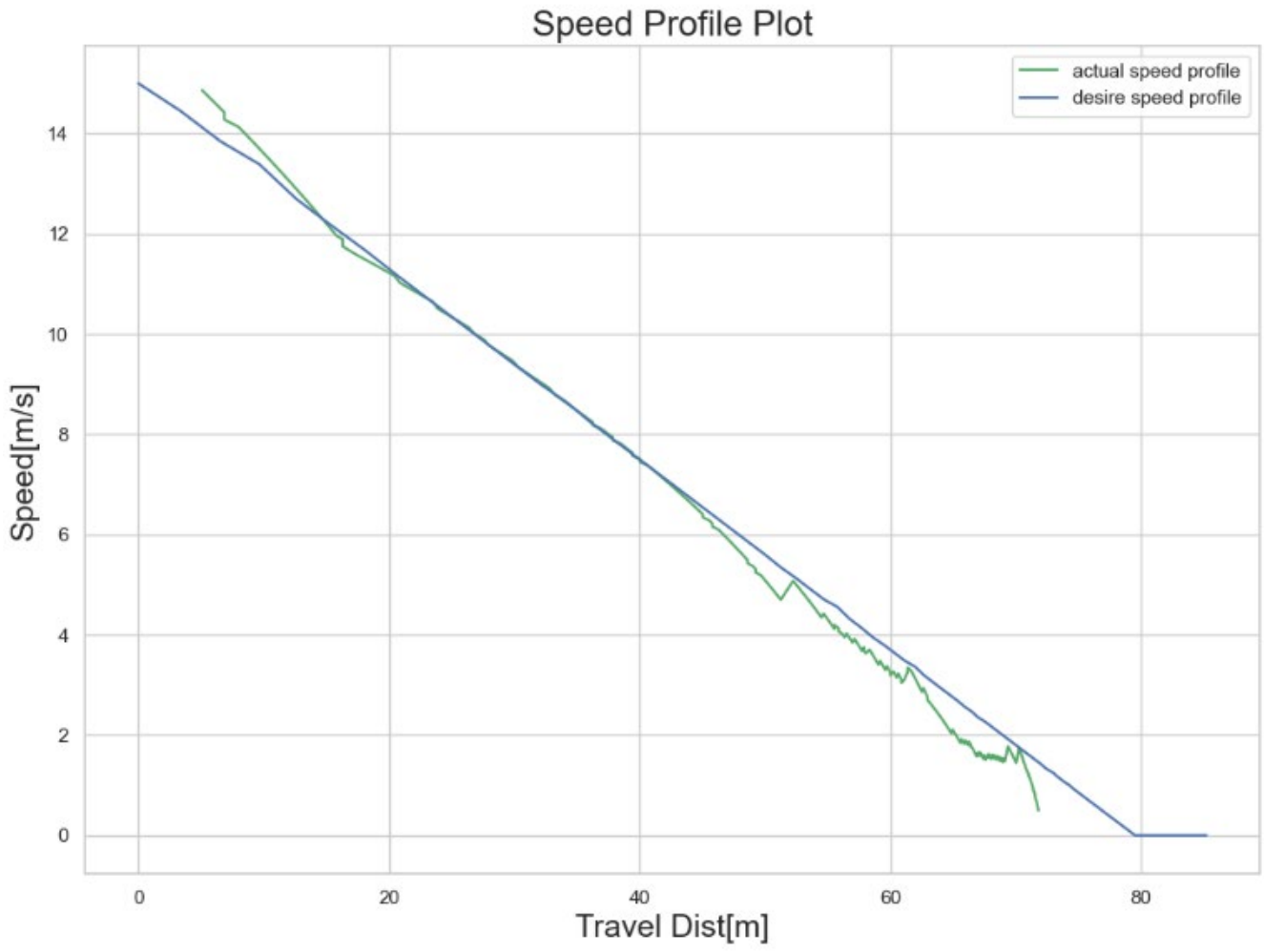

**Figure 5.12.** DDQN autonomous driving agent speed tracking plot with initial speed 15m/s

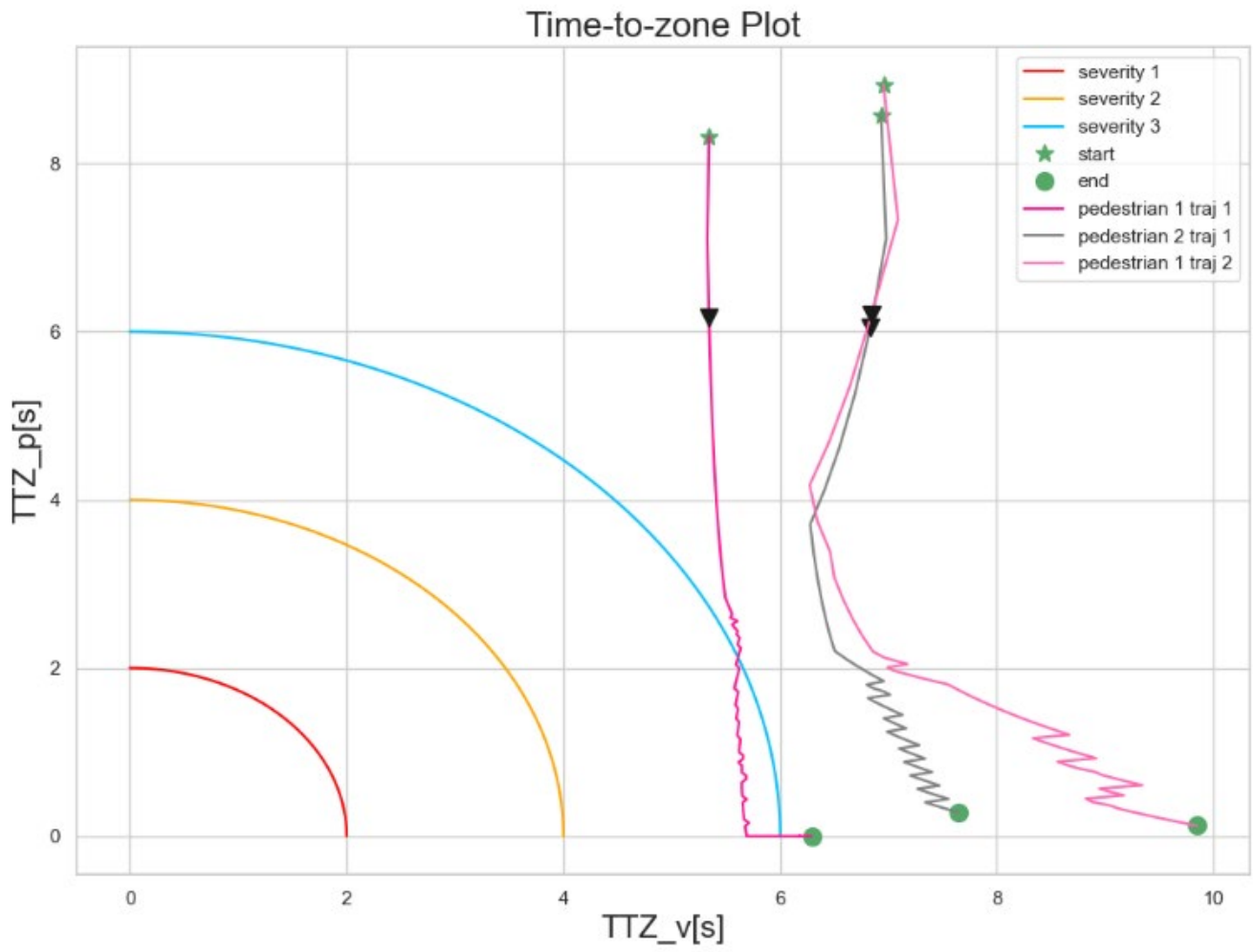

**Figure 5.13.** DDQN autonomous driving agent MIL test TTZ plots with initial speed 15m/s

Figure 5.12 presents the speed tracking performance of the DDQN-based autonomous driving agent, starting with an initial speed of 15 m/s. Initially, the vehicle maintains the set speed of 15 m/s and gradually decelerates to follow the desired braking profile which is decreasing its speed smoothly from 15 m/s to 0 m/s. However, at lower speeds, the vehicle exhibits some abrupt braking behavior, resulting in less precise tracking of the desired braking speed. This occurs because the vehicle's position is very close to the pedestrians, prompting the agent to brake more aggressively to avoid potential collisions. Such behavior demonstrates the agent's prioritization of safety by ensuring timely and sufficient deceleration when immediate threats are detected.

Figure 5.13 illustrates the Time-To-Collision (TTZ) performance for the DDQN autonomous driving agent, starting with an initial speed of 15 m/s. In the TTZ plots, red circles indicate both pedestrian and vehicle's TTZ times less than 2 seconds (indicating very urgent situations), orange circles represent both pedestrian and vehicle's TTZ times less than 4 seconds, and blue circles represent both pedestrian and vehicle's TTZ times less than 6 seconds. Throughout the entire duration of the simulation, the TTZ correspondence between the vehicle and the two pedestrians consistently remains above 4 seconds and often exceeds 6 seconds. This consistent maintenance of safe TTZ values highlights the effectiveness of the DDQN agent in ensuring pedestrian safety, even in emergency scenarios. By keeping socially acceptable distance [42] and reacting promptly to dynamic obstacles, the DDQN agent successfully minimizes the risk of collisions, demonstrating its capability to handle critical situations reliably.

### 5.3.2 Hardware-in-the-Loop Test Results

Next, we evaluate the effectiveness of the proposed emergency braking algorithm using a HIL setup. This stage enables us to assess the algorithm's real-time performance and hardware compatibility under a more realistic traffic simulation environment. The HIL test results are presented below.

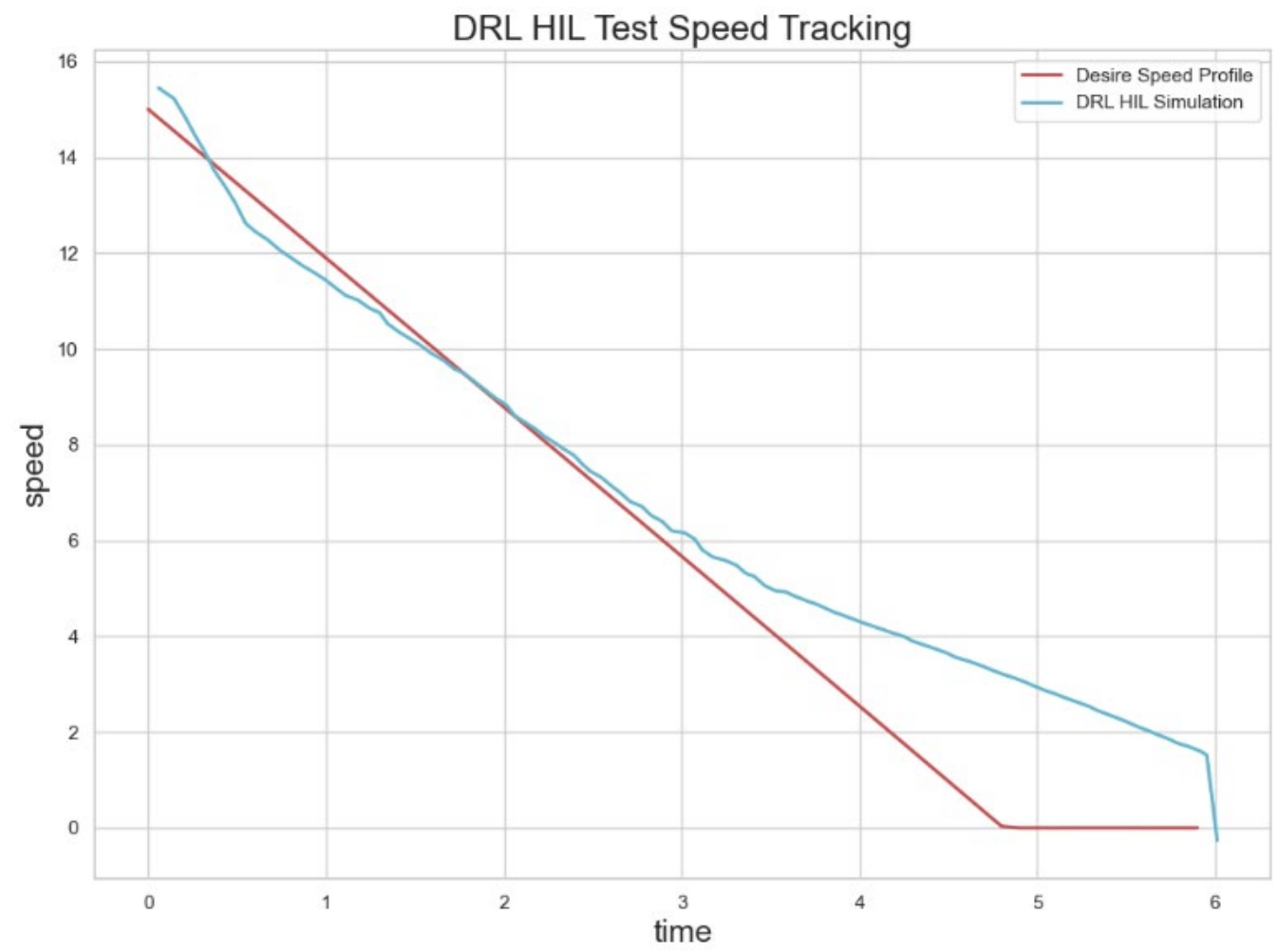

**Figure 5.14.** DDQN autonomous driving agent HIL simulation speed tracking plot with initial speed 15m/s

Figure 5.14 demonstrates the speed tracking performance of the DDQN-based autonomous driving agent during HIL simulation tests, starting with an initial speed of 15 m/s. Overall, the trained model successfully tracks the desired speed profile and comes to a complete stop before the crosswalk under real-time HIL simulation conditions. Consistent with the MIL tests, the agent exhibits a sudden increase in braking intensity at lower speeds, enabling the vehicle to decelerate rapidly and halt to ensure pedestrian safety.

Figure 5.15 illustrates the HIL test Time-To-Collision (TTZ) performance for the DDQN autonomous driving agent, starting with an initial speed of 15 m/s. Throughout the entire duration of the simulation, it can be observed that the TTZ difference between the pedestrian and the vehicle is also larger than 4 seconds most of time, indicating the low likelihood of collision, proving the effectiveness of our emergency brake agent.

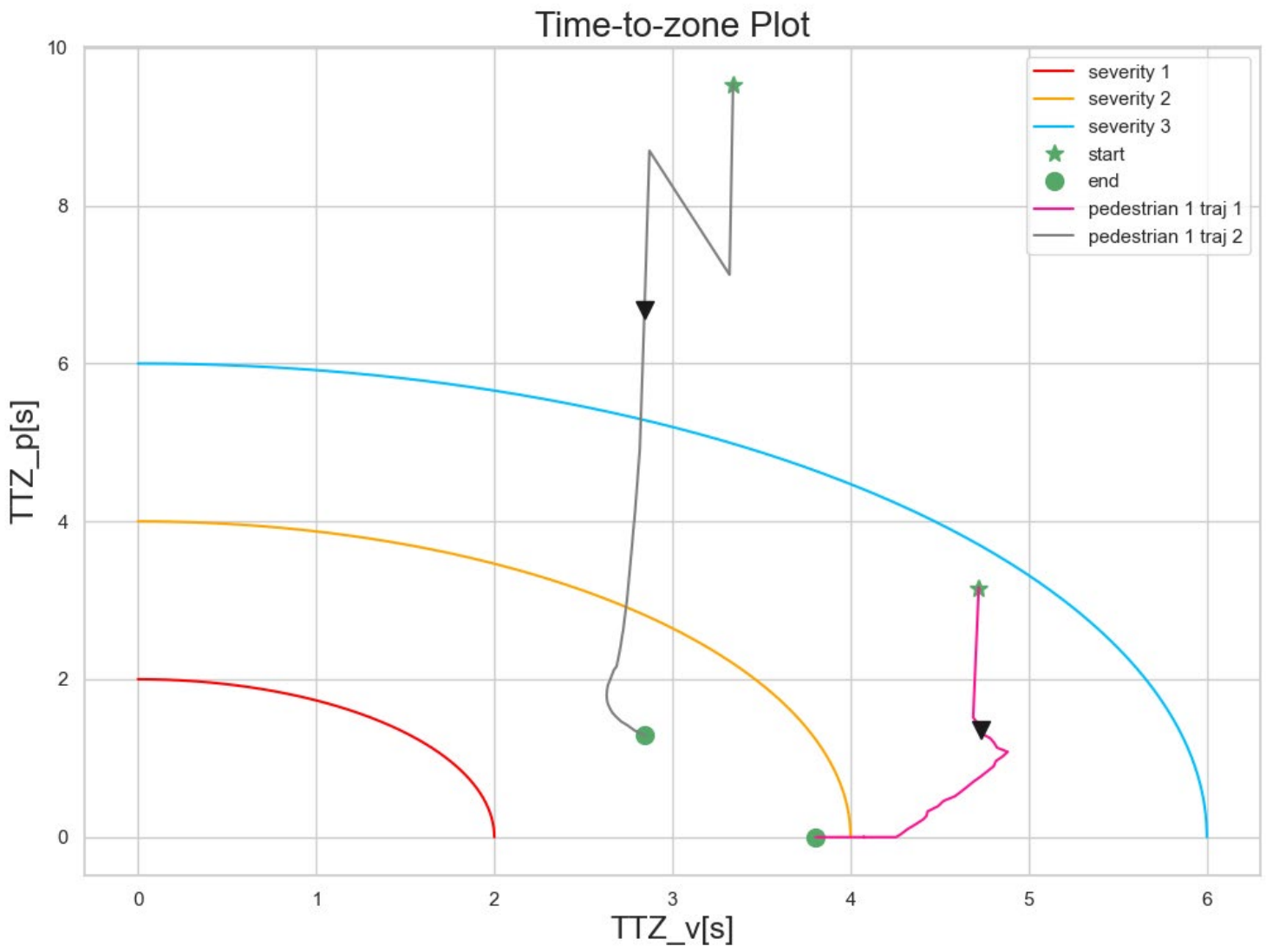

**Figure 5.15.** DDQN autonomous driving agent HIL test TTZ plots with initial speed 15m/s

During the HIL tests, we evaluated the model's performance with the integration of actual hardware, and the results were highly satisfactory. The agent maintained accurate speed tracking and demonstrated reliable emergency braking behavior, even when interacting with physical components of the system. This positive performance in the HIL environment indirectly confirms the viability of our proposed development and testing pipeline. The successful integration and real-time responsiveness of the DDQN agent in the HIL setup indicates that our approach is both feasible and effective for advancing autonomous driving systems.

### 5.3.3 Vehicle-In-Virtual-Environment (VVE) Test Results

Due to limitations in testing ground size and the fact that our DDQN-based method serves as proof of concept without hard-coded safety rules, it is not yet suitable for deployment on real vehicles. As a result, we did not conduct VVE testing for the DDQN emergency brake algorithm. Instead, we first performed VVE testing using manual driving, with the results documented in a previous paper [104]. Subsequently, we tested the vehicle's autonomous path-tracking and obstacle avoidance capabilities using the VVE framework. Specifically, a collision-free trajectory was generated using a collision avoidance algorithm, and a path-tracking controller was employed to follow the planned route. We deployed this path planning and tracking controller onto the experimental vehicle for VVE testing. Using the VVE setup, we then tested the performance of this path planning and tracking system. This process demonstrates that the VVE framework can accurately evaluate not only the functional feasibility of the deployed algorithms but also their real-time performance and interaction with physical components in the real world. The VVE test results are presented below.

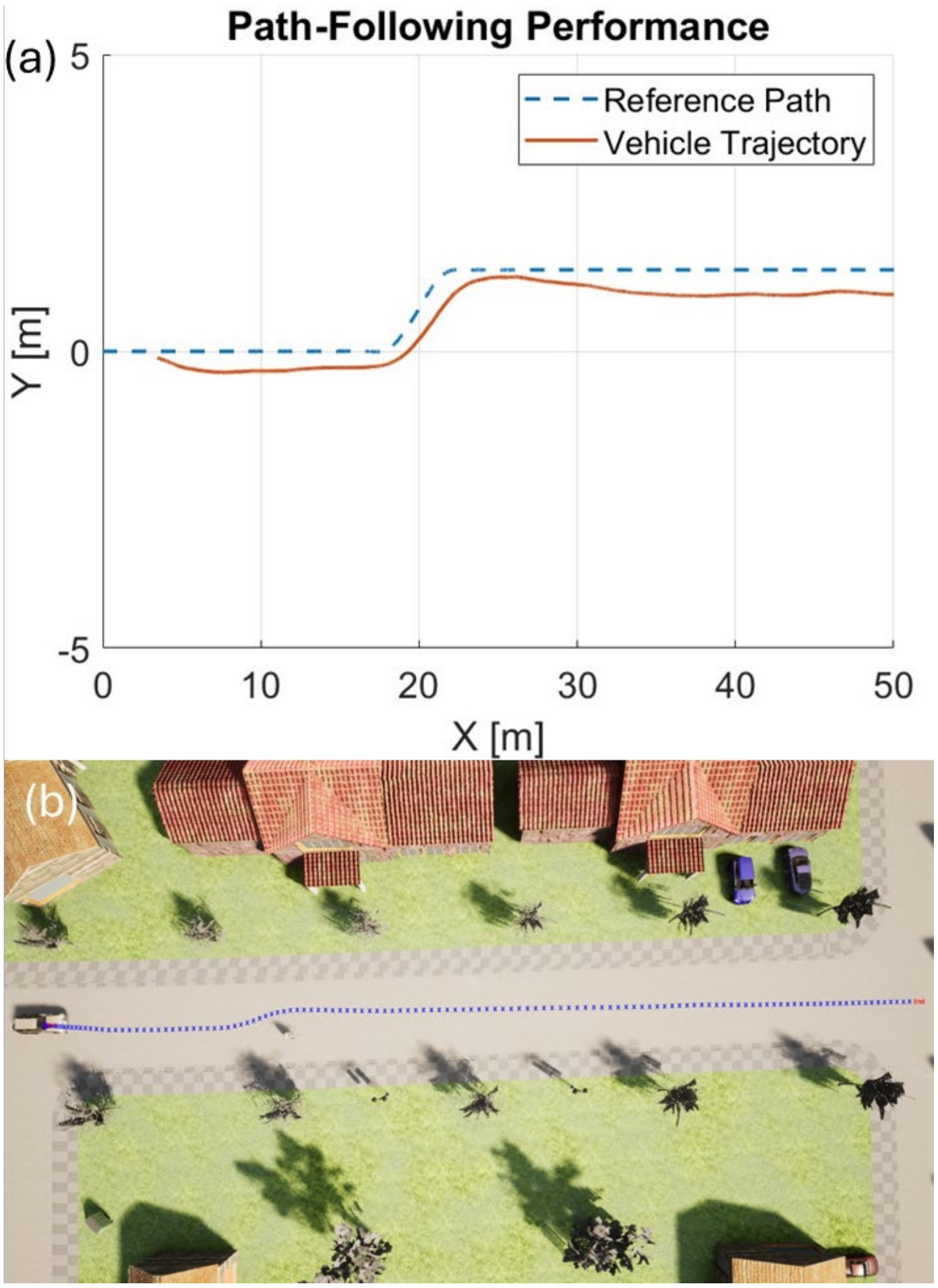

**Figure 5.16.** VVE Autonomous Collision Avoidance Path Tracking Test: (a) Motion trajectory in the real world; (b) Motion trajectory in the virtual world

Figure 5.16 demonstrates autonomous collision avoidance path tracking test result using the proposed VVE framework in both real world and virtual world. In this experiment, we use VVE method to evaluate the capabilities of a pure pursuit-based path-tracking controller. As shown in Figure 5.16 (a), the real-world trajectory of the vehicle is compared with the planned reference path. In this test scenario, a bicyclist is positioned in the current lane ahead of the ego vehicle, requiring the collision avoidance algorithm to generate a single-lane-change trajectory. The pure pursuit controller is then required to follow this trajectory to safely avoid the bicyclist. From Figure 5.16 (a), we can observe that the vehicle closely follows the reference path, though small tracking errors are present. These deviations are likely caused by factors such as GPS inaccuracies and the absence of RTK usage, leading to slight position offsets. Figure 5.16(b) shows the simulation trajectory in the virtual world. From Figure 5.16(b), we replicated the aforementioned traffic scenario and confirmed that the pure pursuit controller can successfully execute the single-lane change maneuver to avoid the bicyclist obstacle.

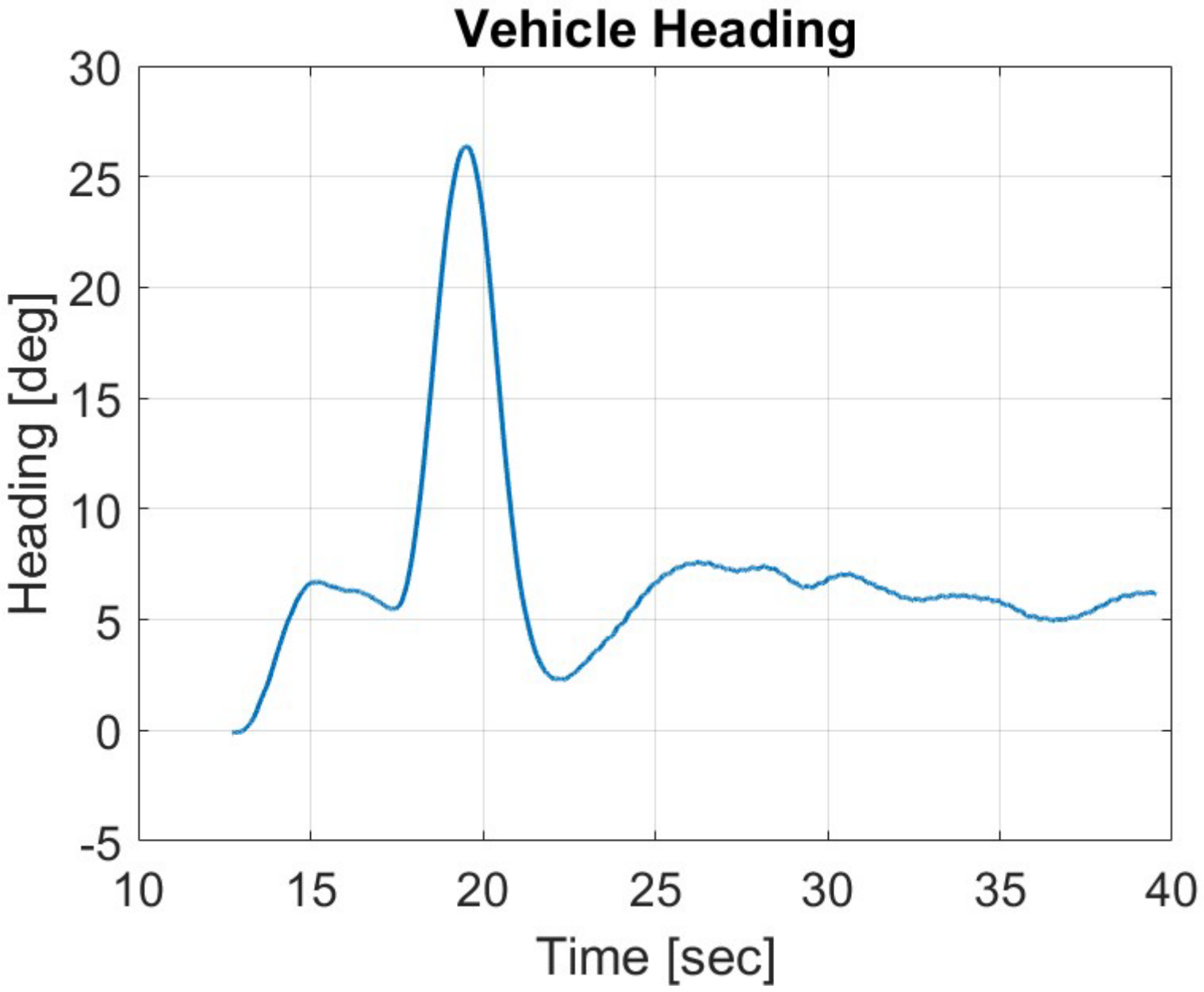

**Figure 5.17.** VVE Autonomous Collision Avoidance Path Tracking Test Yaw Angle History

Figure 5.17 demonstrates yaw angle change of the vehicle in autonomous collision avoidance path tracking test. It can be observed that the vehicle heading remains relatively stable for most of the time, except for a significant change between approximately 17 s and 21 s. This sharp variation indicates that the vehicle is actively executing the single-lane change maneuver to avoid the bicyclist obstacle on its original path. After the lane change is completed, the heading angle stabilizes again, demonstrating the controller's ability to quickly adjust vehicle orientation and return to a steady-state heading.

This experiment demonstrates our capability to test autonomous driving functions using proposed VVE framework. One of the key advantages of VVE is that it allows us to create complex and potentially dangerous real-world scenarios (e.g., collisions with VRUs) in a safe, fully virtual environment, significantly reducing testing risks. Furthermore, the VVE framework enables real-world vehicle dynamics and VRUs dynamics in experiments which large enhance the realism of the test, making it a powerful tool for rapid development, demonstration and evaluation of collision avoidance algorithms.

## 5.4 Conclusion

In this section, we proposed a new testing pipeline that sequentially integrates MIL, HIL, and VVE methods to comprehensively develop and evaluate autonomous driving functions. To demonstrate the effectiveness of this testing pipeline, we designed a deep reinforcement learning based emergency braking algorithm for vulnerable road users and evaluated its performance using the proposed pipeline. We conducted a step-by-step evaluation of each component within the pipeline to clearly demonstrate how each contributes to identifying potential weaknesses and validating the robustness of the algorithm. All of these tests confirmed the feasibility and effectiveness of our proposed experimental methods. By employing this comprehensive testing pipeline, we can efficiently develop advanced and robust autonomous driving systems using artificial intelligence and other cutting-edge algorithms. This approach allows us to thoroughly test the effectiveness of autonomous driving systems under various traffic conditions. Due to time and testing space constraints, a full test of the deep reinforcement learning (DRL)-based collision avoidance algorithm is still in progress and will be included in the future.

Despite the advancements presented in this section, the proposed testing pipeline still has several limitations that need to be addressed in future work. One significant issue is the presence of singular points in the vehicle dynamic model used during Model-in-the-Loop (MIL) and Hardware-in-the-Loop (HIL) testing. Specifically, when the vehicle speed is exactly zero or when input values are extremely large, the simulated data may become inaccurate. This limitation restricts the effectiveness of MIL testing. To overcome this challenge, future efforts will focus on refining the vehicle dynamics model to eliminate these singularities and improve simulation accuracy.

Moreover, transitioning directly from MIL to HIL sometimes leads to overlooking potential software-specific issues that could arise during the interaction between the code and the system. This can result in increased debugging efforts and delays during HIL testing, as these issues are more challenging to diagnose in a hardware-dependent environment. Therefore, we plan to incorporate Software-in-loop (SIL) into our proposed testing pipeline. By incorporating SIL testing after MIL, we can thoroughly validate the software in a simulated yet realistic setting, ensuring that it functions correctly and interacts smoothly with the system. This additional step enhances the overall reliability and efficiency of the testing pipeline, paving the way for a smoother transition to HIL and ultimately contributing to a more robust and well-tested system.

In addition to these improvements, future research will conduct a thorough analysis of the effects of communication latency and computation delays on the Vehicle-in-Virtual-Environment (VVE) method. Understanding these impacts is crucial for optimizing system performance and ensuring real-time responsiveness. Moreover, future work will explore the inclusion of other types of road users, such as vehicular traffic, to expand the range of testing scenarios. Furthermore, integrating extended reality (XR) goggles into the testing system is another aspect we intend to explore and our preliminary analysis and exploration in this area was very promising. By immersing real pedestrians or drivers into the virtual environment, XR technology can further enhance the realism of testing scenarios, providing more authentic interactions and valuable data for refining autonomous driving algorithms.

# Chapter 6: Future Work

In conclusion, the rapid urbanization and increase in privately owned vehicles have worsened traffic congestion and car accidents, posing significant challenges for modern cities. Autonomous driving systems offer a promising solution to mitigate these issues by reducing the human errors responsible for a substantial number of accidents.

In this two-year project, we focused on applying the Vehicle-in-Virtual-Environment (VVE) method to develop, evaluate, and demonstrate safety functions for Vulnerable Road Users (VRUs). In the current second year project, our primary focus was on bicyclist safety. We began by analyzing five key bicyclist crash scenarios identified by the Fatality Analysis Reporting System (FARS), an organization under the National Highway Traffic Safety Administration (NHTSA) that compiles vehicle crash data.

Building on these cases, our research fills critical gaps in the current literature by proposing practical solutions to the types of VRU-related crashes represented in the FARS dataset. Specifically, this research makes the following three contributions:

(1) Proposing a communication-disturbance-observer (CDOB) based delay-tolerant control strategy to enhance robustness of the control under computation latency, network latency and packet loss.

(2) Developing a hierarchical control framework that integrates deep reinforcement learning (DRL) for high-level decision-making with a CLF-CBF-QP-based controller for safe and smooth low-level execution.

(3) Introducing a novel VVE-based testing pipeline that enables efficient and rigorous evaluation of autonomous driving functions under various complex and high-risk traffic scenarios for improving the safety of vulnerable road users.

Looking ahead to the third year, UTC project will continue to focus on utilizing the Vehicle-in-Virtual-Environment (VVE) method to develop, evaluate, and demonstrate safety functions for all road users. Our primary objectives include the implementation, evaluation, and comparison of various collision avoidance algorithms, such as CLF-CLF-QP, Supervised Learning, Deep Reinforcement Learning, and other state-of-the-art (SOTA) methods. We aim to comprehensively understand their performance, characteristics, applicability, and limitations, eventually building a benchmarking framework for future research. Additionally, we plan to integrate mixed-reality technologies into our VVE method. By incorporating XR goggles, participants can fully immerse themselves in the virtual environment, enhancing the realism and overall quality of the experiments.